%% file: PaperForReview.tex
\documentclass[10pt,twocolumn,letterpaper]{article}

\usepackage[pagenumbers]{cvpr} %

\usepackage{graphicx}
\usepackage{amsmath}
\usepackage{amssymb}
\usepackage{booktabs}
\usepackage{multirow}
\usepackage{bm}
\usepackage{array}
\usepackage{makecell}
\usepackage{enumitem}
\usepackage{float}
\usepackage{cuted}
\usepackage{subcaption}

\usepackage[mathscr]{eucal}

\newcommand{\tensor}[1]{\bm{\mathscr{#1}}}

\newcommand{\ra}[1]{\renewcommand{\arraystretch}{#1}}

\newcommand{\figref}[1]{Figure~\ref{#1}}

\newcommand{\XNAS}{\texttt{XNAS}}

\usepackage{pifont}%
\newcommand{\cmark}{\ding{51}}%
\newcommand{\xmark}{\ding{55}}%

\usepackage[pagebackref,breaklinks,colorlinks]{hyperref}

\usepackage[capitalize]{cleveref}
\crefname{section}{Sec.}{Secs.}
\Crefname{section}{Section}{Sections}
\Crefname{table}{Table}{Tables}
\crefname{table}{Tab.}{Tabs.}

\def\cvprPaperID{6033} %
\def\confName{CVPR}
\def\confYear{2023}

\begin{document}

\title{Learning Interpretable Models Through Multi-Objective NAS}

\author{Zachariah Carmichael$^{1,2}$ \quad Tim Moon$^2$ \quad Sam Ade Jacobs$^2$\\
$^1$University of Notre Dame \quad $^2$Lawrence Livermore National Laboratory \\
{\tt\small zcarmich@nd.edu} \quad {\tt\small \{moon13,jacobs32\}@llnl.gov}
}

\maketitle

\begin{abstract}
   Monumental advances in deep learning have led to unprecedented achievements across various domains. While the performance of deep neural networks is indubitable, the architectural design and interpretability of such models are nontrivial. Research has been introduced to automate the design of neural network architectures through \textit{neural architecture search}~(NAS). Recent progress has made these methods more pragmatic by exploiting distributed computation and novel optimization algorithms. However, there is little work in optimizing architectures for interpretability. To this end, we propose a multi-objective distributed NAS framework that optimizes for both task performance and ``introspectability,'' a surrogate metric for aspects of interpretability. We leverage the non-dominated sorting genetic algorithm~(NSGA-II) and explainable AI~(XAI) techniques to reward architectures that can be better comprehended by domain experts. The framework is evaluated on several image classification datasets. We demonstrate that jointly optimizing for task error and introspectability leads to more disentangled and debuggable architectures that perform within tolerable error.
\end{abstract}

\section{Introduction}

The success of deep learning is seemingly ubiquitous in a multitude of domains. A core component of its effectiveness is its ability to automate the feature engineering process. Under this perspective, a natural next step is the automation of the architecture design. To this end, \textit{neural architecture search}~(NAS)~\cite{elsken2019neural} has been proposed.
Progress in NAS has led to results that supersede the state-of-the-art in several applications, such as image classification~\cite{DBLP:conf/aaai/RealAHL19} and object detection~\cite{DBLP:conf/cvpr/ZophVSL18}.

While NAS has been effective in automating architecture discovery and topping leaderboards, little attention has been paid to the discovery of interpretable architectures.
The automation of interpretability would further minimize the need for ``human-in-the-loop'' pipelines. Not only does this reduce the manual  design needed to meet the constraints of an application, but it also enhances the comprehensibility of the discovered models. Increased comprehensibility aids in model debugging, decreases time to deployment, and instills greater trust.

Unfortunately, there is a trade-off between increased interpretability and accuracy that bears a point at which
the degradation in model quality cannot be justified. A method to quantify these trade-offs would be of great importance to NAS endeavors. Furthermore, the acceptable compromise between these metrics will vary between applications -- consider the accuracy needed in biometrics, the explainability required for scientific discovery, and the combination required for medical diagnoses.
To this end, we introduce a framework for the joint optimization of task performance and a surrogate for interpretability. In this work, we put forth the following contributions:
\begin{itemize}[noitemsep,nolistsep,leftmargin=1em]
    \item We develop a new metric to quantify interpretability as the disentanglement between latent representations for different data classes: \textit{introspectability}. We further extend this metric by exploiting hierarchical semantic information from the WordNet database.
    \item We propose a multi-objective evolutionary approach to NAS, e\underline{X}plainable \underline{NAS}~(\XNAS{}), that maximizes both accuracy and introspectability by directly optimizing the Pareto frontier.
    \item We conduct analyses of the accuracy-introspectability trade-off, explore phylogenetic trees to understand the inheritability of objectives, visualize disentangled representations, analyze architectural motifs along the Pareto front, and demonstrate introspectability as a surrogate for trustworthiness and debuggability.
\end{itemize}

\section{Background \& Related Work}\label{sec:background}

Our work is at the intersection of neural architecture search, explainable AI~(XAI), evolutionary algorithms, and multi-objective optimization.
We present a brief overview of these topics and discuss related work to put our contributions in context.

\paragraph{Neural Architecture Search (NAS)}
Akin to how deep learning is used to automate feature engineering, NAS algorithms automate \emph{architectural engineering}~\cite{elsken2019neural}.
NAS algorithms can generally be understood as the composition of three elements: (i) a \emph{search space} that defines the possible neural architectures, (ii) a \emph{search strategy} that explores a search space for candidate solutions, and (iii) a \emph{performance estimation strategy} that determines the fitness of a solution.
Of the many approaches to NAS, Bayesian optimization~(BO), reinforcement learning~(RL), and evolutionary algorithms are the most common.
While BO is typically applied to low-dimensional problems, several works have applied it to NAS~\cite{DBLP:conf/icml/BergstraYC13,DBLP:conf/ijcai/DomhanSH15} and it has even surpassed human experts on competition datasets~\cite{DBLP:conf/icml/MendozaKFSH16}.
However, BO has mostly been overshadowed by RL ever since Zoph and Le achieved unprecedented results on NAS benchmarks~\cite{DBLP:conf/iclr/ZophL17}.
The RL problem can be formulated with the evolutionary search space as the agent's action space and the test set error as the reward~\cite{DBLP:conf/iclr/ZophL17,DBLP:conf/cvpr/ZophVSL18}.
Alternatively, the RL problem can be posed as a sequential control task~\cite{DBLP:conf/aaai/CaiCZYW18,DBLP:journals/corr/WeiWC17}: given the state of the architecture, what network modification
should be applied to improve performance?

While RL-based NAS has achieved state-of-the-art across many benchmarks, it tends to be compute-inefficient and can take thousands of GPU hours to converge~\cite{DBLP:conf/icml/RealMSSSTLK17,DBLP:conf/cvpr/ZophVSL18}.
\textit{Neuro-evolutionary} approaches are generally lightweight in comparison, and they notably perform the same as RL approaches on NAS benchmarks~\cite{DBLP:conf/aaai/RealAHL19}.
The use of evolutionary algorithms for NAS can be traced back decades, e.g.\ \cite{DBLP:conf/icga/MillerTH89} uses genetic algorithms to propose architectures that are then trained using backpropagation. While evolutionary algorithms have been used to search for both weights and network architectures~\cite{DBLP:journals/tnn/AngelineSP94,stanley:ec02}, it is more common to only apply evolution to the architecture and to train the weights with gradient descent~\cite{DBLP:conf/aaai/RealAHL19,DBLP:conf/icml/RealMSSSTLK17,DBLP:conf/iclr/ElskenMH19}.
Evolutionary algorithms evolve a population of candidate solutions to an optimization problem and each generation is derived from the last by applying mating operations to a set of selected parents. In NAS, an offspring may differ from its parents by an added layer, a changed connection, etc. The quality of solutions is judged by a fitness function and evolution is terminated when a resource or time budget is exceeded.

\paragraph{Multi-Objective Optimization \& NAS}

In a multi-objective optimization problem, there are $m$ objectives $\{f_1,\dots,f_m\}$, which in the context of NAS may be accuracy, floating point operations~(FLOPs), energy, etc. When $m > 1$, it becomes nontrivial to select the optimal solution among the set of all objective vectors $Y = \{\mathbf{y} \in \mathbb{R}^m \mid \mathbf{y}=\{f_1(\mathbf{x}),\dots,f_m(\mathbf{x})\}\}$ where $\mathbf{x}$ is a candidate solution.
There exists a variety of strategies to select solutions, such as optimizing for a weighted sum of the (normalized) objectives, lexicographic sorting, or maintaining Pareto-optimal solutions~\cite{Marler2004}.
We are most interested in the latter approach since it captures the trade-offs between objectives and allows the practitioner to choose the optimal compromise for their use case.
The set of Pareto-optimal solutions, also called the Pareto frontier or Pareto front, is the set of non-dominated solutions $\{\mathbf{y}^{\prime} \in Y \mid \{\mathbf{y} \in Y \mid \mathbf{y} \succ \mathbf{y}^{\prime}\} = \emptyset \}$, where $\mathbf{a} \succ \mathbf{b}$ indicates that $\mathbf{a}$ strictly dominates $\mathbf{b}$, i.e.\ $|\{f_i(\mathbf{a}) \mid 1 \le i \le m, f_i(\mathbf{a}) > f_i(\mathbf{b})\}| = m$.

The non-dominated sorting genetic algorithm-II~(NSGA-II)~\cite{nsga2} is an elitist evolutionary approach to multi-objective optimization.
Notably, the authors improve the non-dominated sorting algorithm from cubic to quadratic time complexity.
The surviving members of a generation are selected in a binary tournament with preference given to members of the Pareto front. Additional offspring are generated from members in the ranked fronts, i.e.\ the Pareto fronts computed iteratively after removing the members of the previous front. When a ranked front needs to be subsampled, the crowding distance within the front is used to ensure the full front is represented.

Related to our work,
NSGA-Net~\cite{nsga-net} is an evolutionary framework for NAS that employs NSGA-II for multi-objective optimization. Like most evolutionary NAS algorithms, NSGA-Net explores and exploits the search space with a fixed-size population of candidate architectures. These architectures, encoded as a sequence of phases, are initialized either randomly or seeded from hand-crafted architectures like ResNet.
In the \textit{exploration} stage, homogeneous crossover and bit-flipping mutation operators are applied to the population to create new offspring. 
In the \textit{exploitation} stage, Bayesian optimization is used to exploit correlations in architecture blocks over previous trials.
NSGA-Net is evaluated on CIFAR-10 and CIFAR-100 with a search space similar to DARTS~\cite{darts} using two metrics: classification error and computational complexity (FLOPs).
The authors demonstrate the effectiveness of population-based NAS and the superiority of NSGA-II over a weighted sum of objectives.
While similar to our framework, our focus is on the design of objectives conducive to interpretability. Furthermore, we scale our method to a distributed cluster using \texttt{Ray}~\cite{ray} and evaluate on more datasets.

Multi-objective optimization of a weighted sum of objectives has been employed by many NAS works.
The approach is attractive when the objectives are differentiable since it is amenable to gradient descent by backpropagation.
For instance, Multi-Objective NAS~(MONAS)~\cite{monas} uses RL with a weighted combination of accuracy, power, and multiply-accumulate operations (MACs) as the reward.
However, there are limitations to optimizing for an aggregate of multiple objectives: it relies on manually tuned coefficients, struggles to accommodate objectives that range over multiple orders of magnitude, and tends to cluster in a small region of the Pareto front.

\paragraph{Explainable AI (XAI) \& NAS}

Some of the intersection between interpretability and NAS has been covered in prior work.
In~\cite{DBLP:conf/iclr/RuW0O21}, a NAS framework using the Bayesian optimization search strategy is proposed. For efficiency and interpretability, a Weisfeiler-Lehman graph kernel is used to define a Gaussian process surrogate on the search space, and the gradients are used to identify key motifs that lead to well-performing architectures.
Similarly, \cite{DBLP:journals/corr/abs-1904-00438,DBLP:journals/corr/abs-2009-13266} use alternative techniques to identify key motifs used in the search process.
However, their notions of interpretability and disentanglement focus on the search process rather than on the learned models themselves. In this work, we extend NAS to disentangle the latent space of learned models.

\section{Proposed Framework: \XNAS{}}

Following the taxonomy in~\cite{elsken2019neural}, we break up our method into a search space, search strategy, and performance estimation strategy. We further discuss how we scale the search up to an arbitrary number of compute nodes.

\subsection{Search Space}\label{sec:search_space}

We are interested in exploring complex search spaces beyond simple chains, i.e.\ multi-branch networks such as ResNet~\cite{resnet} or DenseNet~\cite{densenet}.
To this end, we elect to use the popular NAS-Bench-201 search space~\cite{nasbench201}, which is comprised of a macro skeleton and a searched cell. An overview is shown in 
the supplemental material.

The first layer of the macro skeleton is a $3\times 3$ convolutional layer with $F=16$ filters followed by a batch normalization layer. This is followed by a stack of five searched cells ($F=16$).
A basic residual block ($F=32$) with a stride of two proceeds the stacked cell block. The shortcut connection is a $2\times 2$ 2D average pooling layer followed by a $1\times 1$ convolutional layer. These blocks are alternated,
cutting the image dimensions in half and doubling the filters for each set of blocks. The end of the network is a 2D global average pooling layer followed by a fully-connected (dense) classification layer with a softmax activation.

The searched cell can be expressed as a directed acyclic graph where nodes represent data and edges represent operations. The set of operations consists of $3\times 3$ convolutional blocks, $1 \times 1$ convolutional blocks, $3\times 3$ average pooling, ``zeroize'' (equivalent to dropping the edge), and ``skip-connect'' (equivalent to the identity operator). Note that each convolutional block is comprised of convolution, a rectified linear~(ReLU) activation, and batch normalization. All of the convolutions and pooling layers use \texttt{SAME} padding. To prevent cycles, each node is assigned a rank and can only connect to higher-rank nodes.
Since there are $V = 4$ nodes in a cell and five operation candidates in the operation set, the total size of the search space is 
$5 ^ {(\sum_{i=0}^{V-1} i)} = 15,625$ architectures.
There are two issues with the search space definition, which the NAS-Bench-201 authors also point out. First, different architecture encodings can result in the same graph. Like the authors, we do not consider isomorphism in the evaluation of architectures\footnote{The NAS-Bench-201 authors remark that there are 6,466 architectures with unique topology in the search space due to isomorphisms brought about by the ``skip-connect'' and ``zeroize'' operations.}.
Second, architectures can be disconnected due to the zeroize operation. In this case, the mating operations are reapplied to produce valid offspring.

We represent an architecture in the search space as $\mathbf{x}_i$, a fixed-size list of integers of size $\sum_{i=0}^{V-1} i = 6$ with each element in the range $[1..V]$.
Each element of this \textit{encoding} represents (i) a specific operator or operators, such as a convolutional or max pooling layer with specific parameters (e.g. kernel size, strides, etc.), or the lack of an operator (identity) and (ii) how that operator is connected to additional operators in the computational graph.

\subsection{Search Strategy}

As we are interested in discovering neural architectures that are both accurate and interpretable, we propose to use multi-objective optimization. 
We explore and exploit the search space using the Non-Dominated Sorting Genetic Algorithm II~(NSGA-II), as introduced in Section~\ref{sec:background}, with two objectives: accuracy and \textit{introspectability} (introduced in Section~\ref{sec:perf_eval}).
Because we search for architectures that are both accurate and interpretable, we refer to our approach as e\underline{X}plainable \underline{NAS}~(\XNAS{}).
We generate the initial set of solutions by uniformly sampling each of the 6 variables in the optimization problem (as defined in Section~\ref{sec:search_space}). These candidates comprise the first generation of the \textit{population}. Thereafter, the offspring of the proceeding generation are produced by mating the parents comprising the prior generation.
Parents are selected based on the ranked Pareto fronts of the population as described in Section~\ref{sec:perf_eval} and~\cite{nsga2}. Because of this selection, there is no notion of a single best solution, but rather a set of non-dominated solutions that characterize the optimal trade-off between all objectives.

Mating comprises two core operations: crossover and mutation.
The \textit{crossover} operator produces offspring by combining the encodings of two parents. The operator combines the building blocks between successful parents to exploit the \textit{implicit parallelism} of population-based search~\cite{10.7551/mitpress/1090.001.0001}.
Due to the integer-based encoding that we employ in this work, we elect to use simulated binary crossover~\cite{binary-crossover}, which uses a probability density function to simulate the single-point crossover of binary-coded genetic algorithms.
The \textit{mutation} operator produces offspring by modulating one or more of the variables of a single parent. We specifically select polynomial mutation, which follows the same probability distribution as simulated binary crossover.
Both crossover and mutation also have a parameter $p$ that controls the probability that the respective operator is applied to a member of the population.

\subsection{Performance Evaluation Strategy}\label{sec:perf_eval}

We evaluate the performance of an architecture using two objectives: task performance and interpretability.
The former is simple to define quantitatively as the classification accuracy on the held-out validation split of a dataset.
However, interpretability is often treated far more qualitatively and an objective definition eludes community consensus. Furthermore, explaining a model is dependent on the audience, data modality, modeling task, and questions being asked. To disambiguate interpretability in the context of the framework, we state our assumptions: that the user has some technical understanding (e.g.\ a data scientist or domain expert), that we are interested in understanding the model in classification tasks (e.g.\ as opposed to the data or the NAS evolution process), and that models that maximize the metric lead to qualitatively discernible trends.
To this end, we propose to quantify the interpretability of models as the introspectability of disentangled elements, which we describe in the subsequent subsections. We measure this for supervised classification tasks using the pairwise distances between latent representations of individual classes.

\paragraph{Introspectability}

Here we formalize the score that we denote as \textit{introspectability}: the degree to which the representations of disparate classes within a neural network $\mathcal{M}$ are disentangled.
Let us denote the subset of validation data belonging to class $c$ as $\tensor{X}^{(c)} \in \mathbb{R}^{N^{(c)} \times H \times W \times C}$.
Given $\tensor{X}^{(c)}$ as input to $\mathcal{M}$, denote the activations of layer $l$ as $\mathbf{\Phi}^{(c,l)} \in \mathbb{R}^{N^{(c)} \times d^{(l)}_1 \times \dots \times d^{(l)}_n}$.
We reshape the activations to have a single dimension of size
$d^{(l)} = \prod_{i=1}^n d^{(l)}_i$
such that $\mathbf{\Phi}^{(c,l)} \in \mathbb{R}^{N^{(c)} \times d^{(l)}}$.
We denote all activations for class $c$ within $\mathcal{M}$ as
$\mathbf{\Phi}^{(c)} = \big\Vert_{l=1}^L \mathbf{\Phi}^{(c,l)}$
where $\big\Vert$ is the matrix concatenation operator along the columns and $L$ is the number of layers in $\mathcal{M}$.
The mean activations for class $c$ are then
$\bar{\mathbf{\Phi}}^{(c)} = \frac{1}{N^{(c)}} \sum_{i=1}^{N^{(c)}} \mathbf{\Phi}_i^{(c)}$
where $|\bar{\mathbf{\Phi}}^{(c)}| = \sum_{l=1}^L d^{(l)}$.
With these definitions, we then formulate introspectability as~\eqref{eq:introspectability}
\begin{equation}\label{eq:introspectability}
\text{Introspectability}(\mathcal{M}, \tensor{X}) = \frac{1}{\binom{N_C}{2}} \sum_{c = 1}^{N_C} \sum_{k = c + 1}^{N_C} D(\bar{\mathbf{\Phi}}^{(c)}, \bar{\mathbf{\Phi}}^{(k)})
\end{equation}
\noindent
where $D(\cdot, \cdot)$ gives the cosine distance between its two vector arguments and $N_C$ is the number of classes in the classification task.

\paragraph{Introspectability and WordNet}

\begin{figure}
    \centering
    \begin{subfigure}[b]{\linewidth}%
        \centering
        \includegraphics[width=.7\linewidth]{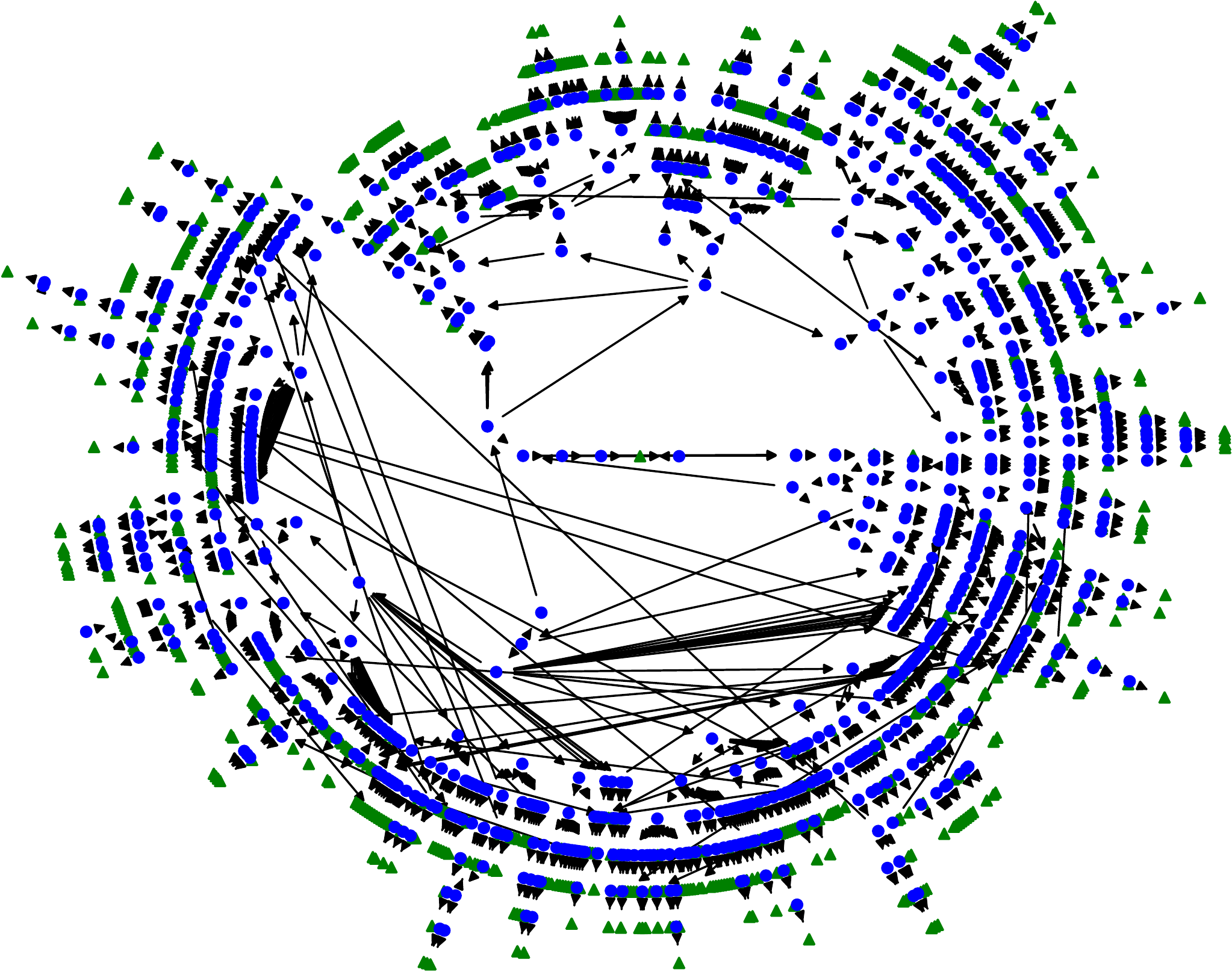}
        \caption{}
        \label{fig:wordnet_imagenet}
    \end{subfigure}%
    \\%
    \begin{subfigure}[b]{\linewidth}%
        \centering
        \includegraphics[width=.82\linewidth]{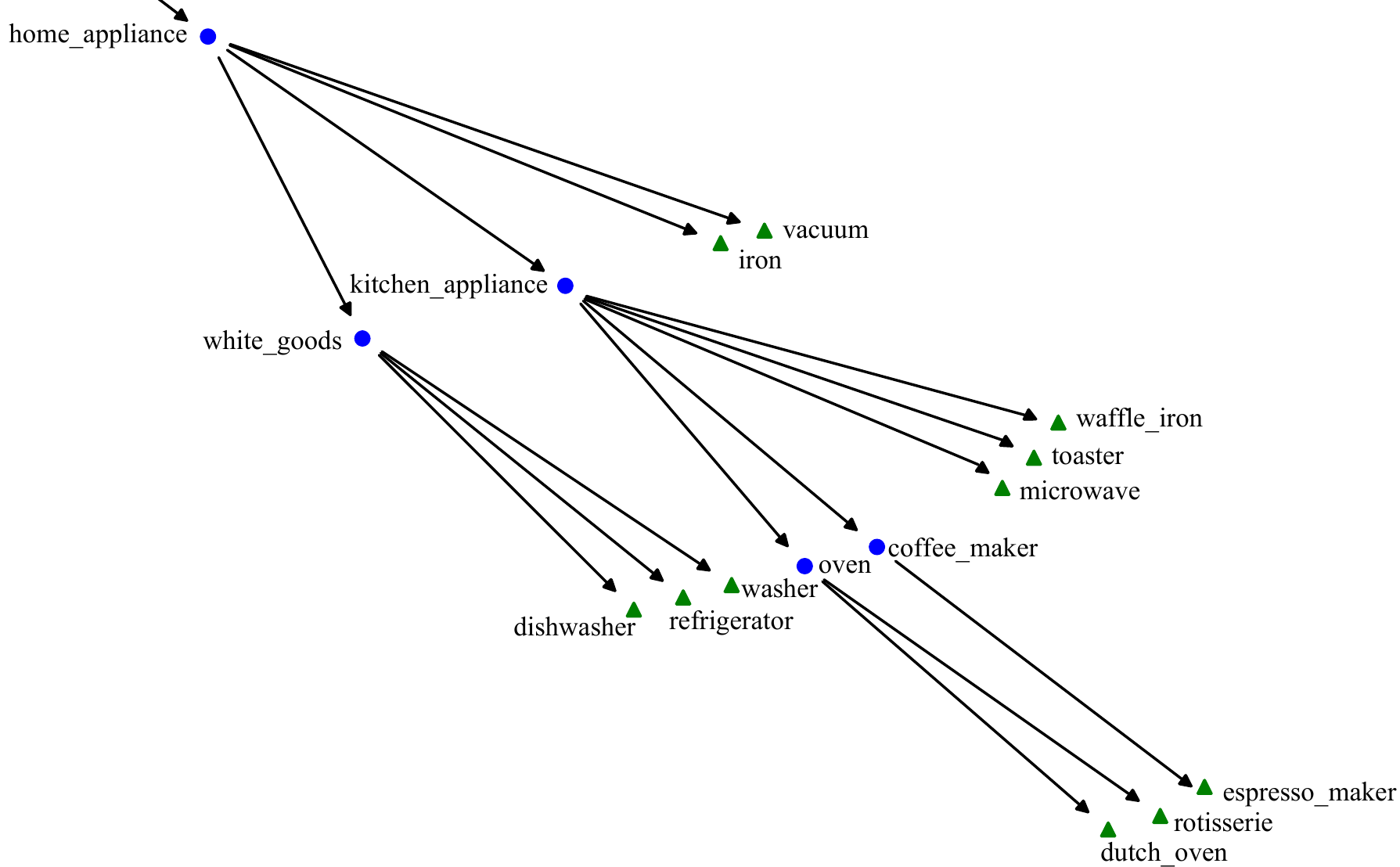}
        \caption{}
        \label{fig:wordnet_imagenet_inset_0}
    \end{subfigure}%
    \caption{(a) The WordNet hyponym-hypernym graph where leaf nodes (green) are the labels of the ImageNet dataset. (b) An inset of the full graph containing the synset ``home appliance'' and its hyponyms.}
    \label{fig:wordnet}
\end{figure}

\begin{table*}%
    \small
    \centering
    \ra{1.1}
    \addtolength{\tabcolsep}{-2.5pt}
    \begin{tabular}{@{}lccccccc@{}}
        \toprule
        \multirow{2}{*}{Dataset} & \multirow{2}{*}{\makecell{Multi-\\Objective}} & \multirow{2}{*}{Gen.} & \multicolumn{4}{c}{Population-Level Statistics} & \multirow{2}{*}{\makecell{Hypervolume}} \\
        & & & Maximum Accuracy & Median Accuracy & Maximum Intros. & Median Intros. & \\
        \midrule
        \multirow{4}{*}{MNIST}
        & reg & -- & 98.9\% & 98.6\% & 0.384 & 0.291 & 0.341 \\
        & \xmark & 18 & 99.1\% & 98.8\% & 0.353 & 0.255 & 0.314 \\
        & \cmark & 18 & 99.1\% & 98.6\% & 0.390 & 0.258 & 0.347 \\
        & \cmark+reg & 18 & 99.0\% & 98.6\% & 0.503 & 0.303 & \textbf{0.424} \\[1ex]
        \multirow{4}{*}{CIFAR-10}
        & reg & -- & 85.9\% & 72.8\% & 0.331 & 0.178 & 0.229 \\
        & \xmark & 34 & 87.7\% & 84.3\% & 0.328 & 0.077 & 0.237 \\
        & \cmark & 34 & 87.9\% & 74.6\% & 0.552 & 0.196 & 0.293 \\
        & \cmark+reg & 34 & 87.7\% & 74.2\% & 0.654 & 0.249 & \textbf{0.361} \\[1ex]
        \multirow{4}{*}{ImageNet-16-120}
        & reg & -- & 44.8\% & 36.2\% & 0.318 & 0.087 & 0.099 \\
        & \xmark & 11 & 47.8\% & 44.9\% & 0.301 & 0.053 & 0.099 \\
        & \cmark & 11 & 47.3\% & 39.1\% & 0.317 & 0.104 & 0.111 \\
        & \cmark+reg & 11 & 47.4\% & 39.0\% & 0.380 & 0.109 & \textbf{0.117} \\
        \bottomrule
    \end{tabular}
    \caption{\XNAS{} experimental results on image classification datasets listing the population-level accuracy and introspectability (Intros.) scores, and hypervolume. The regularization term is denoted as ``reg'' and is described in Section~\ref{sec:results}. Normalized $\text{Introspectability}_{\text{WordNet}}$ is shown for ImageNet-16-120. Multi-objective \XNAS{} with regularization yields the greatest hypervolume across all tasks.}
    \label{tab:results}
    \addtolength{\tabcolsep}{2.5pt}
\end{table*}

We derive a second definition of introspectability based on WordNet~\cite{miller1995wordnet}, a lexical database of the English language. WordNet comprises sets of synonyms (\textit{synsets}) and arises into a hierarchical representation by embedding the transitive relations \textit{hyponyms} (more specific sub-names) and \textit{hypernyms} (more abstract super-names).
In computer vision, the labels of the ImageNet database~\cite{imagenet} are notably derived from WordNet synsets. We visualize all of the labels covered by ImageNet in the hyponym-hypernym graph shown in \figref{fig:wordnet_imagenet} and \ref{fig:wordnet_imagenet_inset_0}.
The shortest path distances between two labels in the hyponym-hypernym graph can be used to compute semantic similarity as shown in \eqref{eq:path_sim}
\begin{equation}\label{eq:path_sim}
    \text{\small\texttt{path\_sim}($w_a, w_b$)} = \frac{1}{\text{\small\texttt{shortest\_path}($w_a, w_b$)} + 1}
\end{equation}
\noindent
where $w_a$ and $w_b$ are label names. We then weigh the pairwise distances between classes by this similarity as \eqref{eq:introspection_imagenet}
\begin{align}
\begin{split}\label{eq:introspection_imagenet}
    &\text{Introspectability}_{\text{WordNet}}(\mathcal{M}, \tensor{X}) = \\
    &\quad\frac{1}{{\binom{N_C}{2}}}\sum_{c = 1}^{N_C} \sum_{k = c + 1}^{N_C}  D(\bar{\mathbf{\Phi}}^{(c)}, \bar{\mathbf{\Phi}}^{(k)}) \times 
    S(c, k).
\end{split}
\\
\intertext{The similarity between labels $S$ is given by \eqref{eq:introspection_imagenet_b}}
\label{eq:introspection_imagenet_b}
& S(c, k) = \text{\small\texttt{path\_sim}(\texttt{name}($c), \text{\texttt{name}(}k$))}
\end{align}
\noindent
where $\texttt{name}(\cdot)$ maps the label index to the corresponding label name.
Intuitively, the score penalizes models with relatively small distances between dissimilar labels and compensates for small distances between similar labels.
To ensure the range of WordNet introspectability is comparable to that of the baseline definition, we normalize the score by dividing by the mean \texttt{path\_sim} value among all pairs of labels in the dataset.

\subsection{Scaling Up}

We scale \XNAS{} to clusters comprising an arbitrary number of compute nodes using the distributed framework, \texttt{Ray}~\cite{ray}.
Given a set of $M$ nodes $\{n_i\}_{i=1}^M$, $n_1$ is treated as a head node that is responsible for running the core NSGA-II optimization loop and the core \texttt{Ray} server. The remaining nodes $\{n_i\}_{i=2}^M$ are configured as workers available to train and evaluate architectures on a dataset. When a new generation of architectures is created, each offspring is submitted for fitness evaluation to a queue by the head node. Each job in the queue is offloaded to a free worker until all workers complete their jobs and the queue is empty. The head node $n_1$ also is treated as an additional worker if it has free resources. 
Each worker node can execute in parallel as many jobs as it has GPUs.

\begin{figure*}
    \centering
    \begin{subfigure}[b]{0.29\linewidth}%
        \centering
        \includegraphics[width=\linewidth]{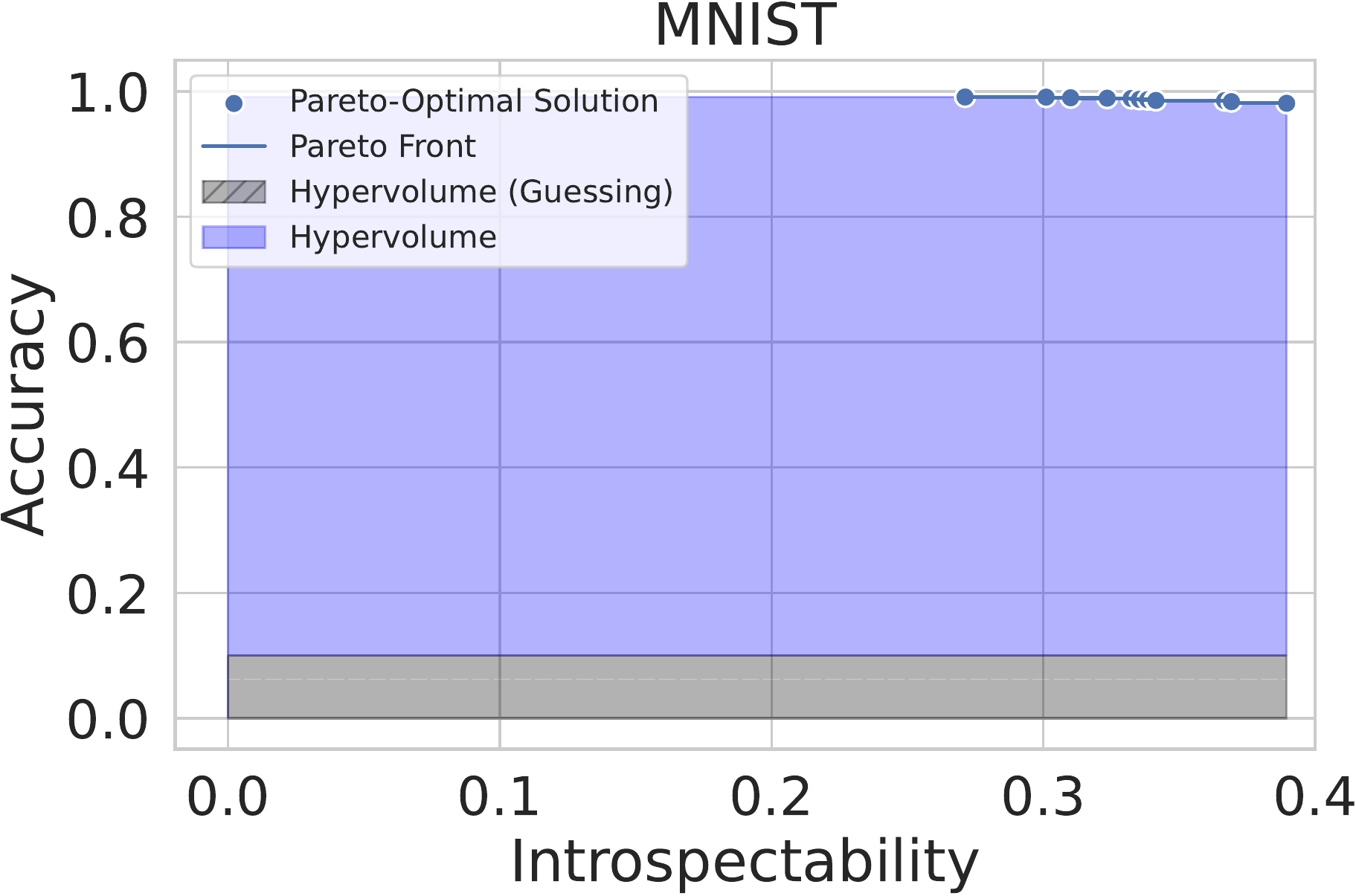}
    \end{subfigure}%
    \hfill
    \begin{subfigure}[b]{0.29\linewidth}%
        \centering
        \includegraphics[width=\linewidth]{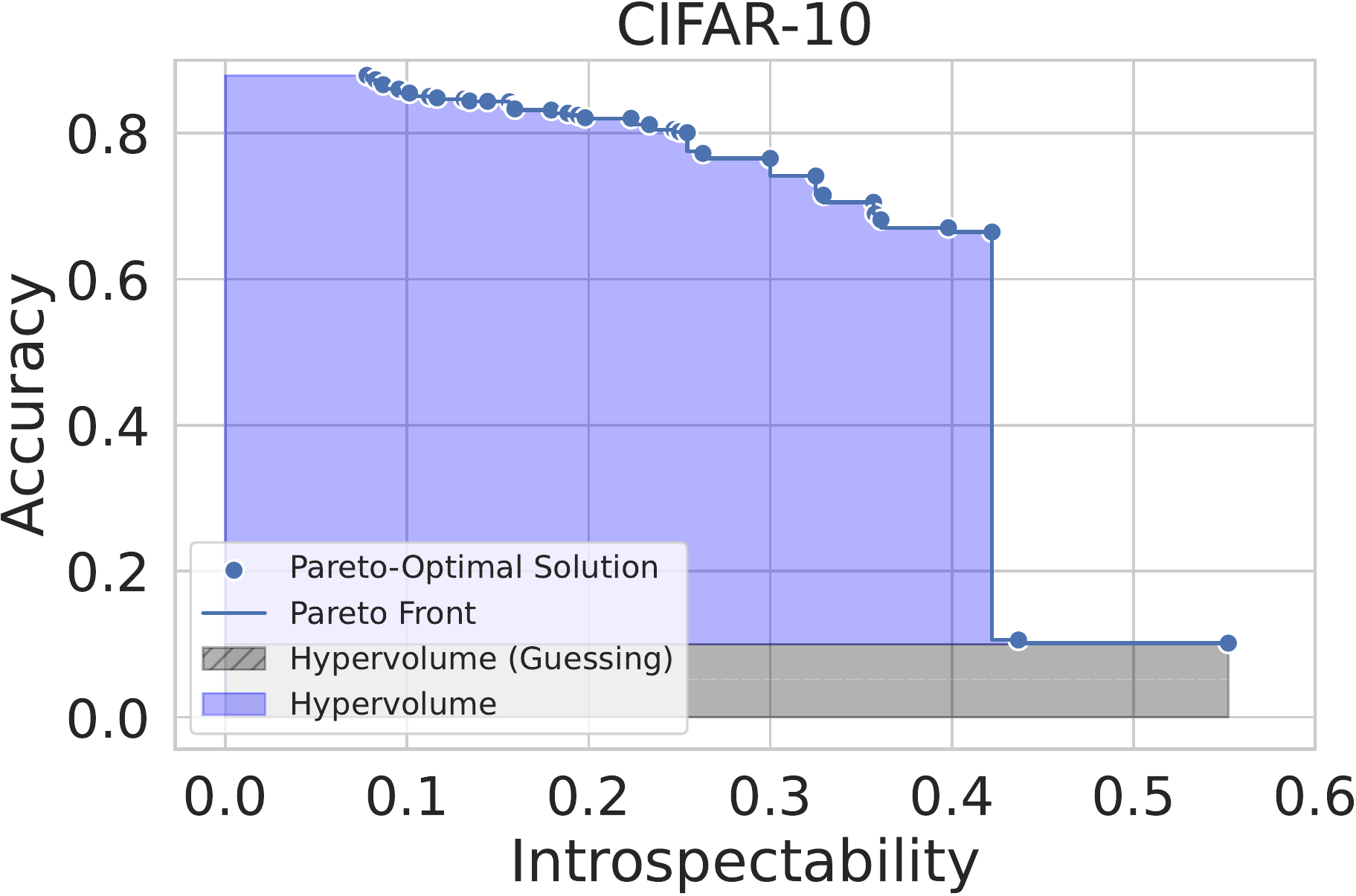}
    \end{subfigure}%
    \hfill
    \begin{subfigure}[b]{0.29\linewidth}%
        \centering
        \includegraphics[width=\linewidth]{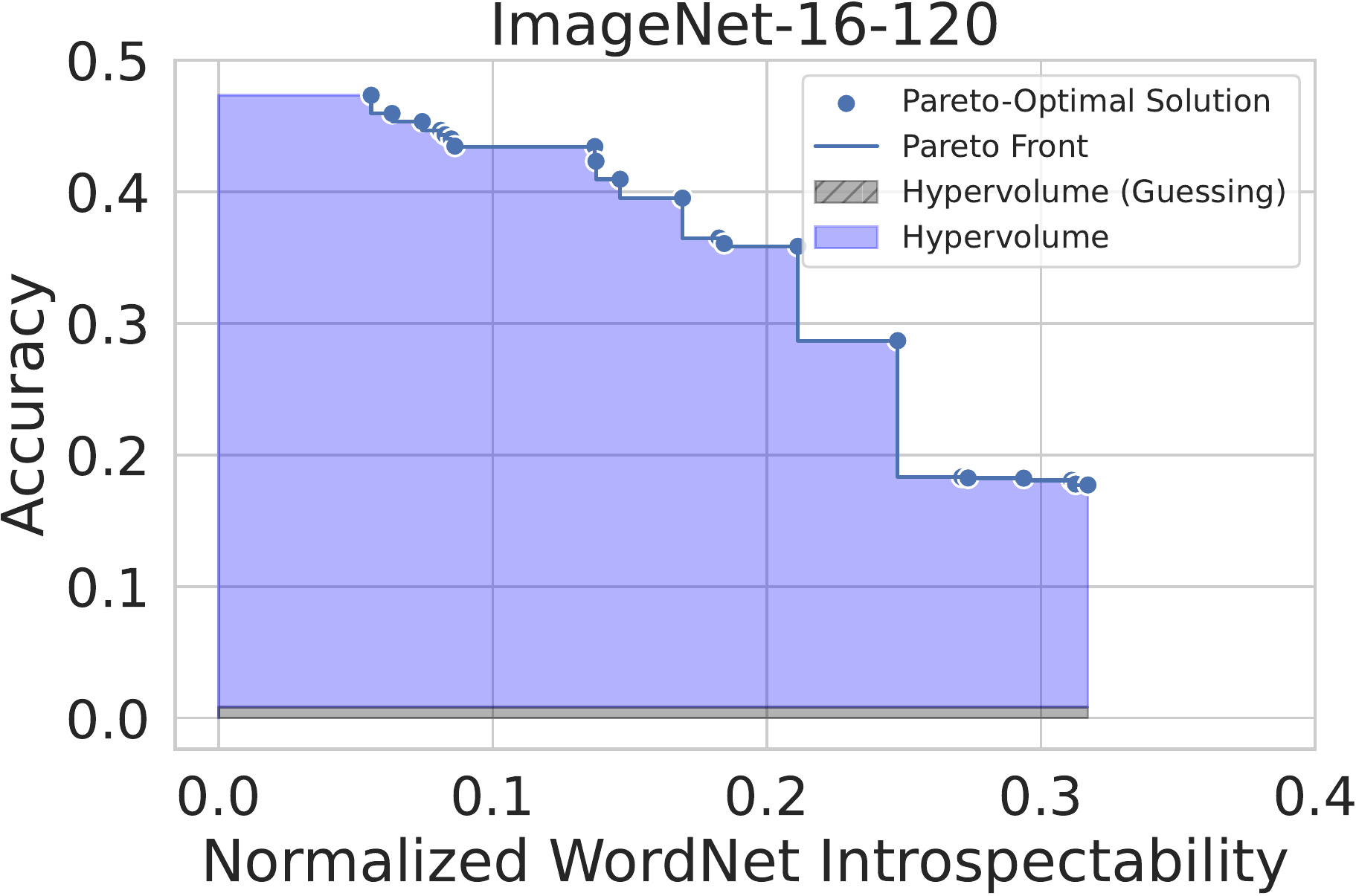}
    \end{subfigure}\\%
    \begin{subfigure}[b]{0.29\linewidth}%
        \centering
        \includegraphics[width=\linewidth]{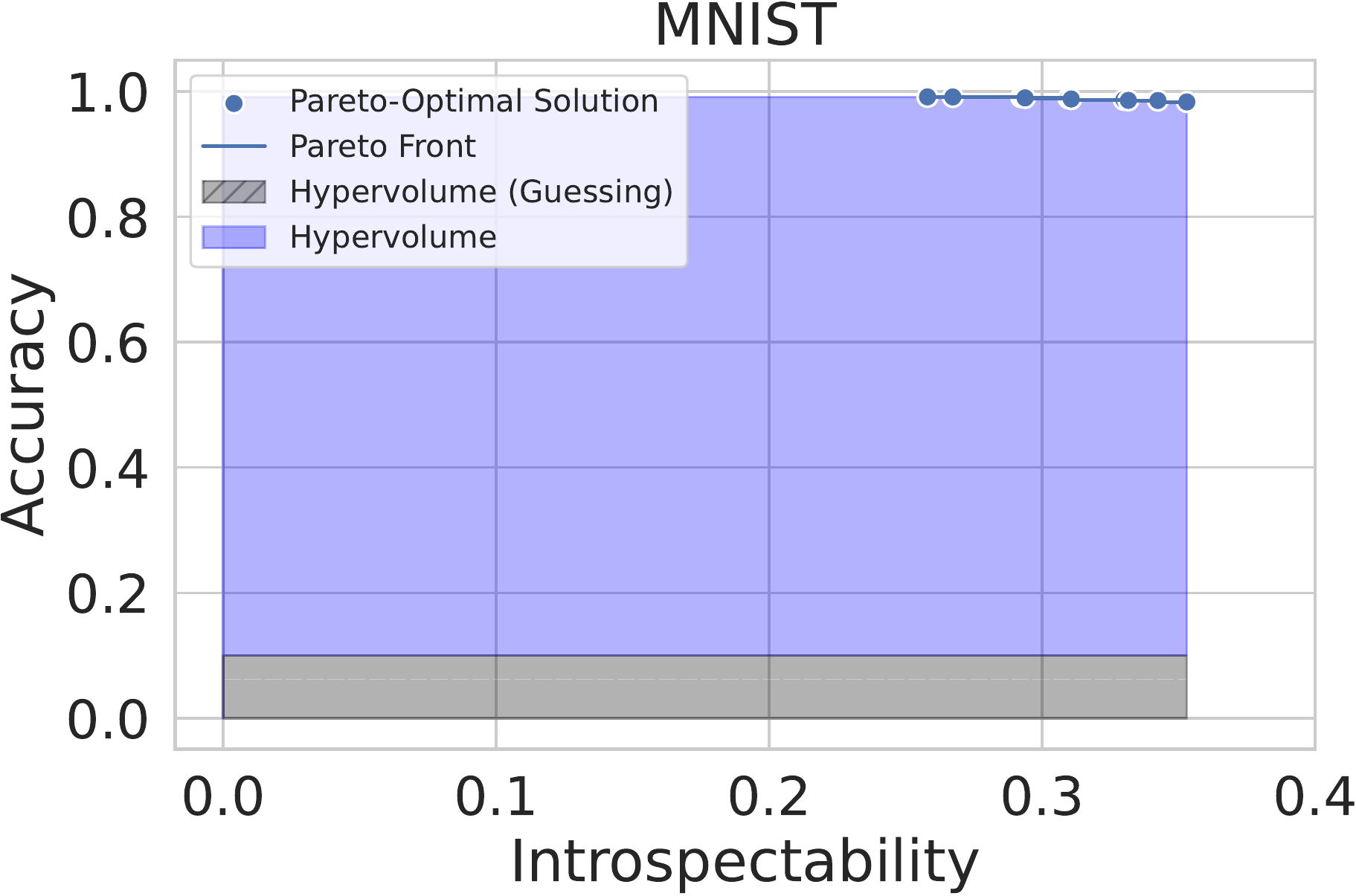}
    \end{subfigure}%
    \hfill
    \begin{subfigure}[b]{0.29\linewidth}%
        \centering
        \includegraphics[width=\linewidth]{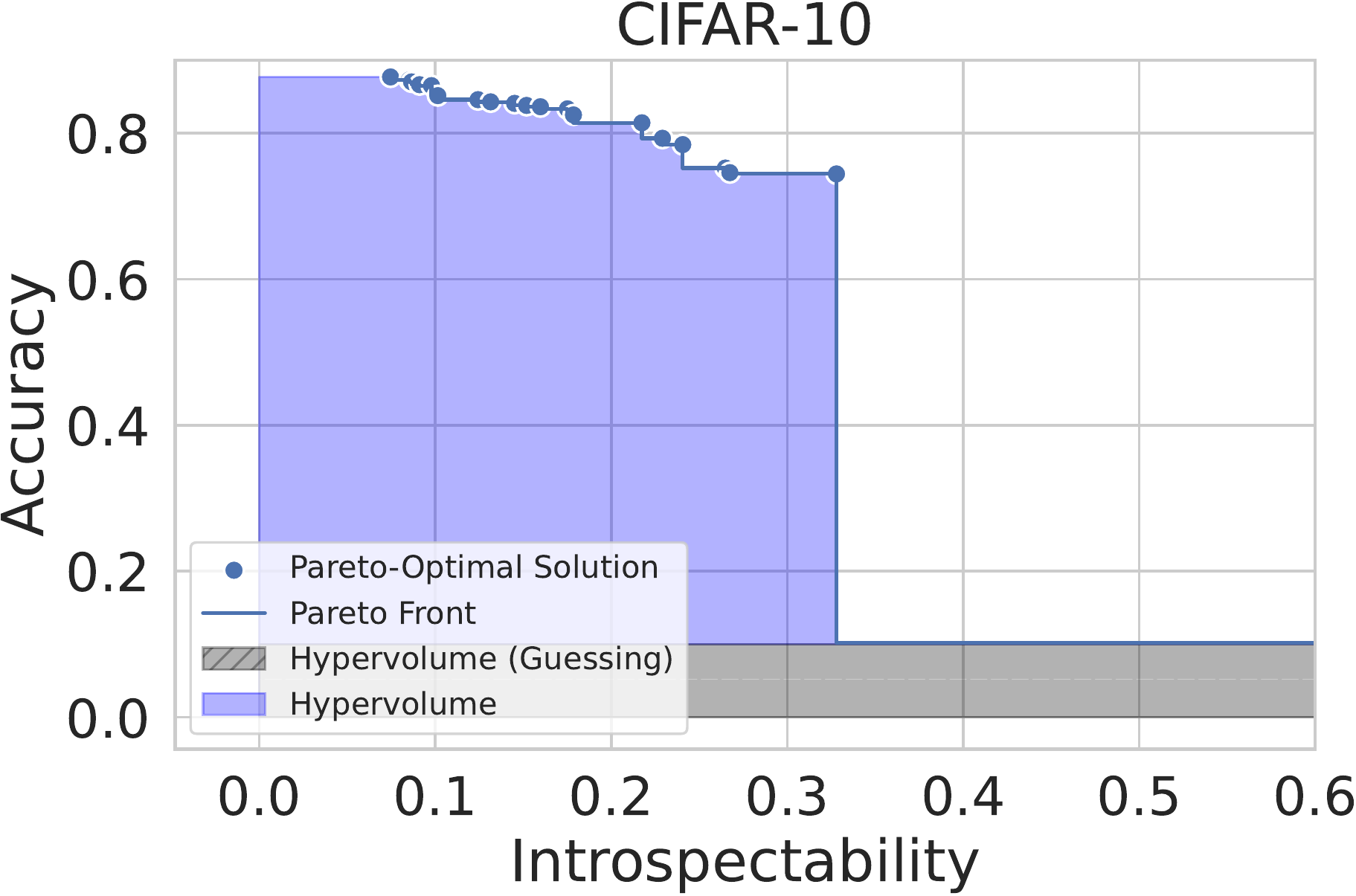}
    \end{subfigure}%
    \hfill
    \begin{subfigure}[b]{0.29\linewidth}%
        \centering
        \includegraphics[width=\linewidth]{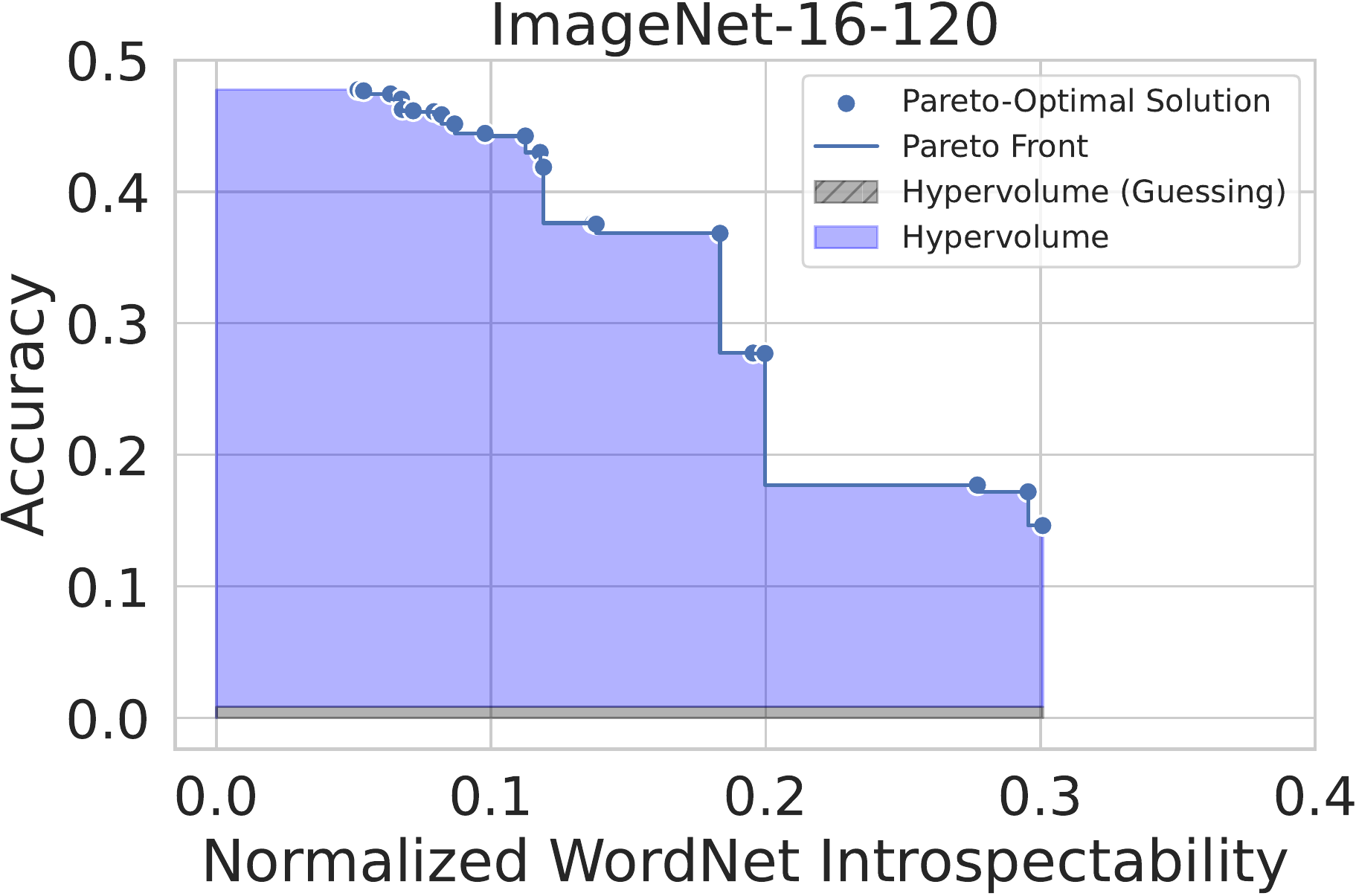}
    \end{subfigure}
    \caption{The Pareto fronts and Pareto-optimal solutions are shown for each task. The shaded region visualizes the hypervolume achieved by \XNAS{}. In the first row, the front is visualized for multi-objective \XNAS{} and for single-objective (accuracy only) in the second row. Normalized $\text{Introspectability}_{\text{WordNet}}$ is shown for ImageNet-16-120.}
    \label{fig:pareto_fronts}
\end{figure*}

\section{Experiments \& Results}

We evaluate \XNAS{} on three image classification datasets: MNIST, CIFAR-10, and ImageNet-16-120~\cite{nasbench201}.
Thereafter, we conduct analyses to understand the evolution process of \XNAS{}, characterize the Pareto front of each task, and 
demonstrate the debuggability of higher-introspectability architectures.

\subsection{Setup and Implementation}
The experimental setup and all hyperparameters are detailed in Appendix~B.
We implement \XNAS{} in Python with code built on the \texttt{\small pymoo}~\cite{pymoo}, \texttt{\small DeepHyper}~\cite{deephyper}, and \texttt{\small Ray}~\cite{ray} libraries. The source code is publicly available%
\ifcsname itIsArxivTime\endcsname%
\footnote{\href{https://github.com/LLNL/XNAS}{github.com/LLNL/XNAS}}.
We performed experiments on the LC systems at Lawrence Livermore National Lab~(LLNL).
\else%
~at $<$redacted$>$.
\fi%

\subsection{Metrics}

To quantitatively assess the quality of the solutions from a multi-objective search algorithm, we look to \textit{hypervolume} as introduced in~\cite{DBLP:conf/ppsn/ZitzlerT98} and improved in~\cite{DBLP:conf/cec/FonsecaPL06}.
Hypervolume can be perceived as the area of the union of rectangles where each rectangle is formed by a point on the Pareto front and a reference point (such as $(0, 0)$).
This notion can easily be extended to higher dimensions, i.e.\ rectangular cuboids are formed with three objectives and hyperrectangles are formed with four or more objectives.
In this work, we set the reference point to $(0, 1/N_C)$ where $1/N_C$ is the classification rate of random guessing with balanced data. Note that this reduces the hypervolume range from $[0,1]$ to $[0, (1{-}1/N_C)]$. We are interested in setting the reference point here to avoid rewarding models that have not learned the task in any significant capacity.

\subsection{Results}\label{sec:results}

\begin{figure}%
    \centering
    \begin{subfigure}[b]{\linewidth}%
        \centering
        \includegraphics[width=.72\linewidth]{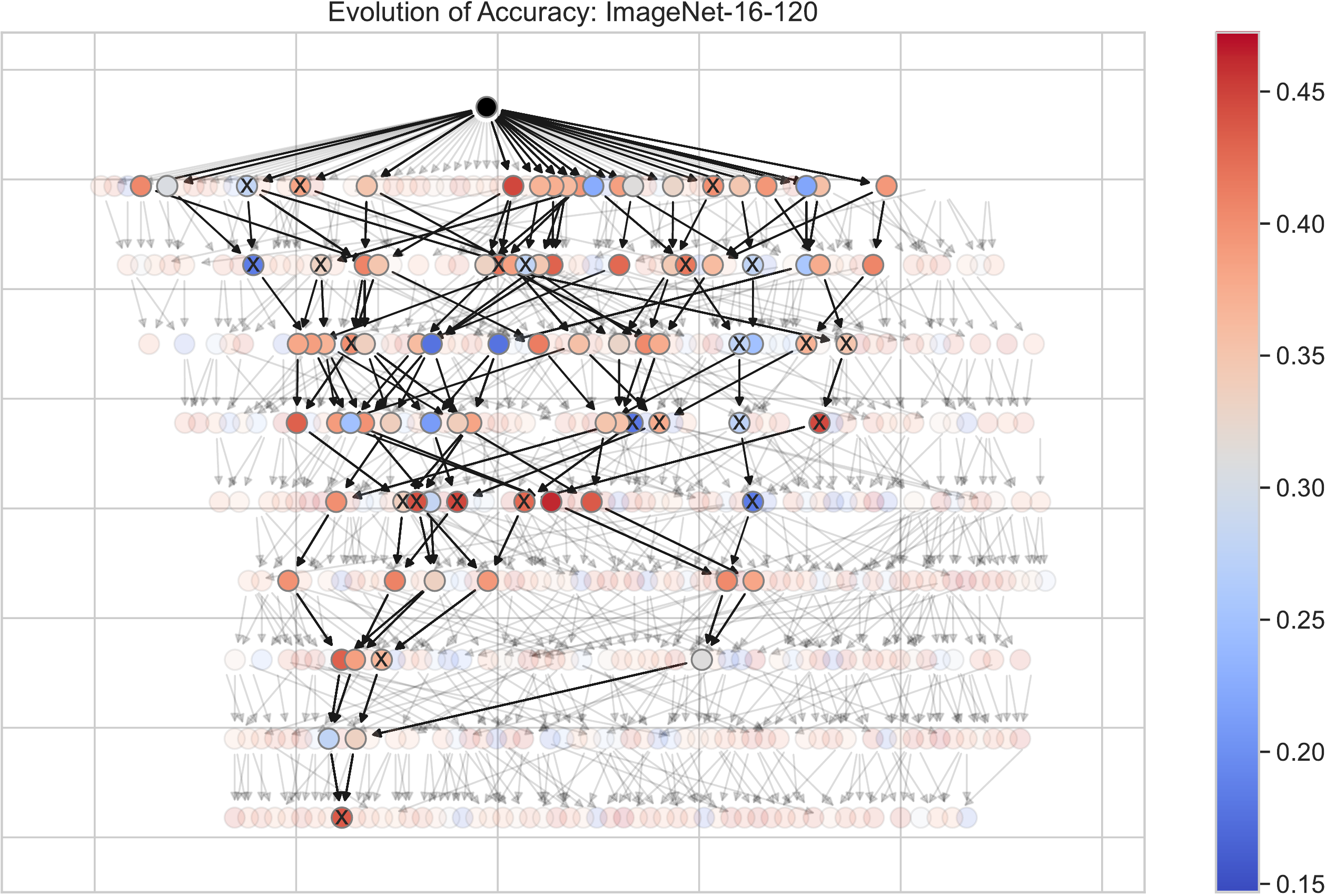}
    \end{subfigure}\\%
    \begin{subfigure}[b]{\linewidth}%
        \centering
        \includegraphics[width=.72\linewidth]{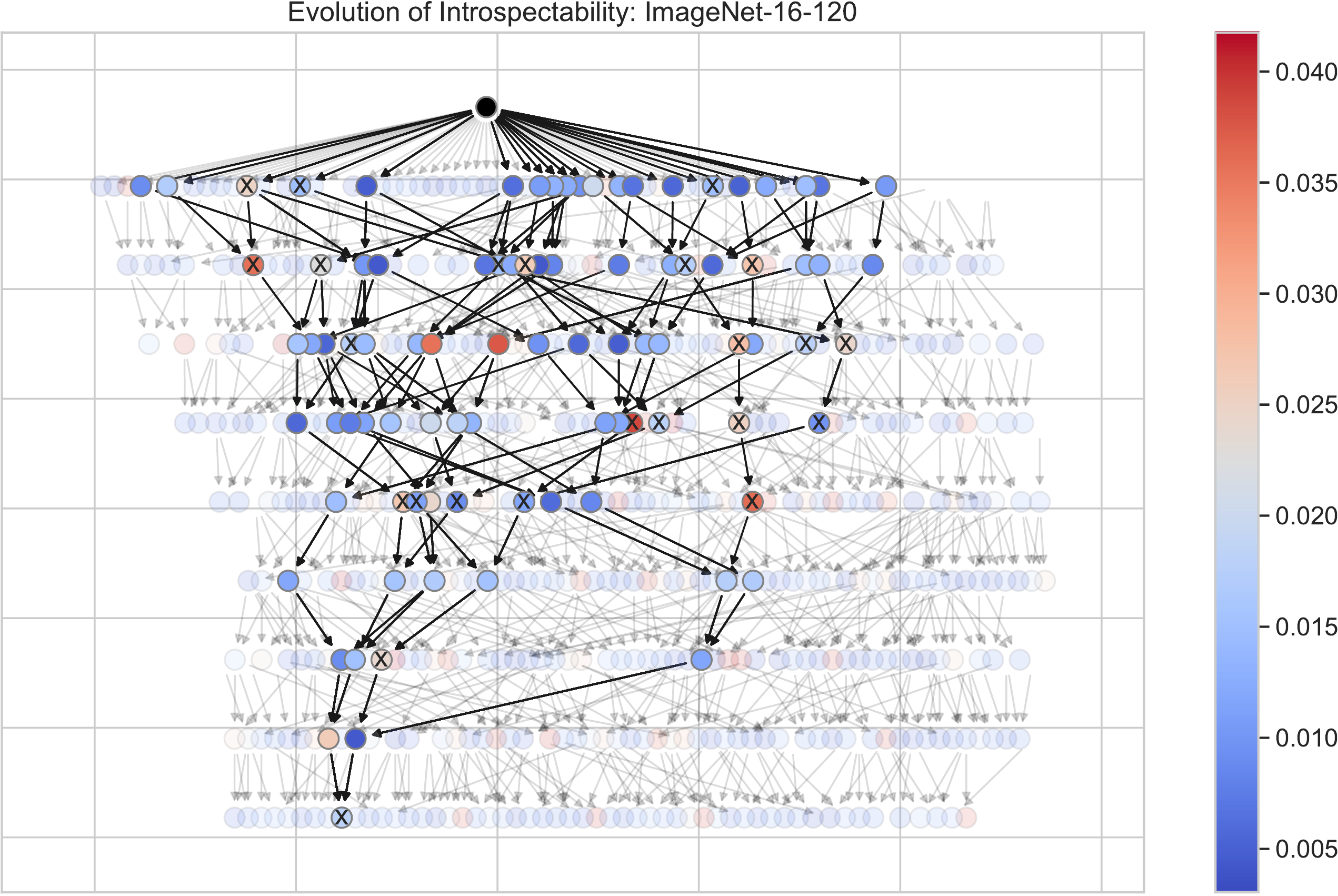}
    \end{subfigure}
    \vspace{-1.9em}
    \caption{Phylogenetic trees showing the evolution of the population in \XNAS{} on the ImageNet-16-120 task up to the eighth generation. The mating path from the initial population to a solution in the final Pareto front of the search is emphasized. Along the path, parents that were Pareto-optimal within their own generation are marked with an $\times$. The evolution of accuracy (top) and introspectability (bottom) within the population are shown. \vspace{-1em}}
    \label{fig:phylo_trees}
\end{figure}

Table~\ref{tab:results} contains the aggregate results for each task and demonstrates the efficacy of using the proposed multi-objective approach.
\XNAS{} is compared with the single-objective NAS baseline.
While the median population-level accuracy falls slightly compared to single-objective NAS, the maximum accuracy is still comparatively high and there is a substantial increase in hypervolume across experiments. Note that the best accuracy of \XNAS{} on ImageNet-16-120 is on par with the best-performing methods as evaluated in~\cite{nasbench201}.
We plot the Pareto front of every NAS search result and shade in the hypervolume in \figref{fig:pareto_fronts}. The visualizations make clear where the multi-objective approach makes up hypervolume over single-objective (accuracy). Across all tasks, multi-objective covers a larger range of introspectability values.
As one would expect, focusing on accuracy tends to cluster the majority of non-dominated solutions in the upper left of the front.
The hypervolume of random guessing illustrates why we set the hypervolume reference point to $(0, 1/N_C)$: some solutions manage to achieve high introspectability but are effectively useless since their predictions are no better than random.

As another baseline, introspectability is employed as a \textit{regularization} term.
Introspectability is differentiable and in turn, can be used to train models as an auxiliary loss. However, this drastically increases the training computational complexity and memory utilization due to the calculation of pairwise distances and the accumulation of activations. In experiments, this slows down training by several orders of magnitude. Accordingly, the number of search space evaluations is limited. Regularization achieves introspectability scores competitive with the multi-objective approach on MNIST and ImageNet-16-120, but not CIFAR-10. In addition, the achieved accuracy and hypervolume are hindered due to the reduced evaluations. Synergistically, we also evaluate regularization applied to the Pareto front of the multi-objective approach -- \XNAS{} is capable of discovering high-accuracy solutions that are predisposed to higher introspectability via the regularization approach. This combination performs best but demands the additional computation.

To understand the evolution process of \XNAS{}, consider the phylogenetic trees shown in \figref{fig:phylo_trees}. It shows the ancestry for a Pareto-optimal solution from the eighth generation with the ImageNet-16-120 task. While most solutions on a given generation's Pareto front are not directly descended from the previous generation's Pareto front, they typically have many Pareto-optimal ancestors. This suggests that Pareto optimality is mildly heritable, although not enough to ensure direct transmission between generations. In addition, the final Pareto-optimal solution had the second-highest cumulative accuracy at its generation. It is not surprising that its ancestors tended to have above-average accuracy and below-average introspectability.

\begin{figure}%
    \centering
    \begin{subfigure}[b]{.5\linewidth}%
        \centering
        \includegraphics[width=.98\linewidth]{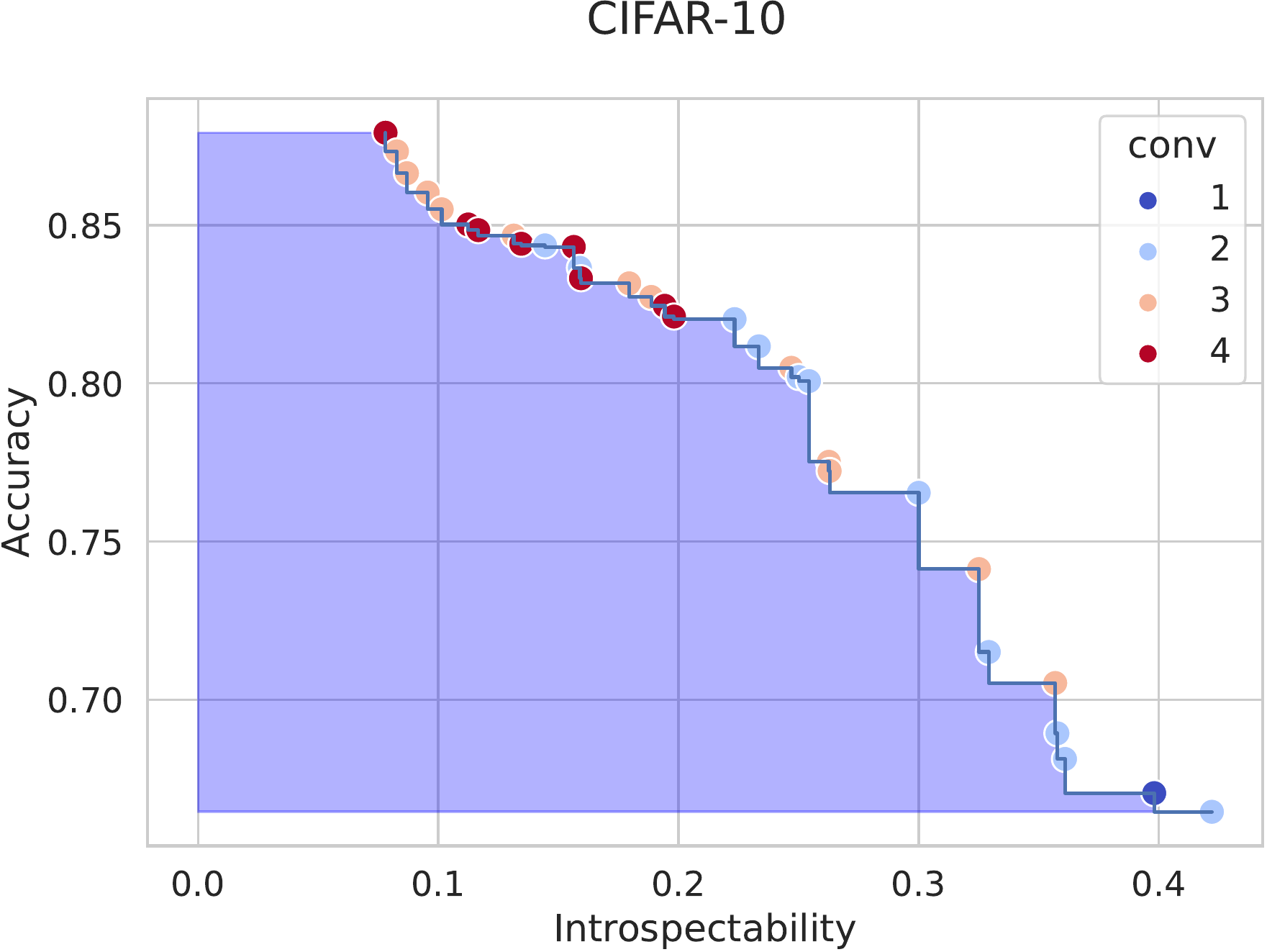}
    \end{subfigure}%
    \begin{subfigure}[b]{.5\linewidth}%
        \centering
        \includegraphics[width=.98\linewidth]{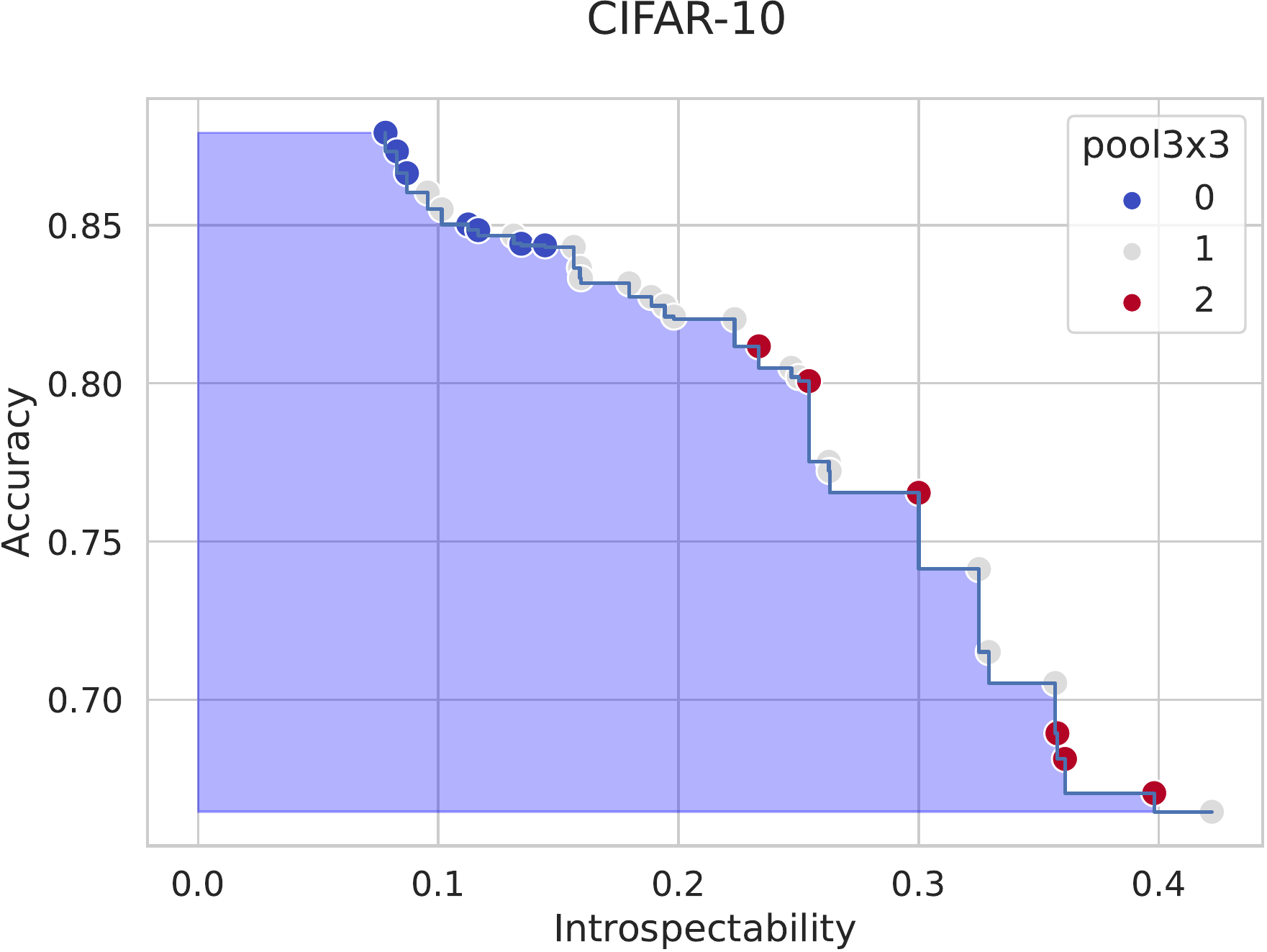}
    \end{subfigure}
    \caption{The Pareto front for the CIFAR-10 task with solutions colored by the number of convolutional (left) and pooling (right) layers per cell.}
    \label{fig:some_motifs}
\end{figure}

We conduct an analysis of emerging patterns in the architectures discovered across the Pareto front for each task.
The methodology for selecting motifs of interest is described in the supplemental material and is accompanied by the corresponding visualizations.
In \figref{fig:some_motifs}, we elect to visualize a pattern that holds for all tasks but is shown for CIFAR-10.
Therein, we observe that more accurate models have fewer pooling layers and more convolutional layers, whereas models with greater introspectability exhibit the opposite tendency.
These layer types can be seen as one knob that controls the accuracy-introspectability trade-off.
Furthermore, a study of the impact of accuracy and introspectability of the generalization error, convergence speed, and the number of parameters is presented in Appendix J. High-introspectability models have lower generalization error, fewer parameters, and faster training times, whereas high-accuracy models exhibit the inverse trend.

\begin{figure}%
    \centering
    \begin{subfigure}[b]{.75\linewidth}%
        \centering
        \includegraphics[width=\linewidth]{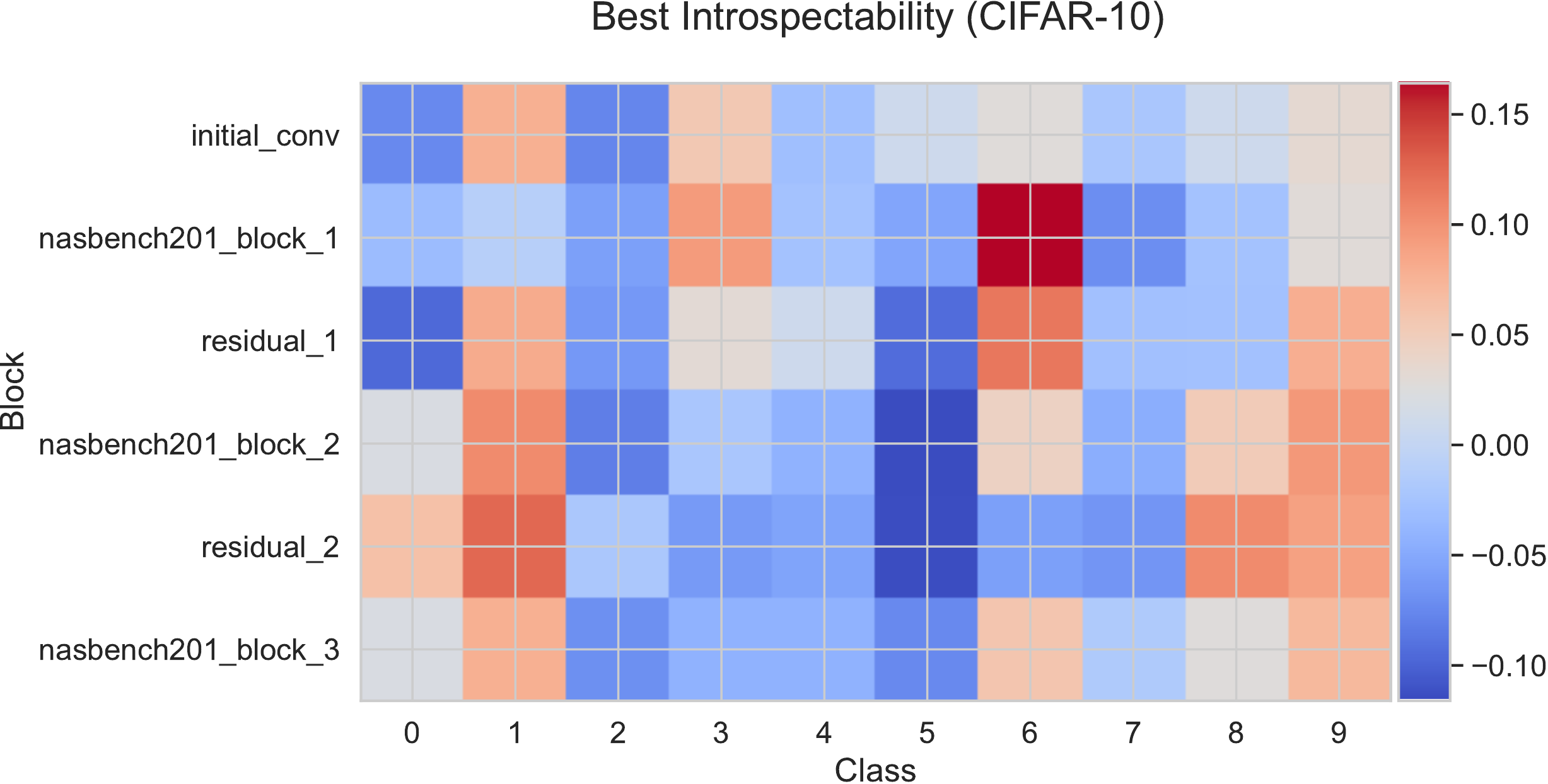}
    \end{subfigure}\\%
    \begin{subfigure}[b]{.75\linewidth}%
        \centering
        \includegraphics[width=\linewidth]{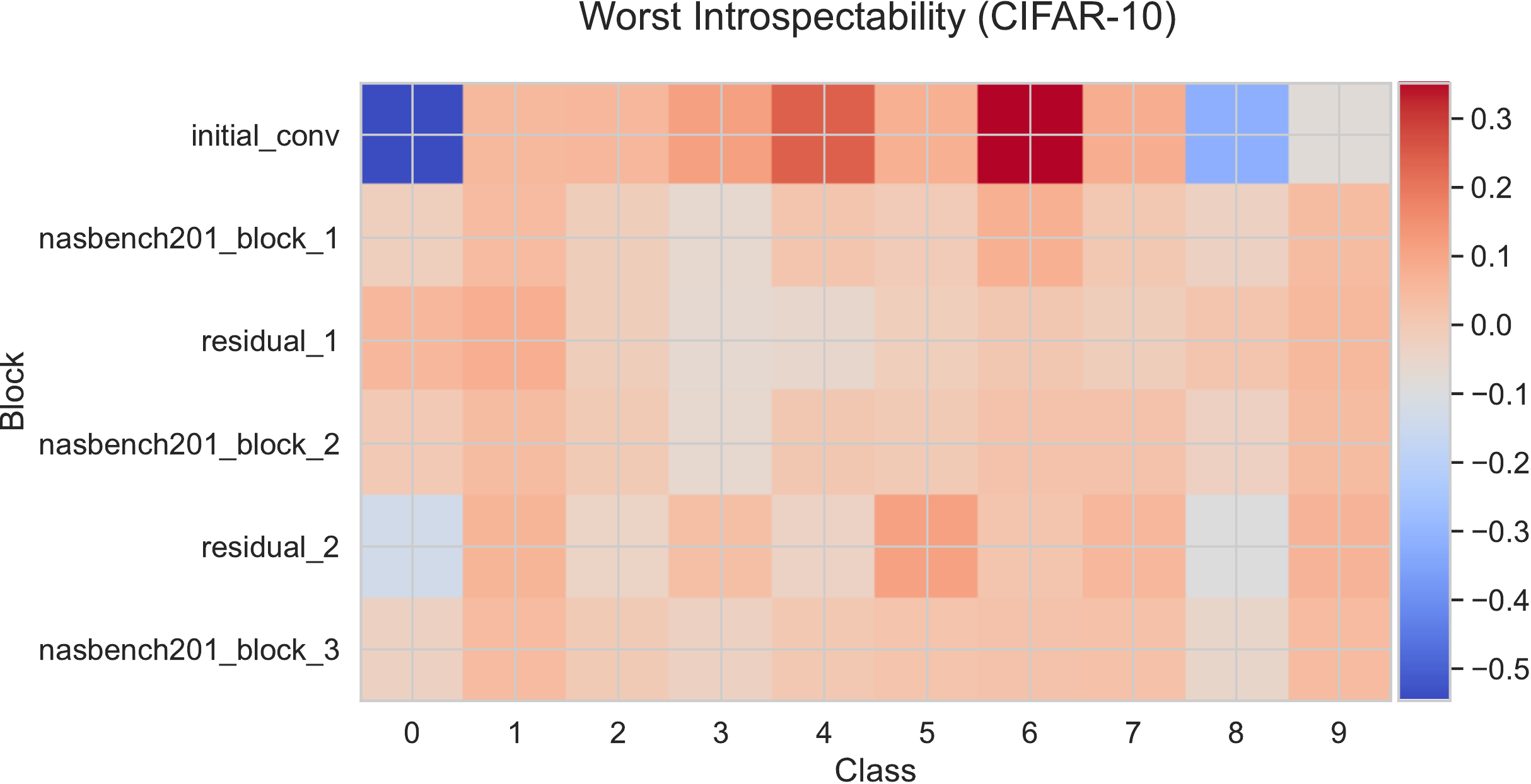}
    \end{subfigure}
    \caption{Mean activations heatmap of the model with highest (top) and lowest (bottom) introspectability on the CIFAR-10 task.}
    \label{fig:heatmap}
\end{figure}

To gain a better qualitative understanding of the introspectability metric, we visualize the activations of the Pareto-optimal solutions of each task.
In \figref{fig:heatmap},
the solutions of the highest and lowest introspectability are shown for CIFAR-10. The MNIST and ImageNet-16-120 activations are shown in the supplemental material. Within each layer, the activations are normalized using z-score normalization. The activations within each block per class are then averaged for the purpose of visualization. The differences between the highest- and lowest-scoring models are quite apparent; the activation patterns for each class in higher-scoring models have notable variance, whereas they are quite constant in lower-scoring models.

\begin{figure}%
    \centering
    \includegraphics[width=.9\linewidth]{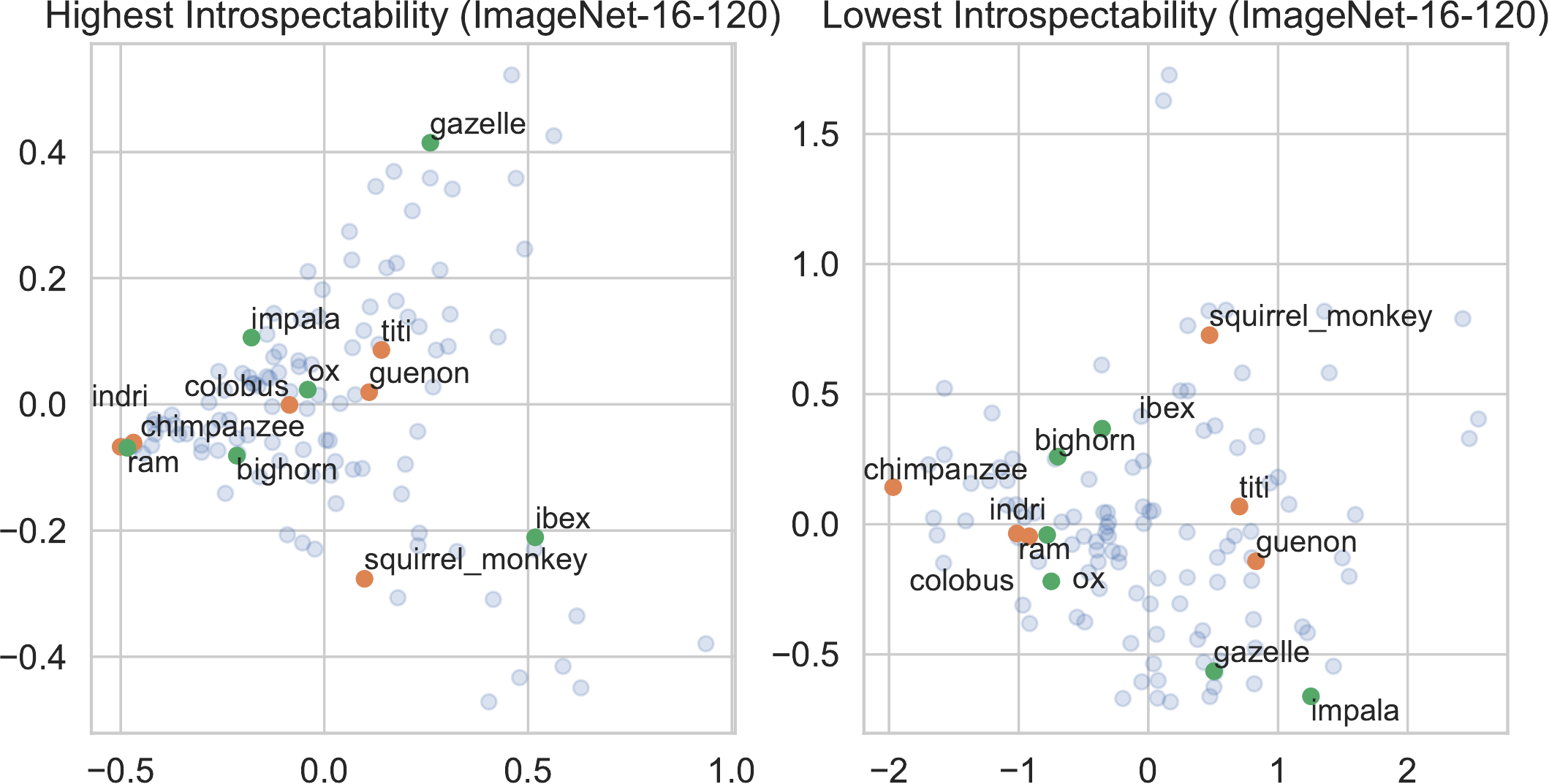}
    \caption{2D PCA of the mean activations per class from the non-dominated models with highest and lowest introspectability on ImageNet-16-120. Classes with the mutual hypernym ``primate'' (orange) and ``bovid'' (green) are emphasized.}
    \label{fig:pca2dinet}
\end{figure}

We use principal component analysis~(PCA) to visualize the mean activations of the best- and worst-scoring non-dominated models for introspectability on ImageNet-16-120, as shown in \figref{fig:pca2dinet}. The labels belonging to the hyponyms of two synsets, primates and bovids, are highlighted for comparison. While neither projection demonstrates complete disentanglement, the higher-introspectability model has more clusters of points within a synset with lower angular distance from the origin.

\begin{figure}[t]
    \centering
    \includegraphics[width=.95\linewidth]{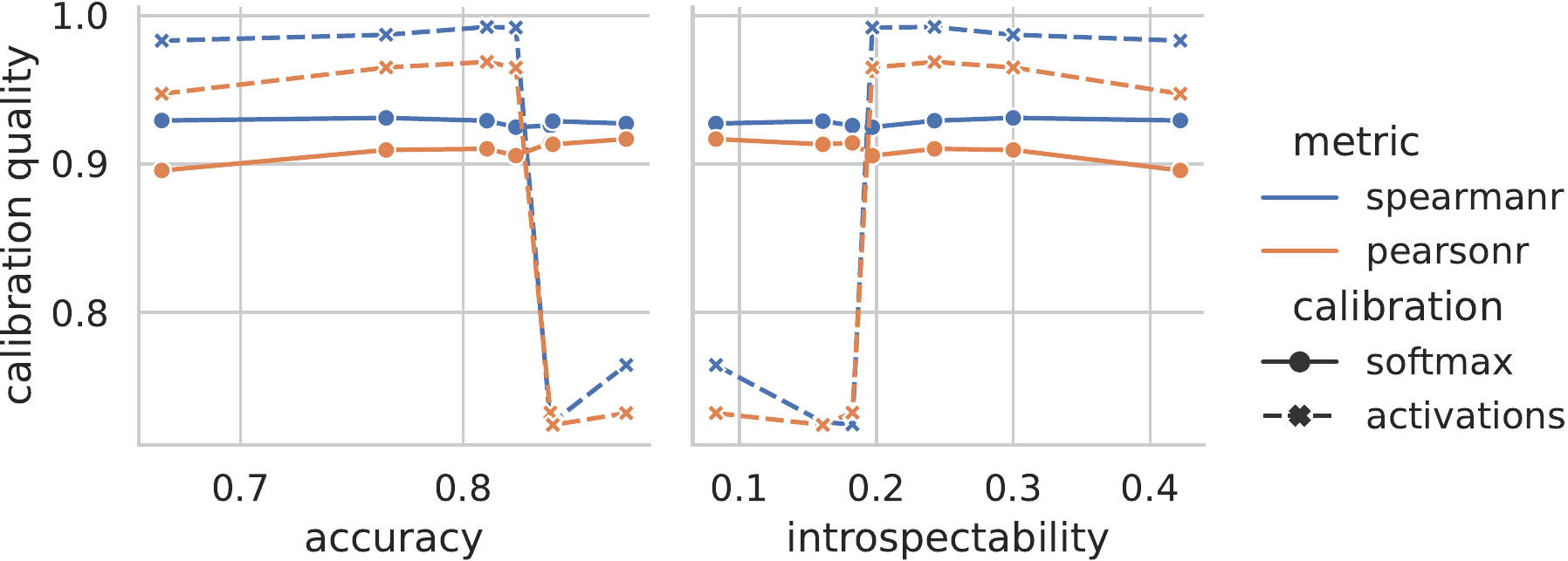}
    \caption{{Calibration quality of Pareto-optimal models discovered by XNAS on CIFAR-10. Calibration using activations improves on softmax for introspectability${>}0.18$.}}
    \label{fig:calibration}
\end{figure}

\paragraph{Introspectability, trustworthiness, and debuggability}
Figures~\ref{fig:heatmap} and~\ref{fig:pca2dinet} demonstrate that increased class-wise disentanglement is a result of optimizing for introspectability. Here, we exploit this to show that introspectability is a surrogate for improved trustworthiness and debuggability of discovered architectures.
Typically, \textit{softmax calibration} is employed to assess the confidence and trustworthiness of predictions. This approach interprets the softmax-activated readout layer outputs of DNN classifiers as probabilities. We propose to use an \textit{activations-based calibration} that \XNAS{} indirectly optimizes for.  %
In activations-based calibration, we 1) collect the mean activations per class from training data, 2) collect the activations of the held-out data, 3) compute the cosine distances between the activations of held-out data and the mean activations per class (similar to Eq.~\eqref{eq:introspectability}), and 4) interpret the distances of the predicted classes as probabilities.
To assess calibration quality, we use the Pearson and Spearman rank correlation coefficients of the calibrated probabilities and the corresponding actual accuracy scores at each probability range on a held-out test set\footnote{We use 50 linear bins to estimate these quantities.}.
Fig.~\ref{fig:calibration} demonstrates that introspectability estimates calibration quality and outperforms softmax calibration when introspectability${>0.18}$.
Notably, the calibration quality tapers off rapidly at this point, indicating that class-wise latent representations are no longer as disentangled. This is an important observation of this approach -- \textit{DNN trustworthiness is a function of class-wise disentanglement}.
This experiment serves as motivation that higher-introspectability models are more conducive to trustworthiness and debuggability, as we will demonstrate.

\begin{figure}
    \centering
    \includegraphics[width=.85\linewidth]{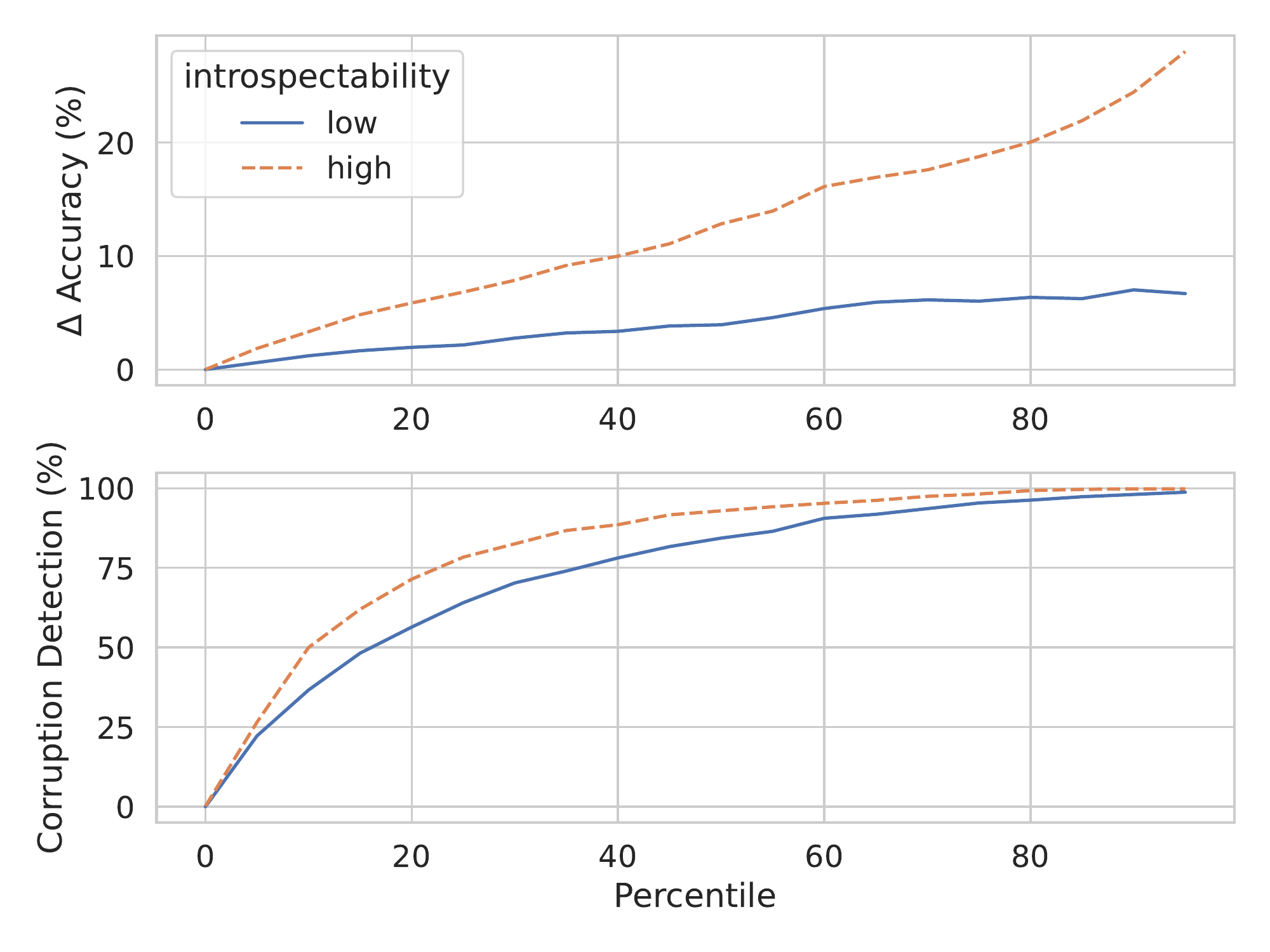}
    \vspace{-1em}
    \caption{(Top) The change in accuracy due to identified mispredictions as a function of the percentile of sample-wise activation calibration scores. (Bottom) The detection rate of corrupted labels as a function of the percentile of sample-wise activation calibration scores. The results are both on the CIFAR-10 dataset.}
    \label{fig:trustworthiness_ab}
\end{figure}

\textit{Identifying mispredictions~~}
We explore the effect of introspectability on the ability to identify mispredictions.
To do so, low-confidence predictions are isolated according to the distribution of activation calibration scores. The quality of identified mispredictions is then quantified as the increase in accuracy due to removing samples.
Figure~\ref{fig:trustworthiness_ab} compares misprediction identification between low- and high-introspectability models on CIFAR-10 -- high-introspectability models demonstrate greater improvement owing to superior calibration and trustworthiness. %

\textit{Debugging data~~}
We further demonstrate the improved model comprehensibility to debug data. We mislabel samples at a corruption rate of 20\% and assess the ability of models to identify the mislabeled data following the distribution of activation calibration scores. Figure~\ref{fig:trustworthiness_ab} shows that high-introspectability models are better equipped to identify bugs in the data on CIFAR-10, achieving a ${\sim}20\%$ higher detection rate at the 20\textsuperscript{th} percentile of scored samples.

\textit{Debugging models~~}
Further model debugging experiments are included in Appendix K -- we identify and repair bugs in models based on pairwise activation distances.

\section{Discussion}

As can be observed in \figref{fig:pareto_fronts}, there exists a clear trade-off between the two metrics, accuracy and introspectability. This trade-off is more pronounced as the classification task grows more complex, i.e.\ in the order of MNIST, CIFAR-10, and ImageNet-16-120.
Furthermore, we discussed how this trade-off is influenced by the selection of operators within a cell.
Interestingly, the observation of this phenomenon seems to contradict the argument by Rudin et al.\ that the accuracy-interpretability trade-off is a false dichotomy~\cite{Rudin2019Why}.
That said, our observation is based on models of the same class and derived from the same search space, while their argument is based on a comparison between different model classes.

The introspectability scores we propose are best suited for technical users, as discussed in Section~\ref{sec:perf_eval}, and should not be confused for an all-telling quantification of fairness, trust, or reliability.
However, the introspectability of models discovered by \XNAS{} has the potential to serve as a requisite criterion before being deployed to users or trusted as a valid
model.
Furthermore, the introspectability score is designed with simplicity and generality in mind. It can be gamed by an adversary with model-level access, e.g.\ by adding futile blocks after the final softmax layer,
but are zeroed out and bypassed with a skip connection. This should be addressed in later work, ideally on a per-application basis.

There are several routes for improvement of the \XNAS{} framework. Foremost, the efficiency can be greatly increased by using weight-sharing techniques~\cite{elsken2019neural,nsga-net}, which reduce the evaluation time of offspring.
Furthermore, there are uninteresting regions of the Pareto front, depending on the application or end user -- NSGA-II can be modified to use reference points of interest to guide the multi-objective search towards more desirable solutions~\cite{10.1145/1143997.1144112}.

{\small
\bibliographystyle{ieee_fullname}
\bibliography{egbib}
}

\clearpage
\begin{strip}
\centering
{\bfseries \Large Supplemental Material}
\end{strip}

\input{supplemental_content}

\end{document}

%% file: supplemental_content.tex
\appendix
\renewcommand\thefigure{\thesection\arabic{figure}}
\renewcommand\thetable{\thesection\arabic{table}}

\tableofcontents

\section{NAS-Bench-201 Overview}\setcounter{figure}{0}\setcounter{table}{0}

\begin{figure}[H]
    \centering
    \includegraphics[width=\linewidth]{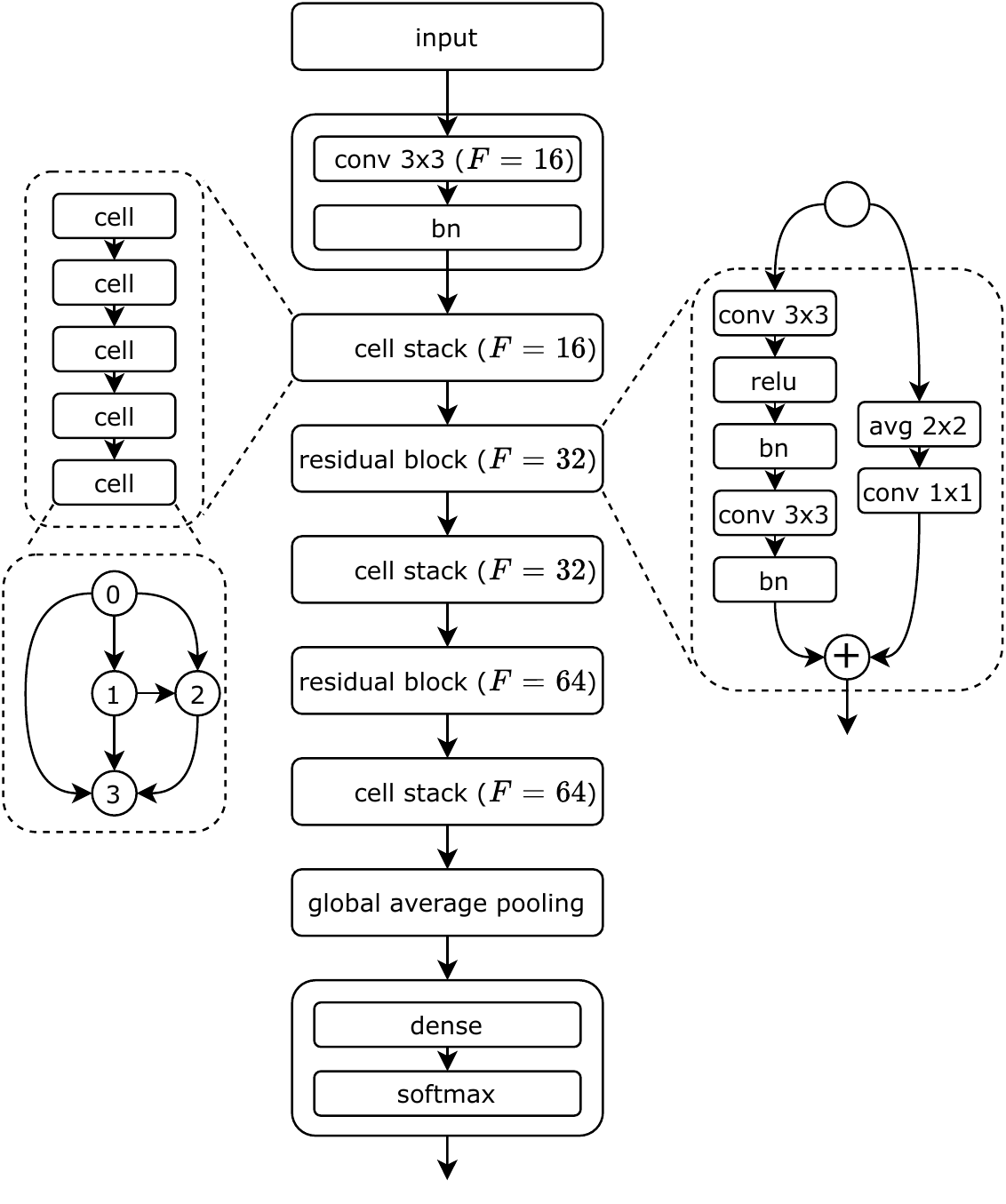}
    \caption{Visualization of the architectures generated by the search space. Note that the first convolutional layer in the main path and the average pooling layer in the shortcut path of each residual block has a stride of 2. \texttt{conv}: 2D convolutional layer. \texttt{bn}: batch normalization. \texttt{relu}: rectified linear~(ReLU) activation layer. \texttt{avg}: 2D average pooling layer. $F$: the number of convolutional filters within each layer of a block.}
    \label{fig:xnas_arch}
\end{figure}

The NAS-Bench-201 search space is comprised of a macro skeleton and a searched cell. An overview is shown in 
\figref{fig:xnas_arch}.

\section{Experiment Setup and Hyperparameters}\setcounter{figure}{0}\setcounter{table}{0}
\paragraph{Setup}

We use a cosine annealing~\cite{cosine} learning rate schedule to decay the learning rate from 0.1 to 0 at the end of the last epoch. We also take half an epoch to warm up the learning rate from 0 to 0.1 at midway through the first epoch.

\paragraph{Data preprocessing} Recall that each raw image $\tensor{X}_i$ has a height of $H$ pixels, width of $W$ pixels, and $C$ color channels. We first scale the image by 255 to map the input domain from $[0..255] \subset \mathbb{N}_0$ to $[0,1] \subset \mathbb{R}$. Then, z-score normalization is applied, i.e.\ the channel-wise mean of the full dataset $\tensor{X}$ is subtracted from each $\tensor{X}_i$ and the result of which is divided by the channel-wise standard deviation of $\tensor{X}$. The resulting data has channel-wise means of zero and standard deviations of one.

\paragraph{Data augmentation} We zero-pad the left and right of each image with $\lceil H / 8 \rceil$ pixels and the top and bottom of each image with $\lceil W / 8 \rceil$ pixels. Then, each image is randomly cropped following a uniform distribution back to shape $H \times W \times C$. Next, the image is flipped horizontally with a probability of 0.5. The final augmentation applied is cutout~\cite{cutout}. Randomly centered rectangular windows with height $2\lceil H / 8 \rceil$ and width $2\lceil W / 8 \rceil$ are selected to be filled with zeros within the bounds of each image.

We do not allow offspring that have the same architecture as another offspring or a previously evaluated architecture.
There are 6 integer variables in the optimization problem, so we set the probability of polynomial mutation per variable to $1/6$.
Table~\ref{tab:hparams} contains the summary of all hyperparameters used across the experiments.

\paragraph{ImageNet-16-120}
The ImageNet-16-120 dataset, originally introduced in~\cite{DBLP:journals/corr/ChrabaszczLH17} and adapted by the NAS-Bench-201 benchmark~\cite{nasbench201}, is a downsampled version of the ImageNet dataset. The dataset facilitates substantially faster experimentation while permitting satisfactory classification results -- performance on ImageNet-16-120 has been shown to be indicative of performance across all of ImageNet.
Each image in the dataset is resized to $16 \times 16$ pixels and only the data for the first 120 classes are retained.

\begin{table}%
    \small
    \centering
    \ra{1.1}
    \begin{tabular}{@{}ccrcl@{}}
        \toprule
        && Hyperparameter && Value \\
        \midrule
        \parbox[t]{2mm}{\multirow{11}{*}{\rotatebox[origin=c]{90}{Model Training}}}
        &&Loss && Cross Entropy \\
        &&Optimizer && SGD \\
        &&Learning Rate (LR) && 0.1 \\
        &&LR Schedule && Cosine Decay \\
        &&Nesterov && Yes \\
        &&Momentum && 0.9  \\
        &&Weight Decay && 0.0005 \\
        &&Batch Size && 512 \\
        &&Epochs && 5$^{*}$, 12$^{\dag}$, 200$^{\ddag}$ \\
        &&Data Normalization && Z-Score (Channel-Wise) \\
        &&Data Augmentation && See Text \\
        \midrule
        \parbox[t]{2mm}{\multirow{4}{*}{\rotatebox[origin=c]{90}{NSGA-II}}}
        &&Population Size && 64 \\
        &&Sampling && Uniform Random \\
        &&Crossover && Simulated Binary $p = 0.9$, $\eta = 3$ \\
        &&Mutation && Polynomial $p = 1/6$, $\eta = 3$ \\
        \bottomrule
    \end{tabular}
    \caption{Summary of hyperparameters used across each experiment. $^{*}$MNIST; $^{\dag}$CIFAR-10; $^{\ddag}$ImageNet-16-120}
    \label{tab:hparams}
\end{table}

\paragraph{Introspectability Regularizer}

Introspectability can be used as a regularization term as it is differentiable. We add this as an auxiliary loss term and naively balance the term with cross entropy with a regularizer weight of 0.5 -- this bounds introspectability to the range $[0,1]$. Because we want to maximize introspectability, we take the cosine similarity instead of the distance. To accumulate activations grouped by classes in TensorFlow, the \texttt{tf.scatter\_nd} operator is used in implementation. The remaining implementation is straightforward.

\section{Additional Activation Heat Maps}\setcounter{figure}{0}\setcounter{table}{0}

\begin{figure*}%
    \centering
    \begin{subfigure}[b]{\linewidth}%
        \centering
        \includegraphics[width=\linewidth]{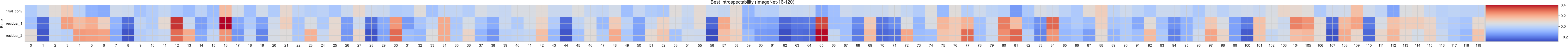}
        \caption{}
        \label{fig:a}
    \end{subfigure}\\%
    \begin{subfigure}[b]{\linewidth}%
        \centering
        \includegraphics[width=\linewidth]{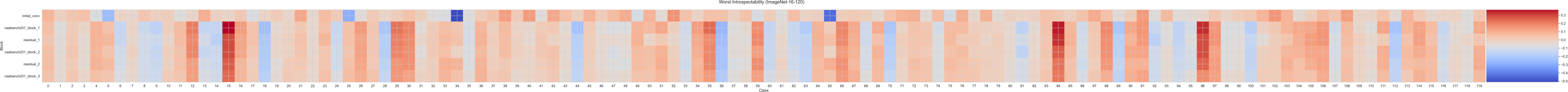}
        \caption{}
        \label{fig:b}
    \end{subfigure}
    \caption{Mean activations heatmap of the model with highest introspectability on the ImageNet-16-120 task.}
    \label{fig:heatmap_last}
\end{figure*}

To gain a better qualitative understanding of the introspectability metric, we visualize the activations of the Pareto-optimal solutions of each task.
In \figref{fig:heatmap_last},
the solutions of the highest and lowest introspectability are shown for MNIST and ImageNet-16-120 (see main text for CIFAR-10). Within each layer, the activations are normalized using z-score normalization. The activations within each block per class are then averaged for the purpose of visualization. The differences between the highest- and lowest-scoring models are quite apparent; the activation patterns for each class in higher-scoring models have notable variance, whereas they are quite constant in lower-scoring models. The heat maps are best viewed digitally.

\begin{figure}[H]
    \centering
    \begin{subfigure}[b]{\linewidth}%
        \centering
        \includegraphics[width=\linewidth]{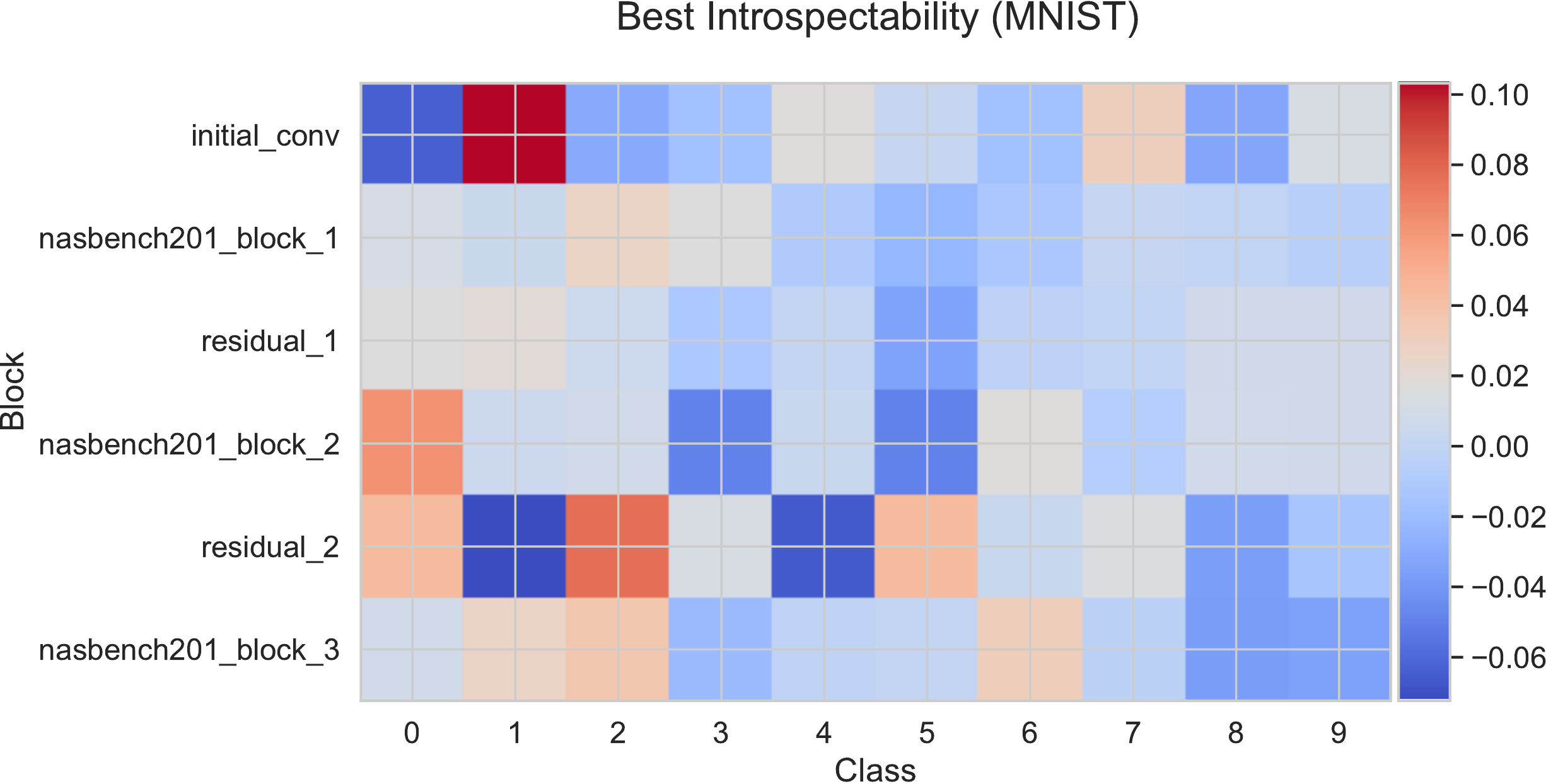}
        \caption{}
        \label{fig:a}
    \end{subfigure}\\%
    \begin{subfigure}[b]{\linewidth}%
        \centering
        \includegraphics[width=\linewidth]{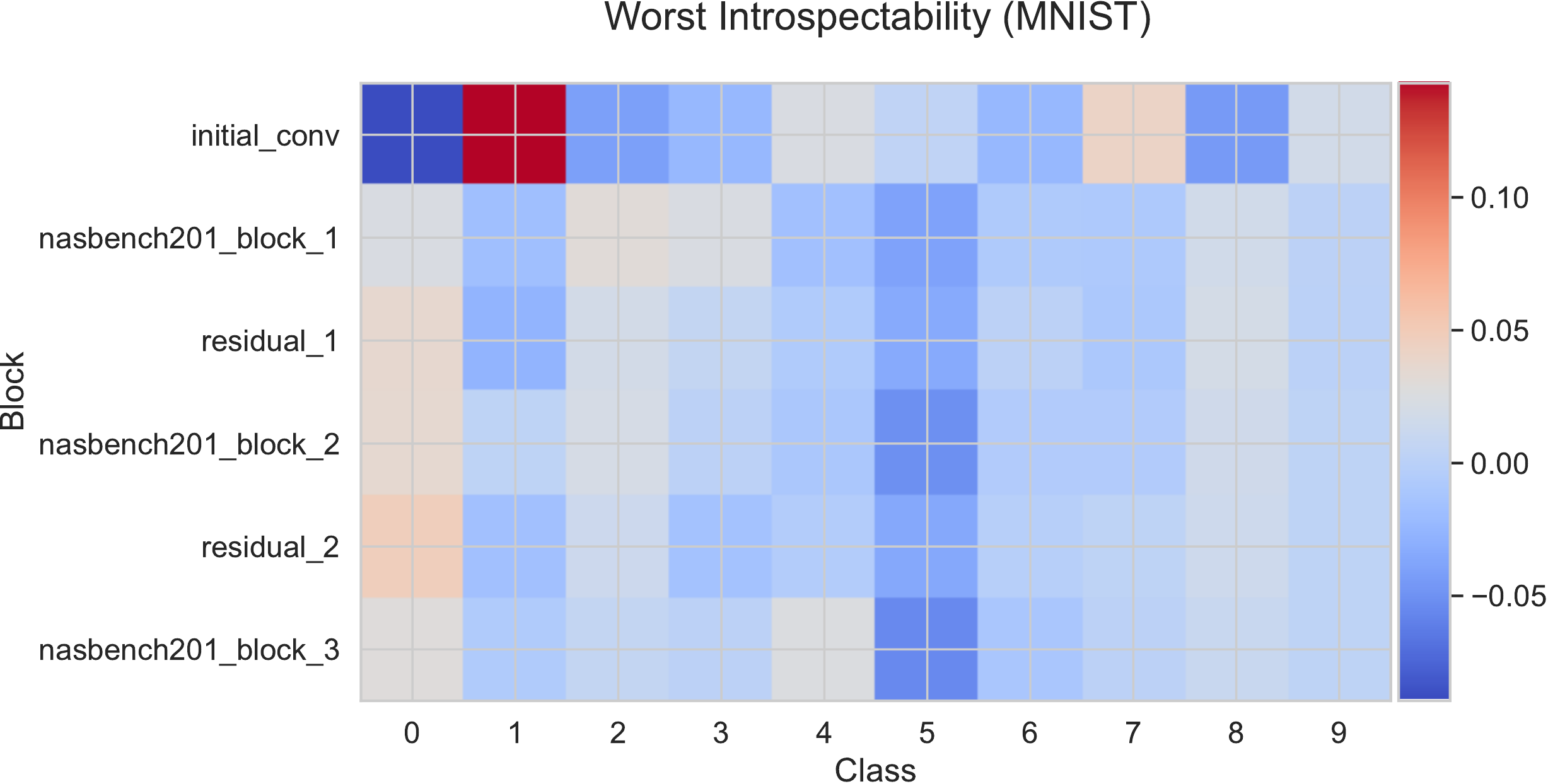}
        \caption{}
        \label{fig:b}
    \end{subfigure}
    \caption{Mean activations heatmap of the models with (a) highest and (b) lowest introspectability on the MNIST task.}
    \label{fig:heatmap_first}
\end{figure}

\ifcsname itIsArxivTime\endcsname%
\else%
\section{Additional PCA Visualizations}\setcounter{figure}{0}\setcounter{table}{0}
The remaining 2D PCA activation visualizations are shown for the MNIST and CIFAR-10 tasks in \figref{fig:pcamnist} and \figref{fig:pcacifar}, respectively. For MNIST, there is little discernible difference between the models with highest and lowest introspectability -- this is expected as the difference between these introspectability scores is small (see the main text). For CIFAR-10, an apparent difference between the two models can be observed; the spread of points about the origin is more Gaussian with the higher-scoring model, which, empirically, should indicate a greater mean cosine distance between class representations. It is important to recall that the PCA projection eliminates thousands of dimensions used to represent activations. Naturally, this causes small changes in introspectability to be less apparent in visualizations.

\begin{figure}[H]
    \centering
    \includegraphics[width=\linewidth]{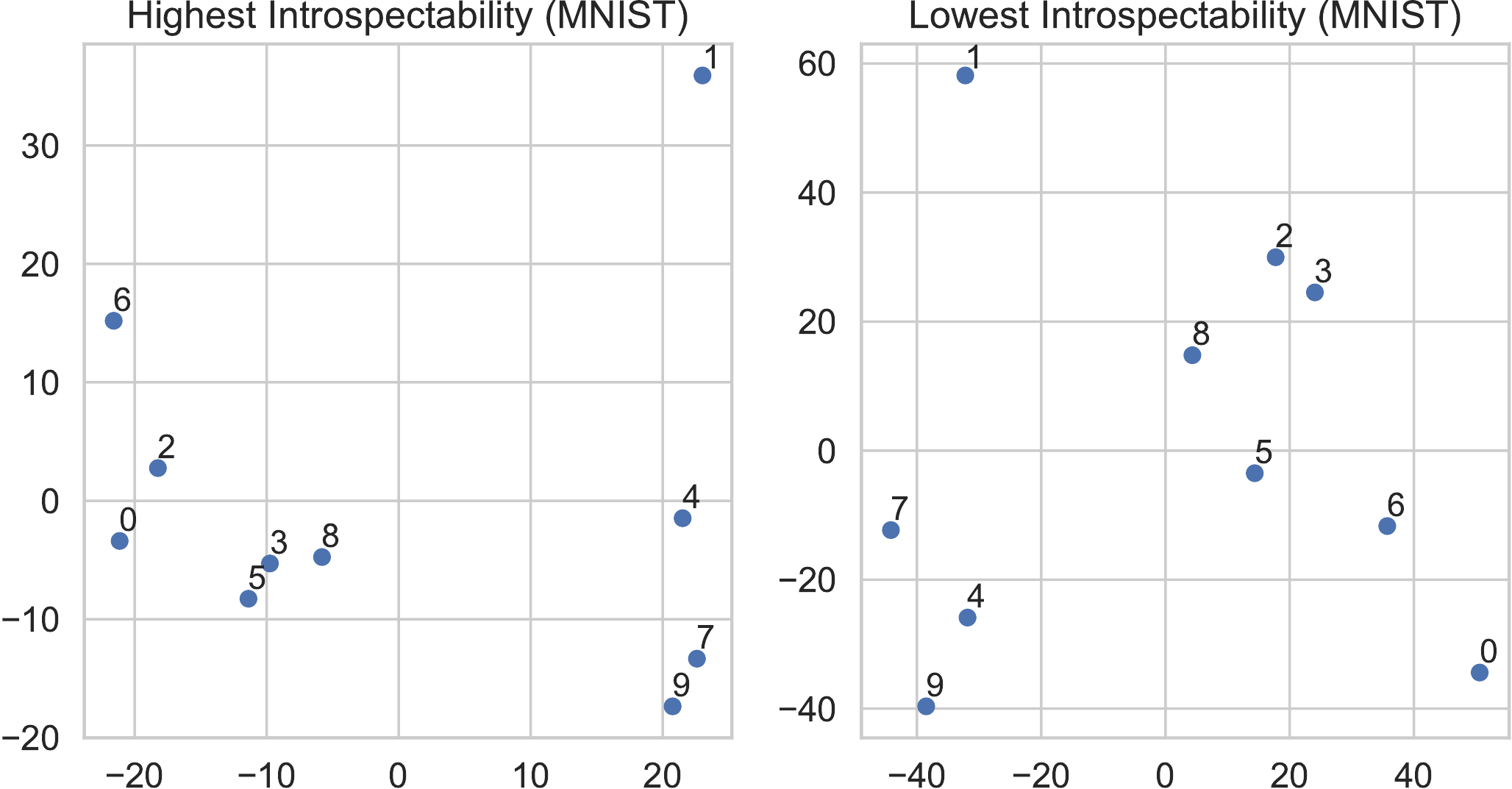}
    \caption{2D PCA of the mean activations per class from the non-dominated models with highest and lowest introspectability on MNIST.}
    \label{fig:pcamnist}
\end{figure}

\begin{figure}[H]
    \centering
    \includegraphics[width=\linewidth]{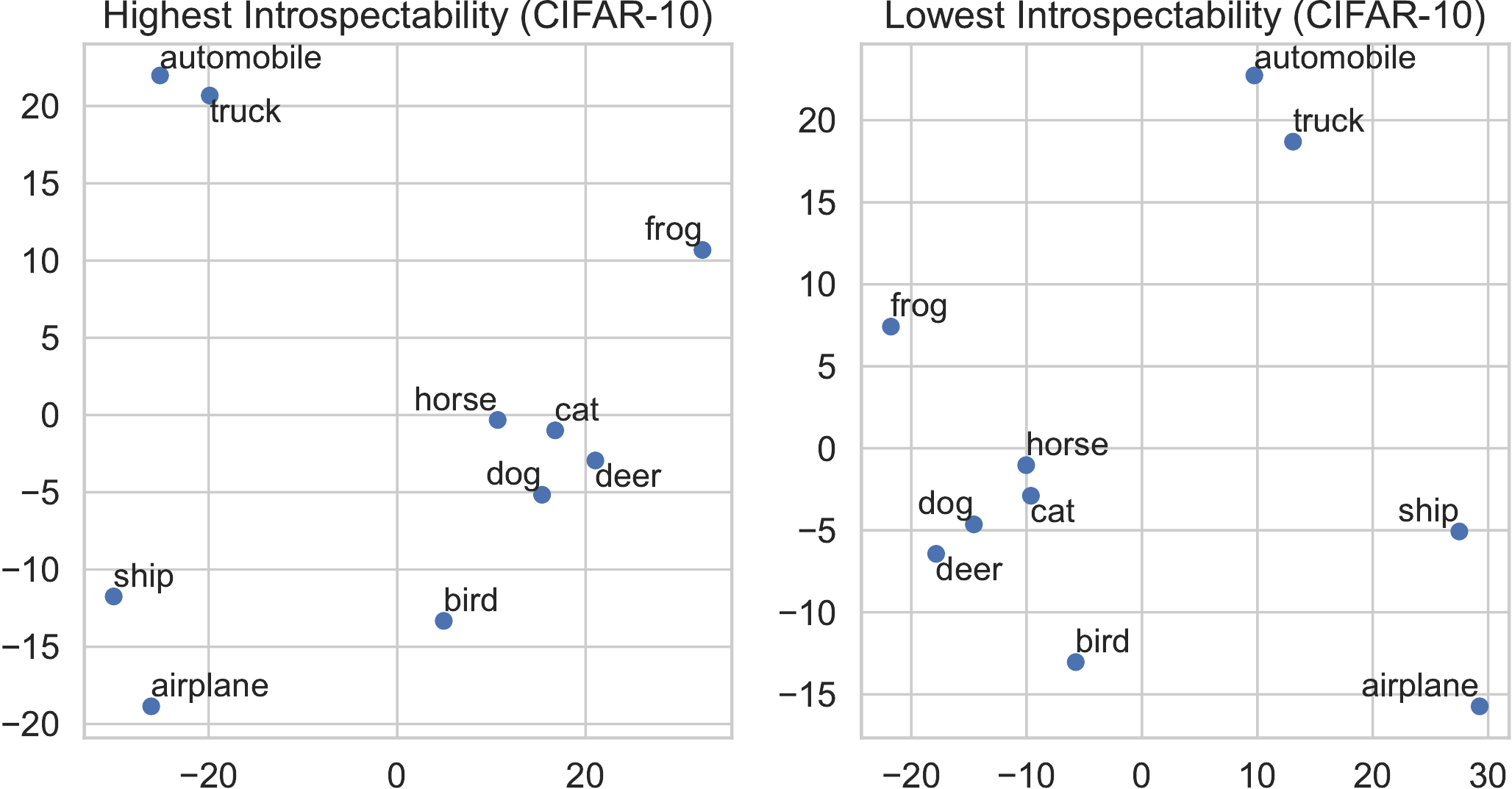}
    \caption{2D PCA of the mean activations per class from the non-dominated models with highest and lowest introspectability on CIFAR-10.}
    \label{fig:pcacifar}
\end{figure}
\fi%

\section{Analysis of Operator Selection}\setcounter{figure}{0}\setcounter{table}{0}

We show the operator-level normalized frequencies selected in the Pareto-optimal solutions of each task in Tables~\ref{tab:op_freq_first}-\ref{tab:op_freq_last}. The 3x3 convolutions are most popular across all tasks, followed by either 3x3 average pooling or ``zeroize'' operators. The skip-connect and 1x1 convolutions are least frequent among these solutions.

\begin{table}[H]
    \centering
    \ra{1.1}
    \begin{tabular}{@{}lr@{}}
        \toprule
        Operation & \makecell[r]{Normalized\\Frequency} \\
        \midrule
        3x3 Conv2D      &      0.51515 \\
        3x3 AvgPool2D   &      0.16667 \\
        Zeroize         &      0.16667 \\
        1x1 Conv2D      &      0.09091 \\
        Skip-Connect    &      0.06061 \\
        \bottomrule
    \end{tabular}
    \caption{Frequency of operations of solutions in the Pareto front (normalized by total cell operations across the Pareto front models) on the MNIST task}
    \label{tab:op_freq_first}
\end{table}

\begin{table}[H]
    \centering
    \ra{1.1}
    \begin{tabular}{@{}lr@{}}
        \toprule
        Operation & \makecell[r]{Normalized\\Frequency} \\
        \midrule
        3x3 Conv2D   &          0.44444\\
Zeroize           &     0.24306 \\
3x3 AvgPool2D    &      0.19097 \\
Skip-Connect    &       0.06250 \\
1x1 Conv2D     &        0.05903 \\
        \bottomrule
    \end{tabular}
    \caption{Frequency of operations of solutions in the Pareto front (normalized by total cell operations across the Pareto front models) on the CIFAR-10 task}
    \label{tab:op_freq}
\end{table}

\begin{table}[H]
    \centering
    \ra{1.1}
    \begin{tabular}{@{}lr@{}}
        \toprule
        Operation & \makecell[r]{Normalized\\Frequency} \\
        \midrule
        3x3 Conv2D  &           0.48039 \\
3x3 AvgPool2D      &    0.21078 \\
Zeroize           &     0.10784 \\
Skip-Connect     &      0.10784 \\
1x1 Conv2D      &       0.09314 \\
        \bottomrule
    \end{tabular}
    \caption{Frequency of operations of solutions in the Pareto front (normalized by total cell operations across the Pareto front models) on the ImageNet-16-120 task}
    \label{tab:op_freq_last}
\end{table}

\section{Frequentist Analysis of Motifs}\setcounter{figure}{0}\setcounter{table}{0}

We conduct analysis of the most common motifs across the Pareto-optimal solutions of each task, as shown in Tables~\ref{tab:motif_freq_first}-\ref{tab:motif_freq_last}.
Recall that the integer-coded cells are encoded as follows:

\begin{itemize}
    \item 0: 3x3 Conv2D
\item 1: 1x1 Conv2D
\item 2: 3x3 AvgPool2D
\item 3: Zeroize
\item 4: Skip-Connect
\end{itemize}

\noindent
We also use an asterisk (*) to match any operator.
Within each table, the encodings of sizes 1 through 5 are shown alongside its normalized frequency of that size. Motifs of size 6 are not shown as we do not evaluate duplicate architectures (other than isomorphisms). The most common motifs reflect the operator frequencies discussed in the previous section. Interestingly, ${>}67\%$ of the Pareto-optimal solutions of each task all have a common motif of size 1, and ${>}45\%$ a common motif of size 2. This suggests that certain cell topologies exhibit inductive biases specific to the task.

\begin{table}[H]
    \centering
    \ra{1.1}
    \begin{tabular}{@{}ccc@{}}
        \toprule
        Size & \makecell[c]{Normalized\\Frequency} & Encoding \\
        \midrule
        1 & 0.81818 & [0 * * * * *] \\
        2 & 0.45455 & [0 * * * 3 *] \\
  2 & 0.45455 & [0 * 0 * * *] \\
  2 & 0.45455 & [0 * * * * 0] \\
  2 & 0.45455 & [* * 0 * * 0] \\
  3 & 0.36364 & [0 * 0 * * 0] \\
  4 & 0.27273 & [0 * 0 4 * 0] \\
  5  & 0.18182 & [0 0 0 4 * 0] \\
  5 & 0.18182 & [0 * 0 4 3 0] \\
        \bottomrule
    \end{tabular}
    \caption{Frequency of encodings of solutions in the Pareto front (normalized by the number of Pareto-optimal solutions) on the MNIST task. The top motif (motifs if tied frequency) for each size is shown only}
    \label{tab:motif_freq_first}
\end{table}

\begin{table}[H]
    \centering
    \ra{1.1}
    \begin{tabular}{@{}ccc@{}}
        \toprule
        Size & \makecell[c]{Normalized\\Frequency} & Encoding \\
        \midrule
  1 & 0.70833 & [* * * * 3 *] \\
  2 & 0.50000 & [* * * * 3 0] \\
  3 & 0.29167 & [0 * 0 * 3 *] \\
  3 & 0.29167 & [* * 0 * 3 0] \\
  4 & 0.18750 & [0 * 0 * 3 0] \\
  5 & 0.08333 & [0 * 0 1 3 0] \\
        \bottomrule
    \end{tabular}
    \caption{Frequency of encodings of solutions in the Pareto front (normalized by the number of Pareto-optimal solutions) on the CIFAR-10 task. The top motif (motifs if tied frequency) for each size is shown only}
    \label{tab:motif_freq}
\end{table}
\begin{table}[H]
    \centering
    \ra{1.1}
    \begin{tabular}{@{}ccc@{}}
        \toprule
        Size & \makecell[c]{Normalized\\Frequency} & Encoding \\
        \midrule
  1 & 0.67647 & [* 0 * * * *] \\
  2 & 0.47059 & [* 0 * * * 0] \\
  3 & 0.23529 & [* 0 * 0 * 0] \\
  4 & 0.11765 & [2 0 * 1 * 0] \\
  4 & 0.11765 & [* 0 * 0 0 0] \\
  4 & 0.11765 & [0 0 * * 0 0] \\
  4 & 0.11765 & [0 0 * 0 * 0] \\
  4 & 0.11765 & [0 * * 0 0 0] \\
  5 & 0.05882 & [3 0 2 2 * 3] \\
  5 & 0.05882 & [2 0 1 1 * 0] \\
  5 & 0.05882 & [2 0 * 1 4 0] \\
  5 & 0.05882 & [2 0 4 0 0 *] \\
  5 & 0.05882 & [2 0 * 0 0 0] \\
        \bottomrule
    \end{tabular}
    \caption{Frequency of encodings of solutions in the Pareto front (normalized by the number of Pareto-optimal solutions) on the ImageNet-16-120 task. The top motif (motifs if tied frequency, up to 5) for each size is shown only}
    \label{tab:motif_freq_last}
\end{table}

\section{Comparing Motifs Across the Pareto Front}\setcounter{figure}{0}\setcounter{table}{0}

\paragraph{Motif Discovery}

\begin{enumerate}
\item Assemble the following data into a tabular structure: architecture encoding, accuracy, and introspectability for the Pareto front of the solutions
\item Sort the data by accuracy and then introspectability which results in data with ascending accuracy and descending introspectability
\item Record the count of each block for each architecture encoding
\item For each architecture encoding in the sorted data, enumerate all applicable motifs of size 1 to 5 (motifs of size 6 cannot exist as architectures are not evaluated multiple times). For example, some architecture encoding $[a, b, c, d, e, f]$ has $\sum_{k=1}^{5}{6 \choose k}$ motifs, e.g.\ $[a, *, *, *, e, *]$ and $[*, b, c, *, *, f]$ but not, say, $[*, b, c, *, *, e]$. This is nearly equivalent to its power set minus $\emptyset$ and the original sequence. An asterisk here implies a match with any other operator, and thus allows for the comparison of motifs between different architectures
\item  For each motif, compute the absolute value of the Spearman rank correlation coefficient between the ranks of the solutions in the sorted data and whether each solution has the motif. If applicable, the count of the operator is used instead of a simple indicator flag. The intuition here is that we discover interesting architectures that demonstrably are favored more in one part of the Pareto front than another, e.g.\ the high-accuracy vs. high-introspectability regions
\item  In addition to each motif having a correlation score, we also record the support (the number of solutions with the motif) and the motif size
\item  Compute the Pareto front of the scored motifs (the costs being the correlation score, the support and the motif size) to identify the most salient motifs. We heuristically eliminate motifs that have support less than 3 or correlation less than 0.2
\end{enumerate}

\begin{figure}[H]
    \centering
    \begin{subfigure}[b]{.5\linewidth}%
        \centering
        \includegraphics[width=.95\linewidth]{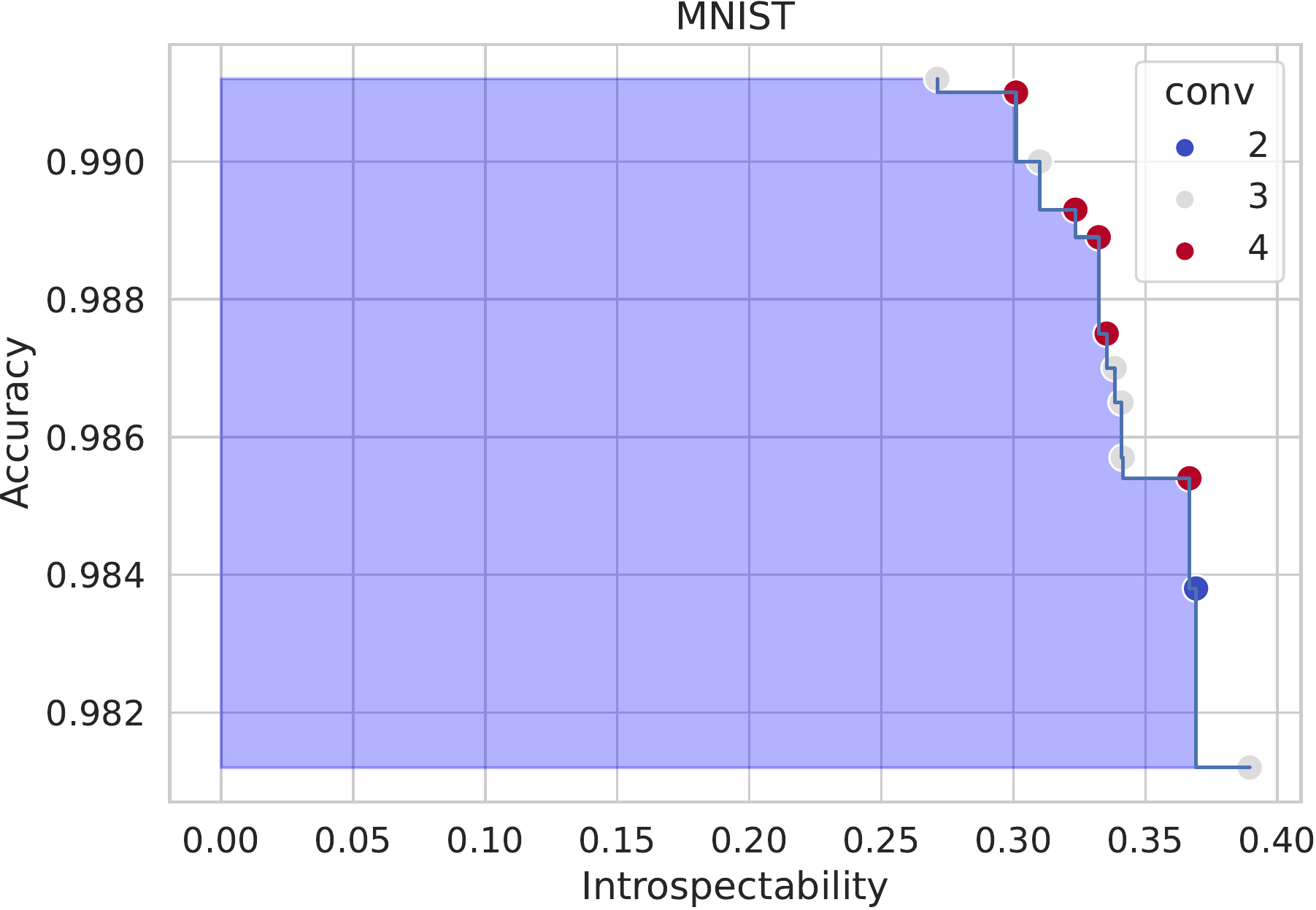}
    \end{subfigure}%
    \begin{subfigure}[b]{.5\linewidth}%
        \centering
        \includegraphics[width=.95\linewidth]{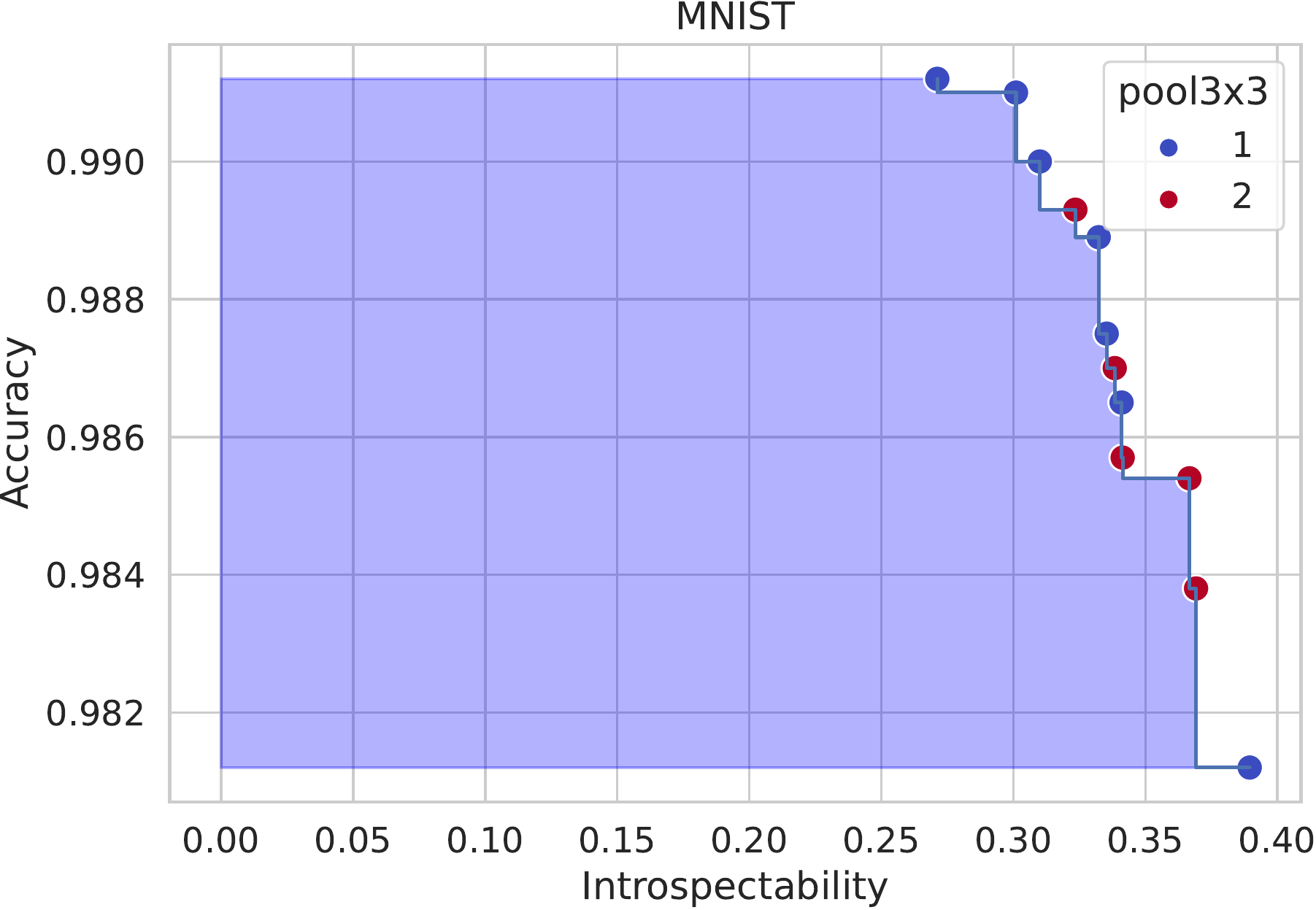}
    \end{subfigure}
    \caption{The Pareto front for the MNIST task with solutions colored by the number of convolutional (left) and pooling (right) layers.}
\end{figure}

\begin{figure}[H]
    \centering
    \begin{subfigure}[b]{.5\linewidth}%
        \centering
        \includegraphics[width=.95\linewidth]{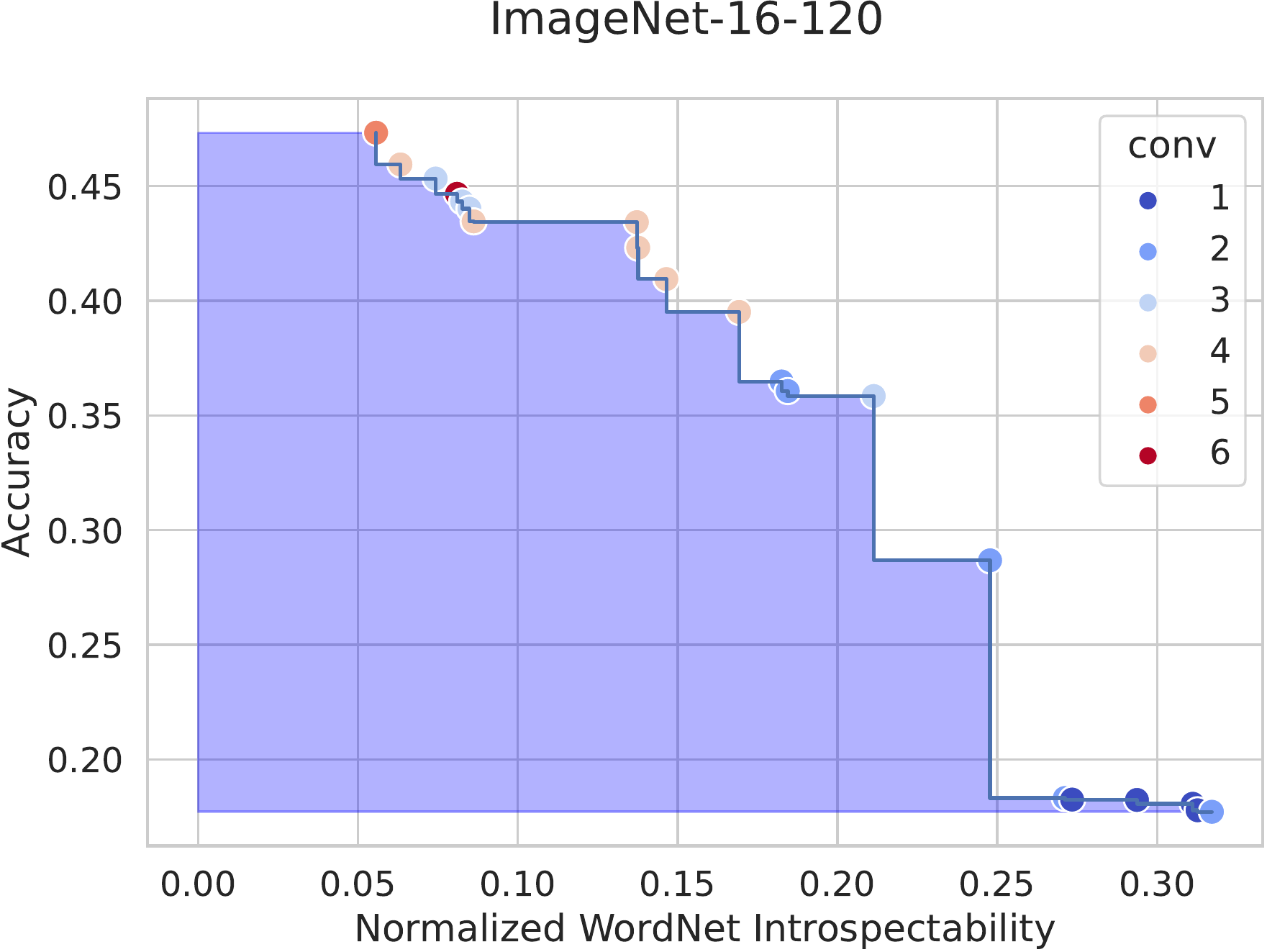}
    \end{subfigure}%
    \begin{subfigure}[b]{.5\linewidth}%
        \centering
        \includegraphics[width=.95\linewidth]{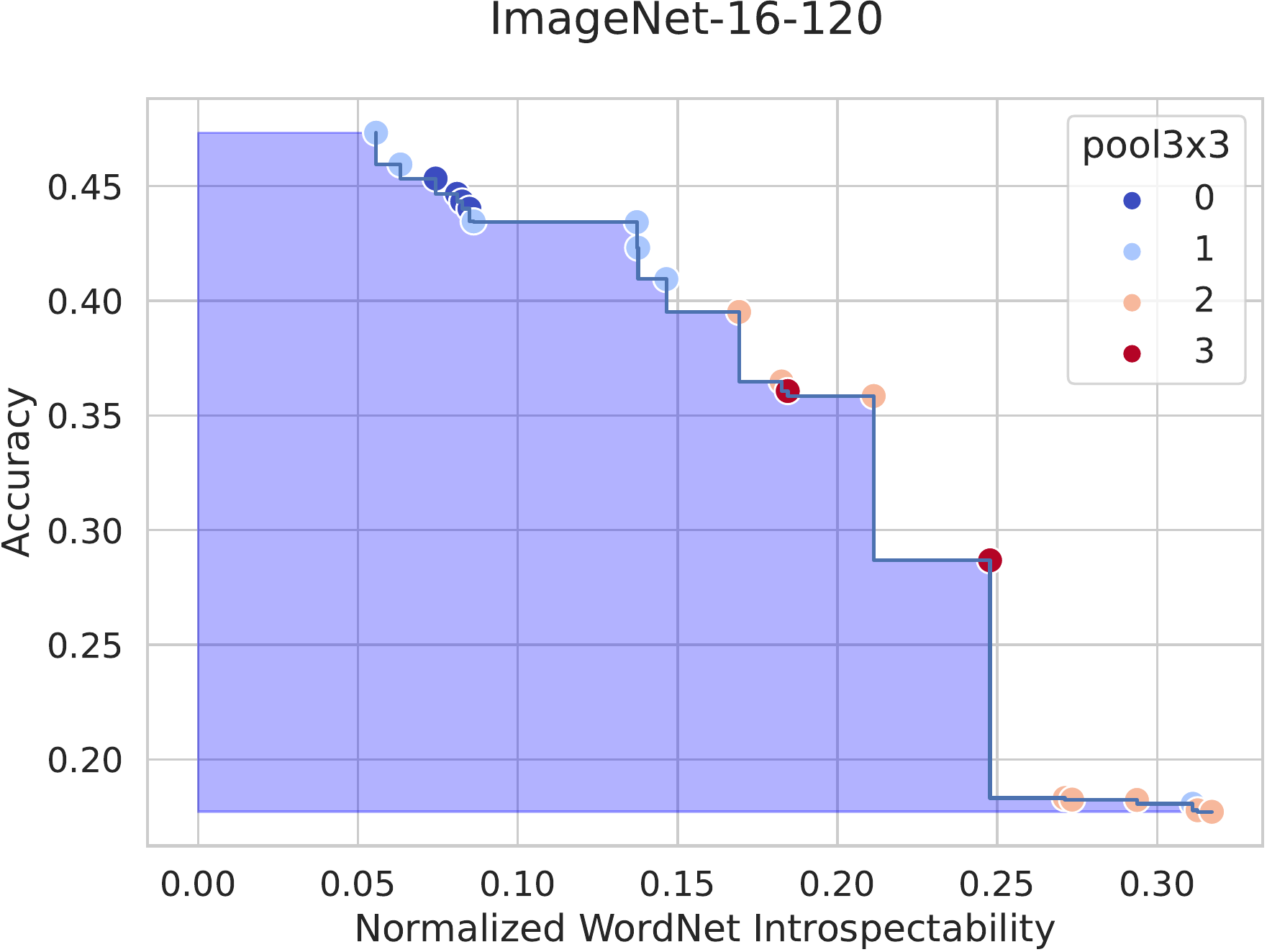}
    \end{subfigure}
    \caption{The Pareto front for the ImageNet-16-120 task with solutions colored by the number of convolutional (left) and pooling (right) layers.}
\end{figure}

\begin{figure}[H]
    \centering
    \includegraphics[width=\linewidth]{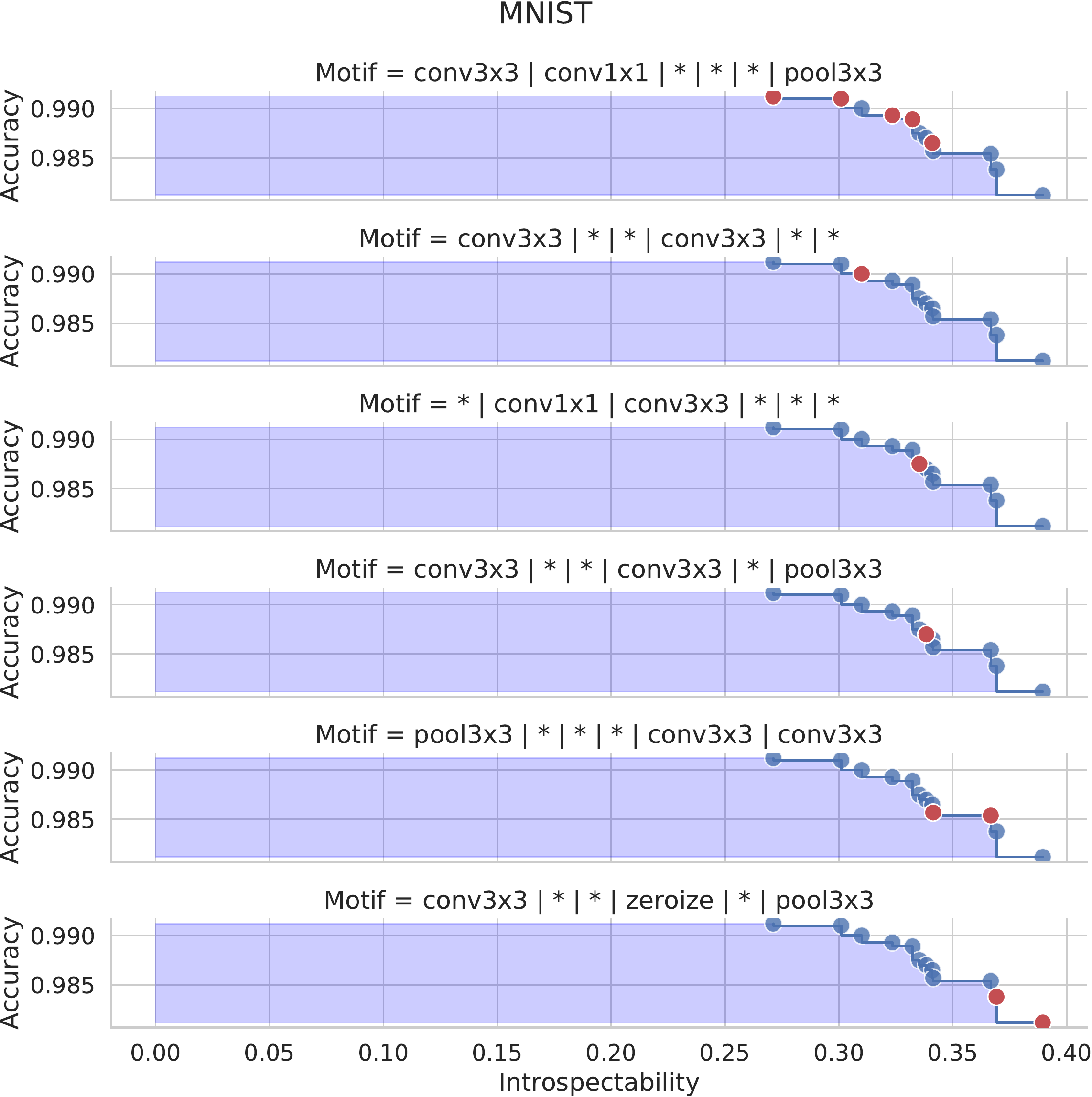}
    \caption{All discovered motifs among the Pareto optimal solutions on the MNIST task. See text for description of the motif discovery process. Each red solution indicates that its architecture has the motif shown in the sub-plot title. The remaining solutions are shown in blue. For the N/A plot, none of the discovered motifs apply to the architecture.}
\end{figure}
\begin{figure}[H]
    \centering
    \includegraphics[width=\linewidth]{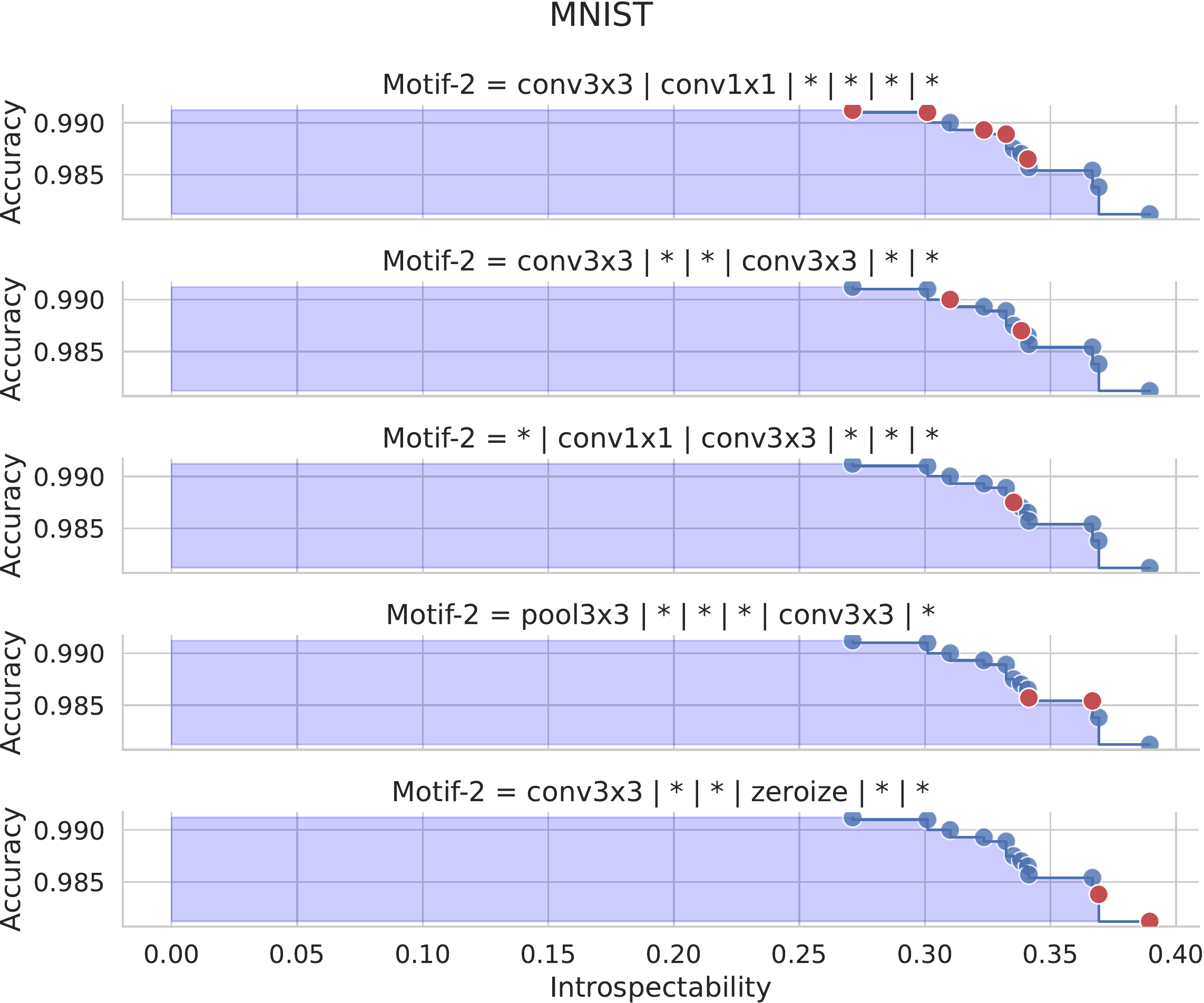}
    \caption{Discovered motifs of size 2 among the Pareto optimal solutions on the MNIST task. See text for description of the motif discovery process. Each red solution indicates that its architecture has the motif shown in the sub-plot title. The remaining solutions are shown in blue. For the N/A plot, none of the discovered motifs apply to the architecture.}
\end{figure}
\begin{figure}[H]
    \centering
    \includegraphics[width=\linewidth]{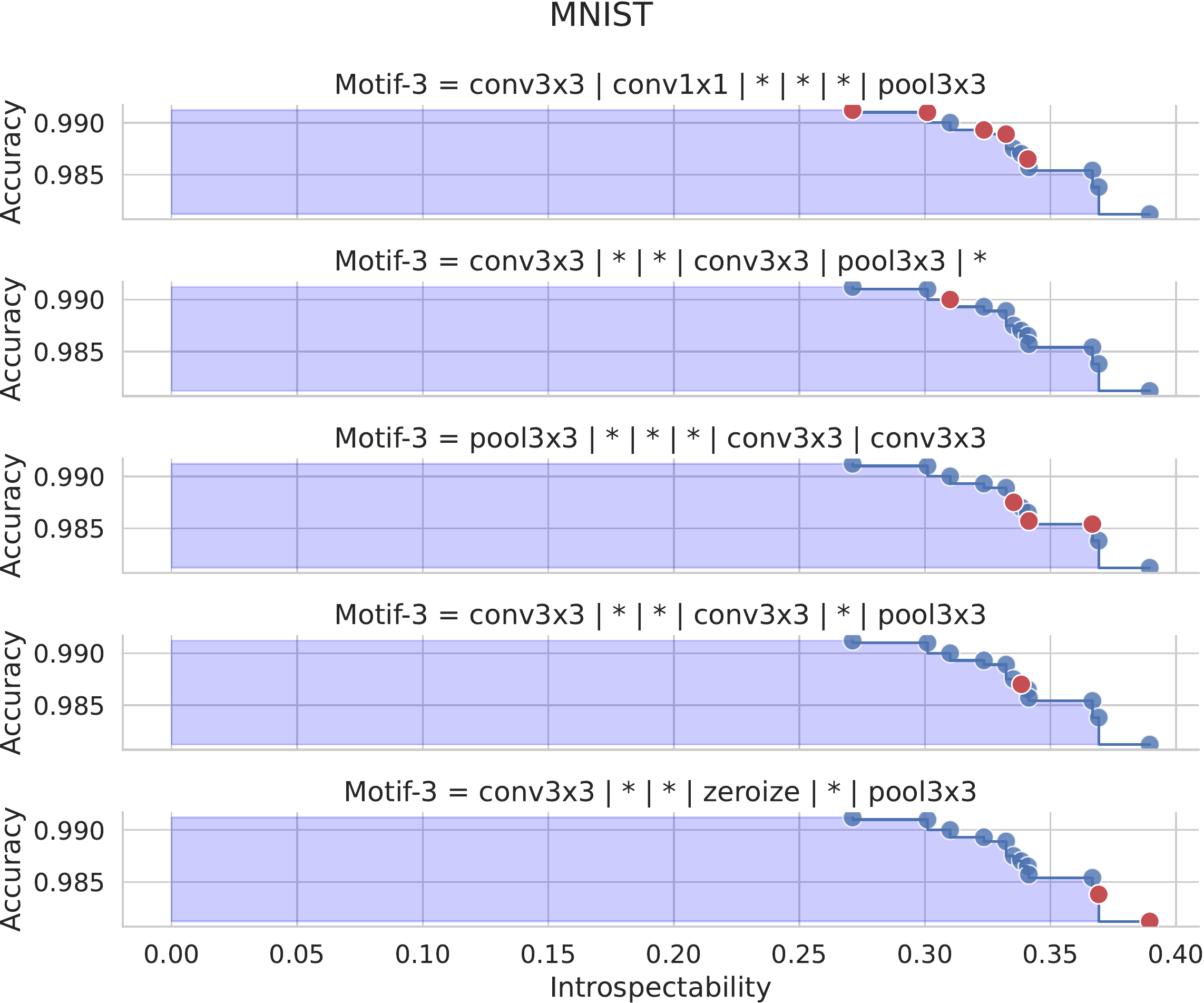}
    \caption{Discovered motifs of size 3 among the Pareto optimal solutions on the MNIST task. See text for description of the motif discovery process. Each red solution indicates that its architecture has the motif shown in the sub-plot title. The remaining solutions are shown in blue. For the N/A plot, none of the discovered motifs apply to the architecture.}
\end{figure}
\begin{figure}[H]
    \centering
    \includegraphics[width=\linewidth]{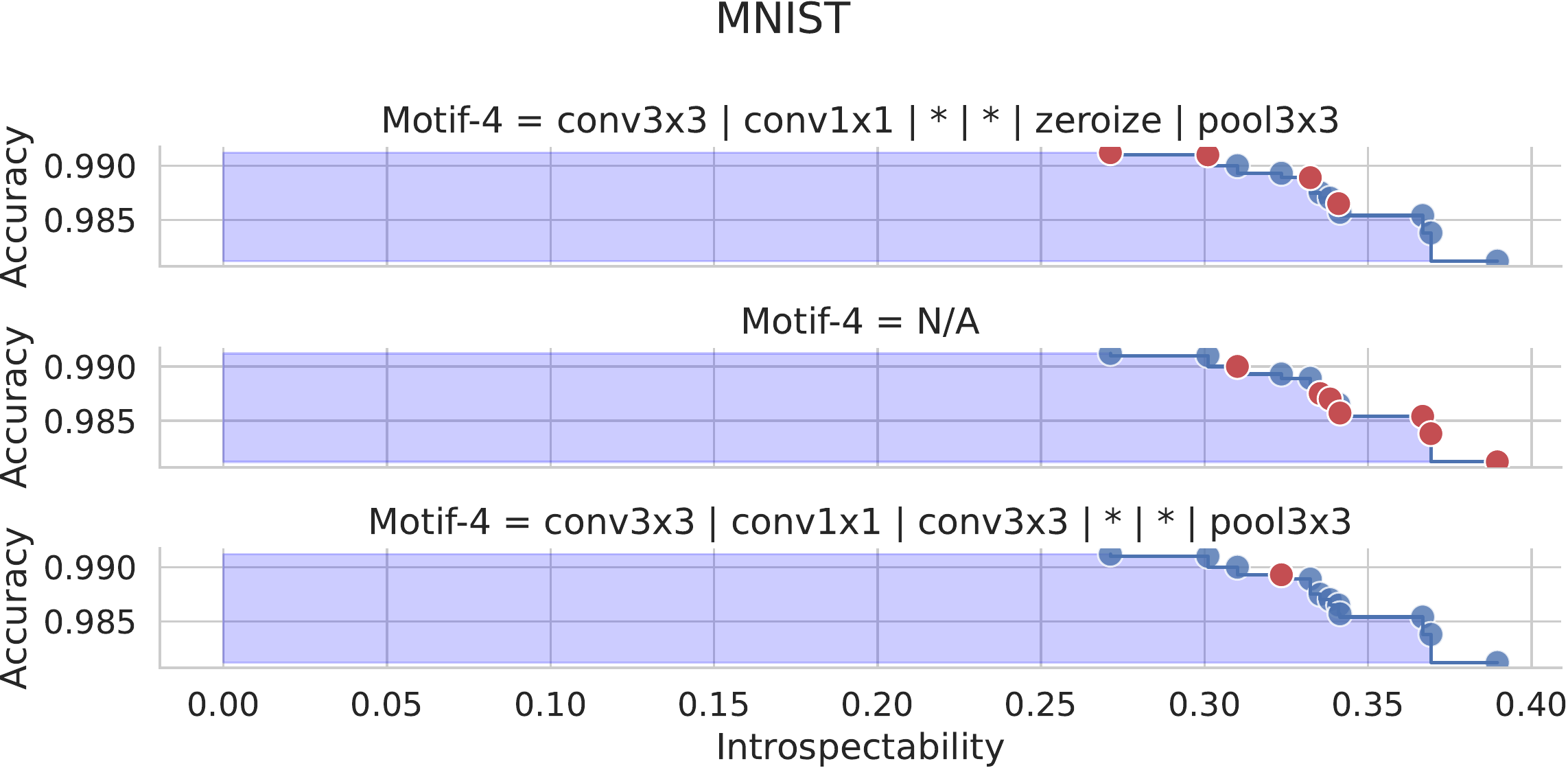}
    \caption{Discovered motifs of size 4 among the Pareto optimal solutions on the MNIST task. See text for description of the motif discovery process. Each red solution indicates that its architecture has the motif shown in the sub-plot title. The remaining solutions are shown in blue. For the N/A plot, none of the discovered motifs apply to the architecture.}
\end{figure}
\begin{figure}[H]
    \centering
    \includegraphics[width=\linewidth]{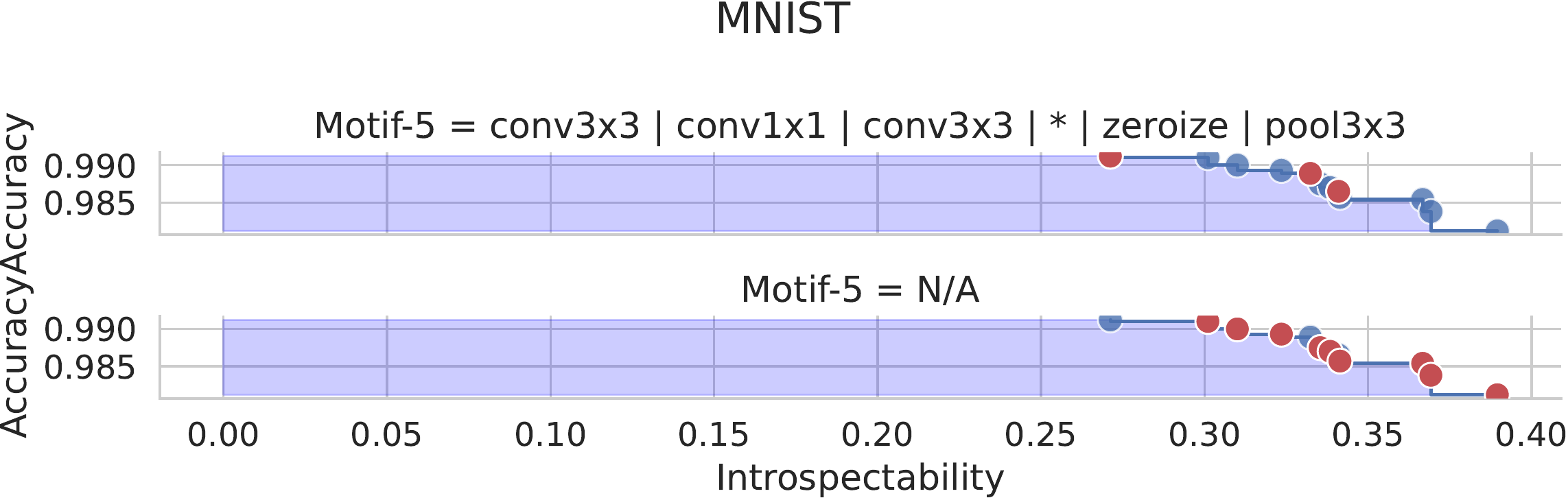}
    \caption{Discovered motifs of size 5 among the Pareto optimal solutions on the MNIST task. See text for description of the motif discovery process. Each red solution indicates that its architecture has the motif shown in the sub-plot title. The remaining solutions are shown in blue. For the N/A plot, none of the discovered motifs apply to the architecture.}
\end{figure}

\begin{figure}[H]
    \centering
    \includegraphics[width=\linewidth]{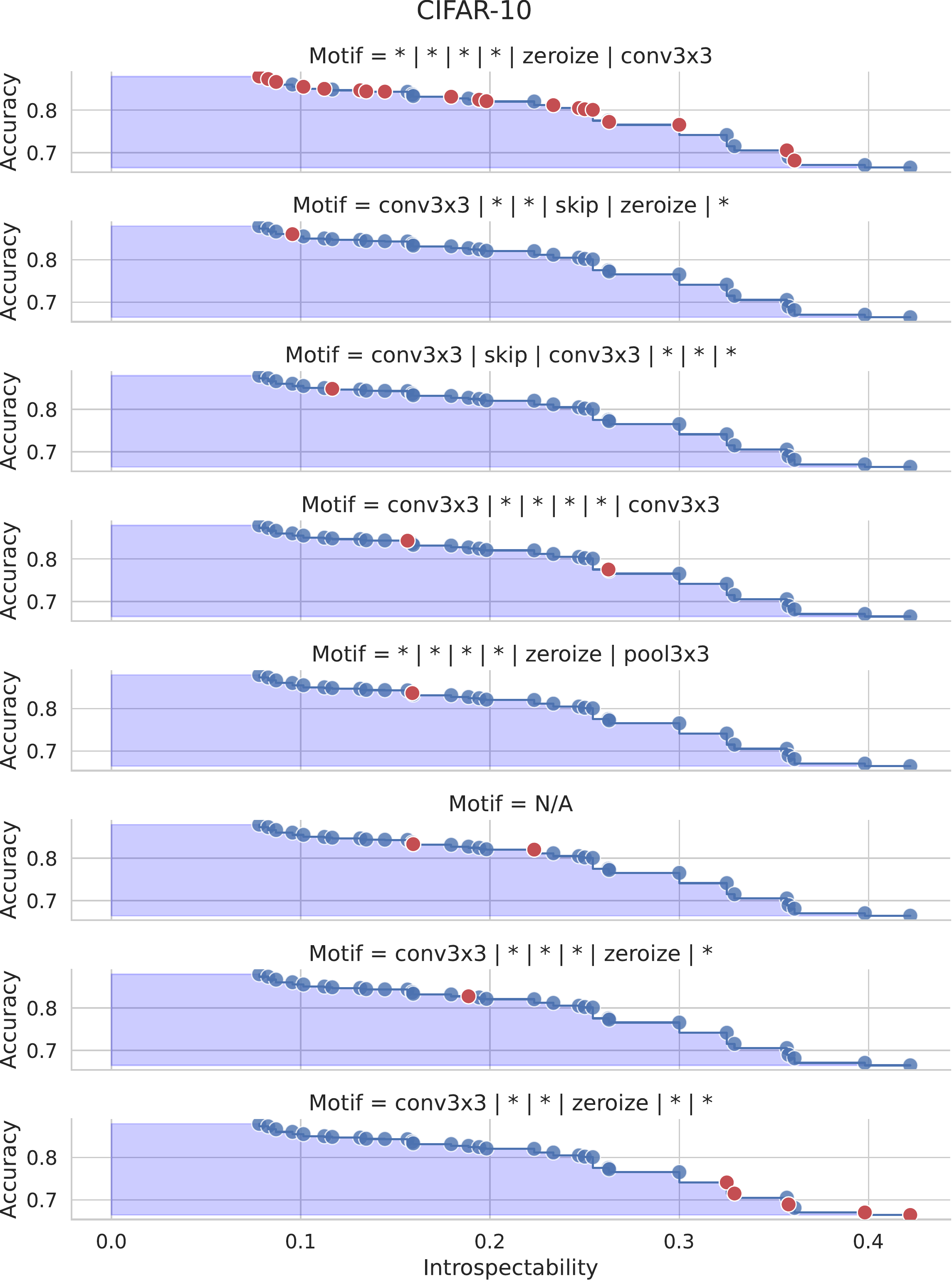}
    \caption{All discovered motifs among the Pareto optimal solutions on the CIFAR-10 task. See text for description of the motif discovery process. Each red solution indicates that its architecture has the motif shown in the sub-plot title. The remaining solutions are shown in blue. For the N/A plot, none of the discovered motifs apply to the architecture.}
\end{figure}
\begin{figure}[H]
    \centering
    \includegraphics[width=\linewidth]{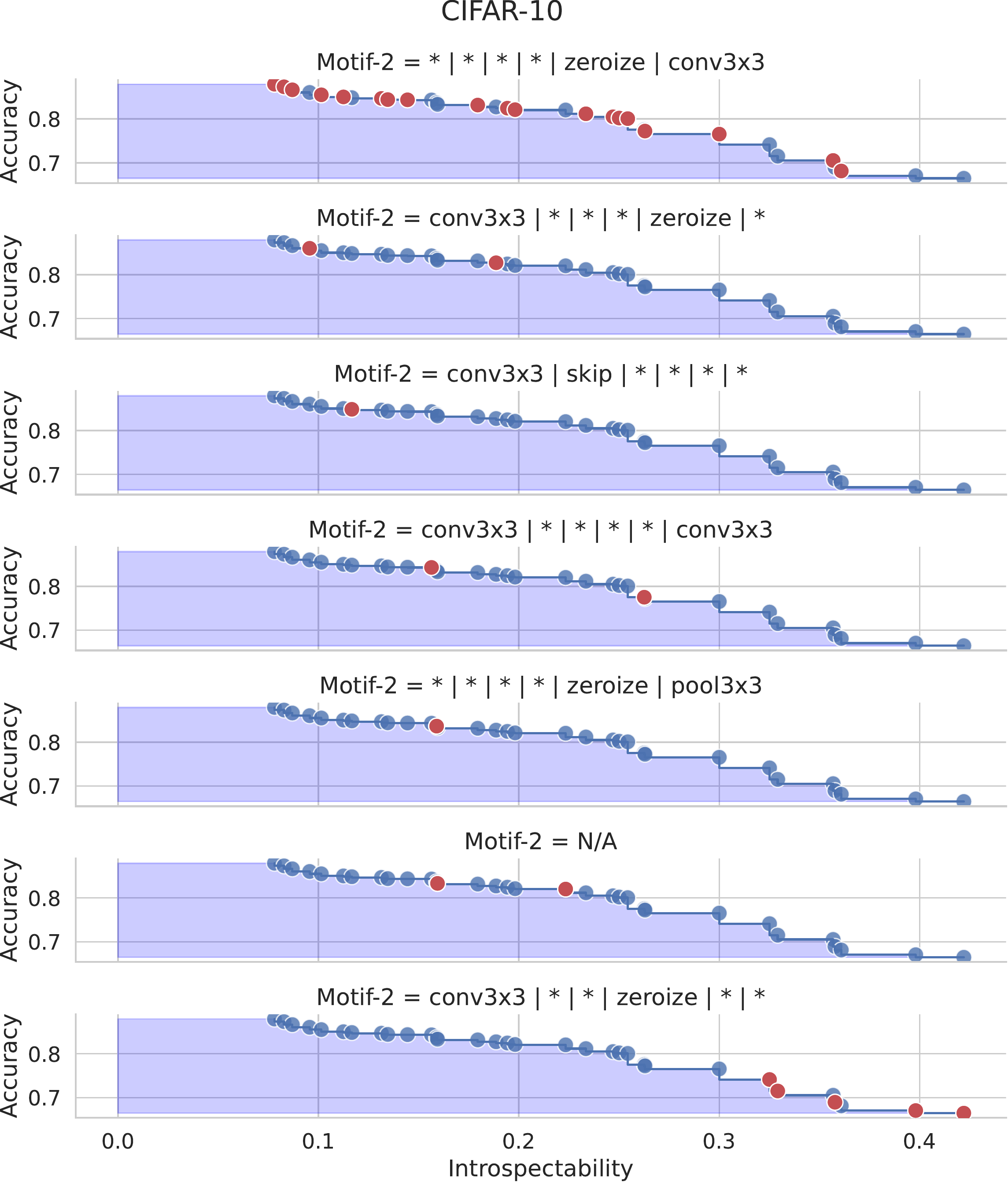}
    \caption{Discovered motifs of size 2 among the Pareto optimal solutions on the CIFAR-10 task. See text for description of the motif discovery process. Each red solution indicates that its architecture has the motif shown in the sub-plot title. The remaining solutions are shown in blue. For the N/A plot, none of the discovered motifs apply to the architecture.}
\end{figure}
\begin{figure}[H]
    \centering
    \includegraphics[width=\linewidth]{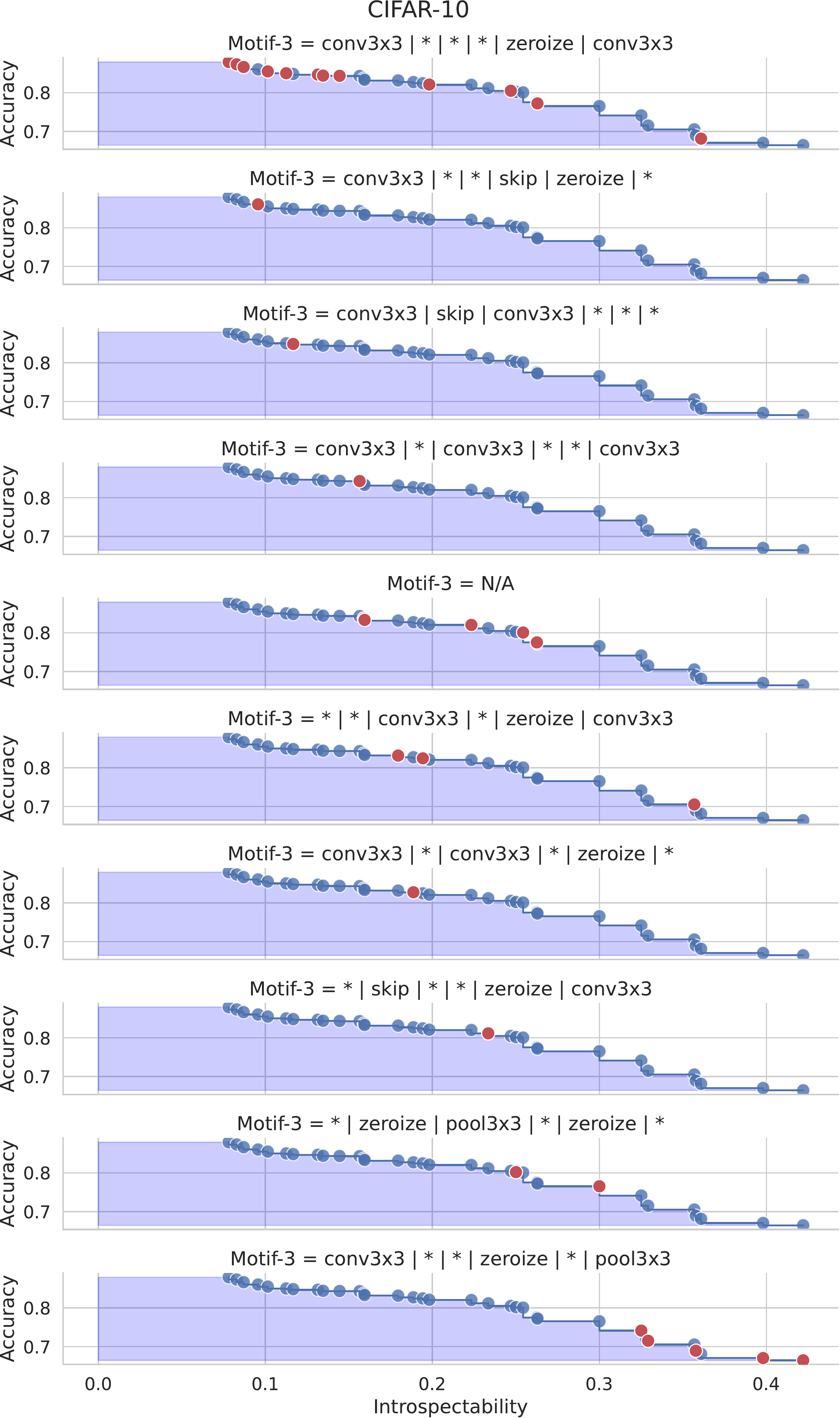}
    \caption{Discovered motifs of size 3 among the Pareto optimal solutions on the CIFAR-10 task. See text for description of the motif discovery process. Each red solution indicates that its architecture has the motif shown in the sub-plot title. The remaining solutions are shown in blue. For the N/A plot, none of the discovered motifs apply to the architecture.}
\end{figure}
\begin{figure}[H]
    \centering
    \includegraphics[width=\linewidth]{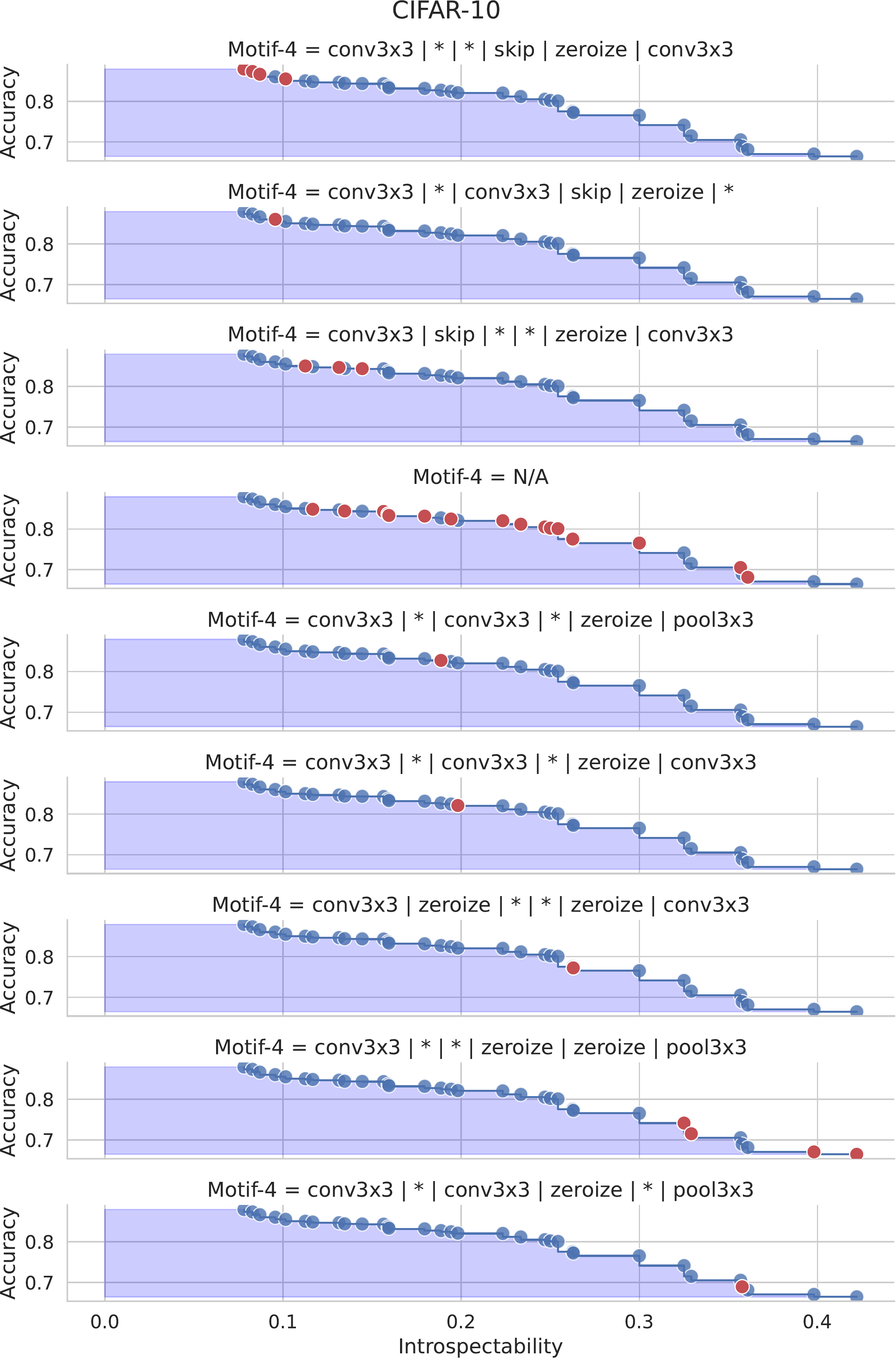}
    \caption{Discovered motifs of size 4 among the Pareto optimal solutions on the CIFAR-10 task. See text for description of the motif discovery process. Each red solution indicates that its architecture has the motif shown in the sub-plot title. The remaining solutions are shown in blue. For the N/A plot, none of the discovered motifs apply to the architecture.}
\end{figure}

\begin{figure}[H]
    \centering
    \includegraphics[width=\linewidth]{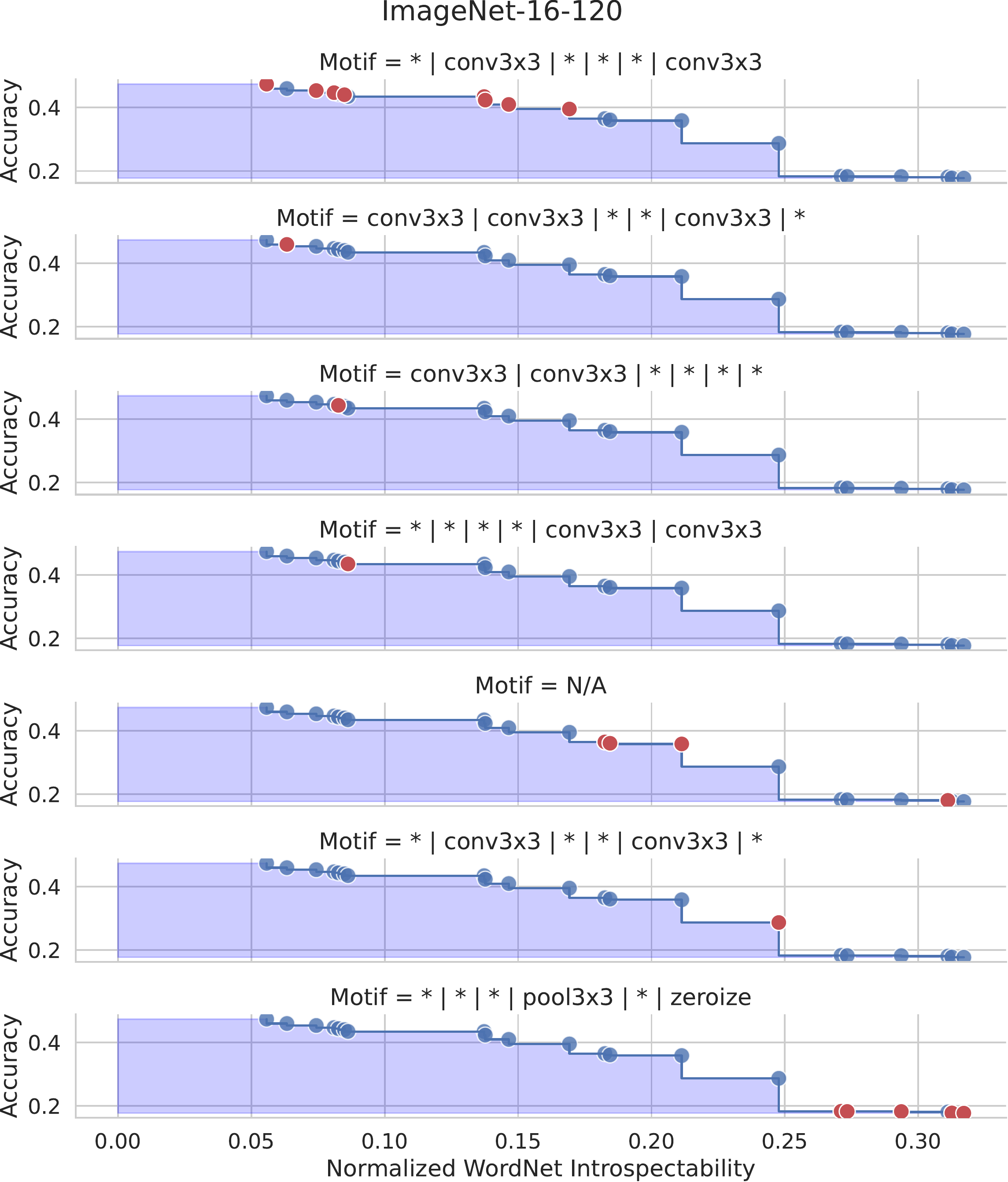}
    \caption{All discovered motifs among the Pareto optimal solutions on the ImageNet-16-120 task. See text for description of the motif discovery process. Each red solution indicates that its architecture has the motif shown in the sub-plot title. The remaining solutions are shown in blue. For the N/A plot, none of the discovered motifs apply to the architecture.}
\end{figure}
\begin{figure}[H]
    \centering
    \includegraphics[width=\linewidth]{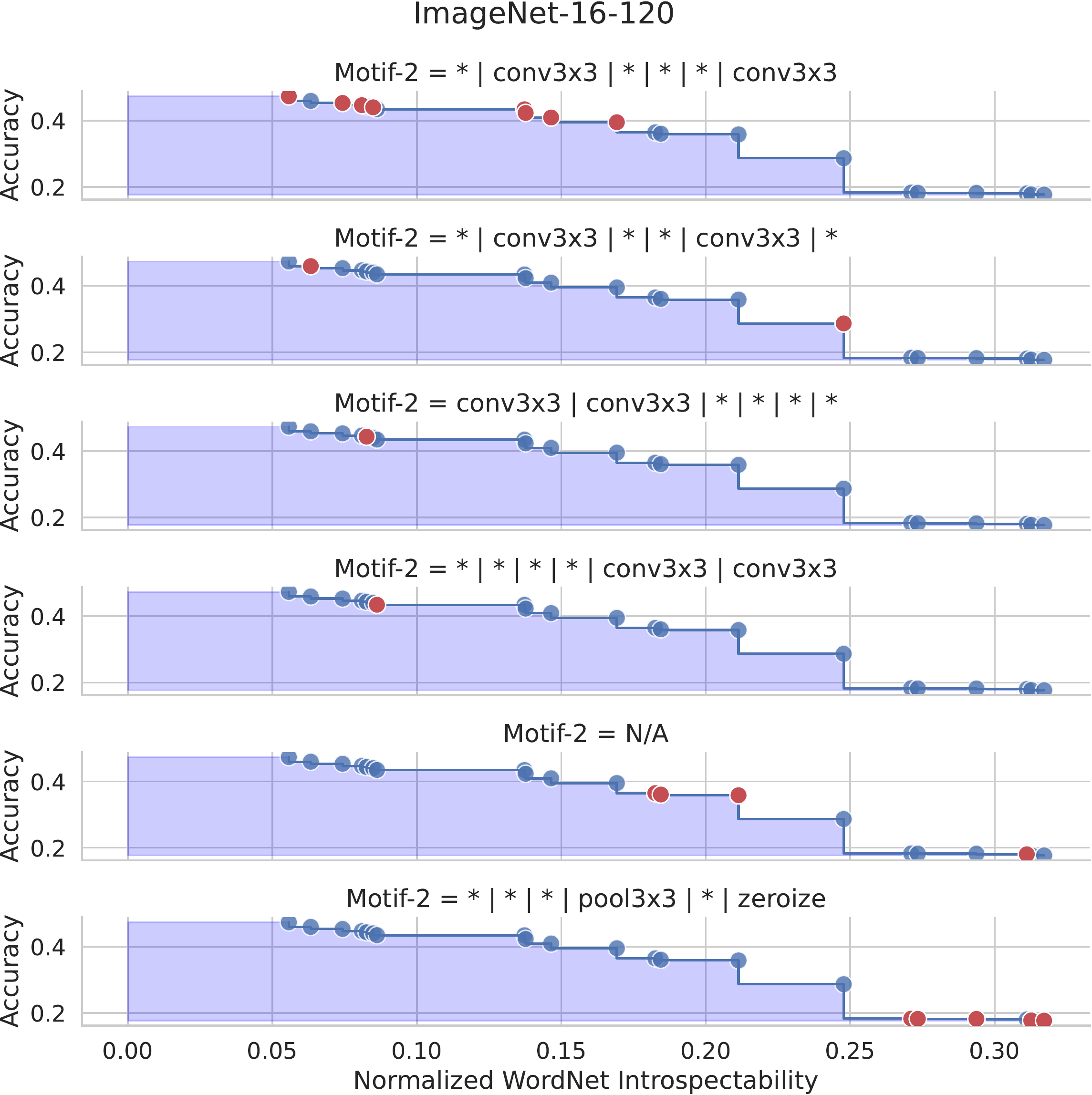}
    \caption{Discovered motifs of size 2 among the Pareto optimal solutions on the ImageNet-16-120 task. See text for description of the motif discovery process. Each red solution indicates that its architecture has the motif shown in the sub-plot title. The remaining solutions are shown in blue. For the N/A plot, none of the discovered motifs apply to the architecture.}
\end{figure}
\begin{figure}[H]
    \centering
    \includegraphics[width=\linewidth]{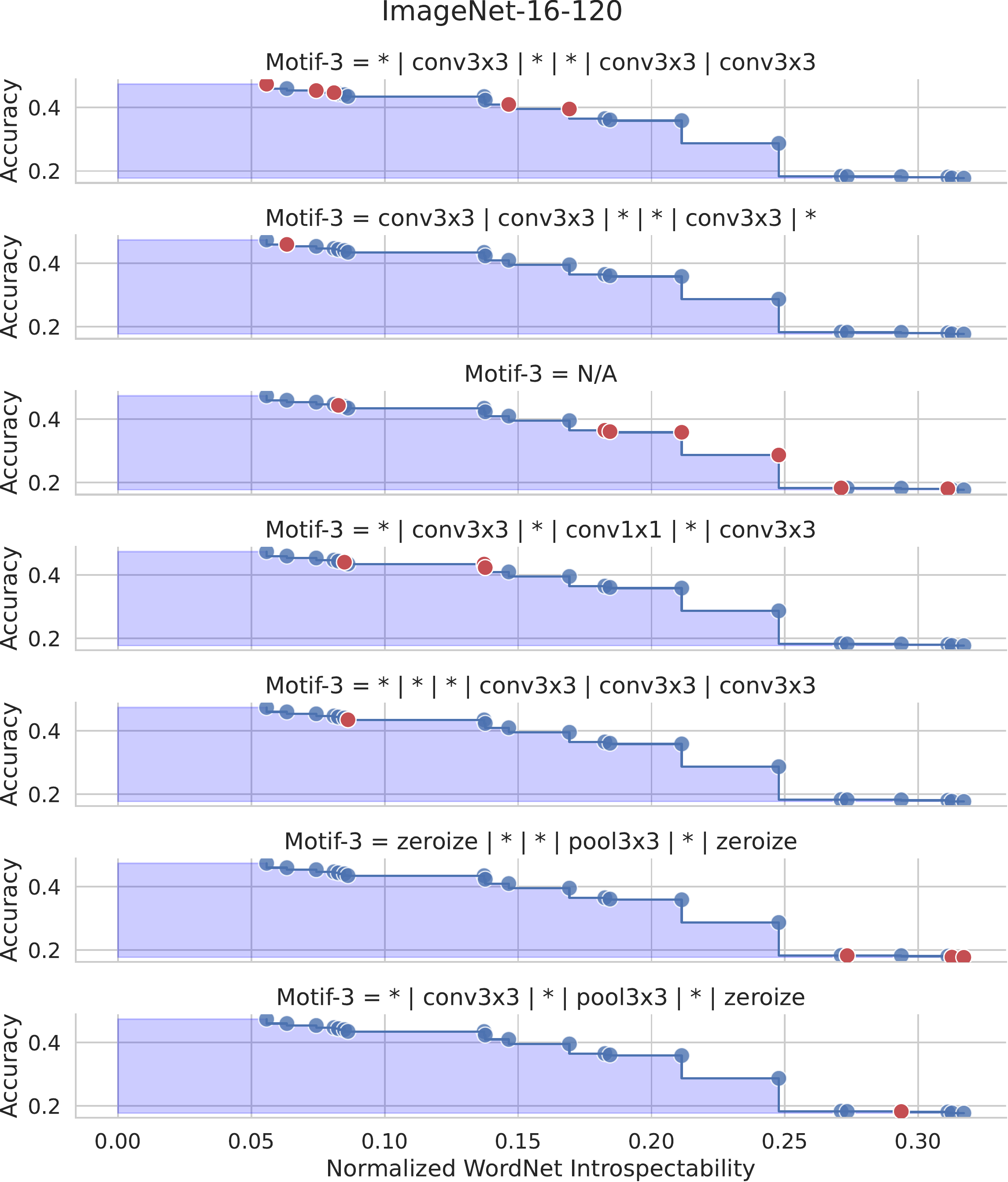}
    \caption{Discovered motifs of size 3 among the Pareto optimal solutions on the ImageNet-16-120 task. See text for description of the motif discovery process. Each red solution indicates that its architecture has the motif shown in the sub-plot title. The remaining solutions are shown in blue. For the N/A plot, none of the discovered motifs apply to the architecture.}
\end{figure}
\begin{figure}[H]
    \centering
    \includegraphics[width=\linewidth]{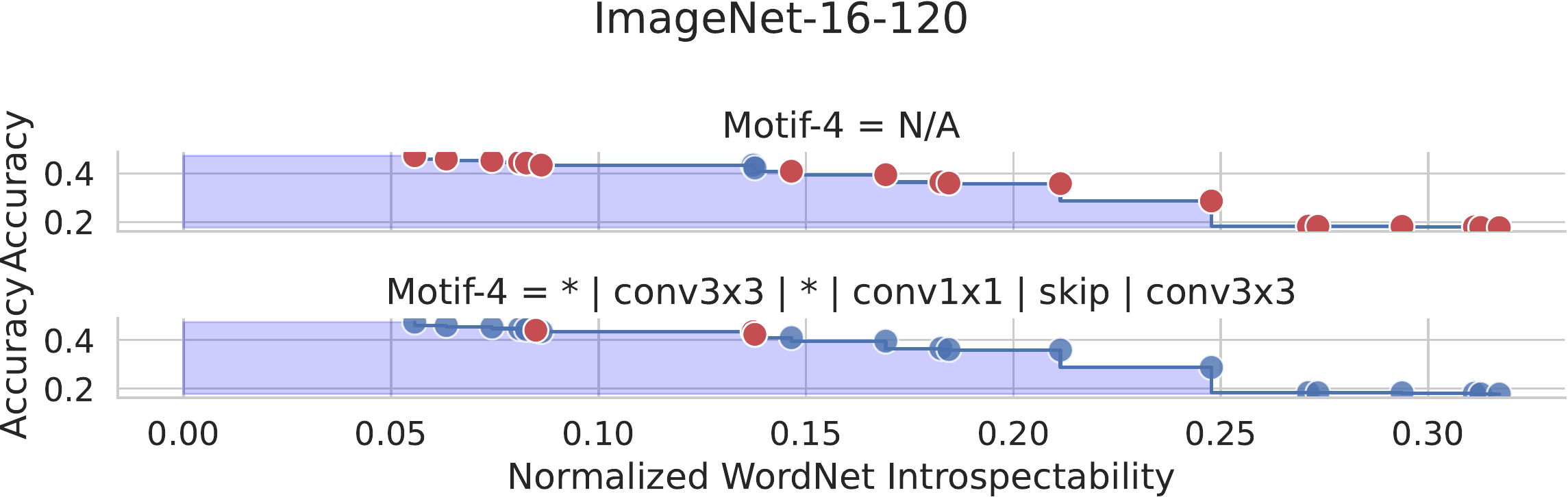}
    \caption{Discovered motifs of size 4 among the Pareto optimal solutions on the ImageNet-16-120 task. See text for description of the motif discovery process. Each red solution indicates that its architecture has the motif shown in the sub-plot title. The remaining solutions are shown in blue. For the N/A plot, none of the discovered motifs apply to the architecture.}
\end{figure}

\ifcsname itIsArxivTime\endcsname%
\else%
\section{Comparing Evolution of Single- and Multi-Objective Search}\setcounter{figure}{0}

We illustrate the evolution of accuracy and introspectability of the models on the Pareto front over each generation in \figref{fig:metrics_over_gens_first}-\figref{fig:metrics_over_gens_last}. Each figure contrasts single-objective with multi-objective optimization to better understand the benefit of NSGA-II in our framework. Note that we do not expect a strict increase in each objective at each generation, which would be expected for population-level statistics, as opposed to statistics within the Pareto front. With single-objective optimization, we can observe that solutions with higher introspectability tend to lie beyond the 95\% confidence interval. This indicates fluke solutions, whereas multi-objective more confidently produces higher-introspectability solutions.

\begin{figure}[h]
    \centering
    \begin{subfigure}[b]{.6\linewidth}%
        \centering
        \includegraphics[width=\linewidth]{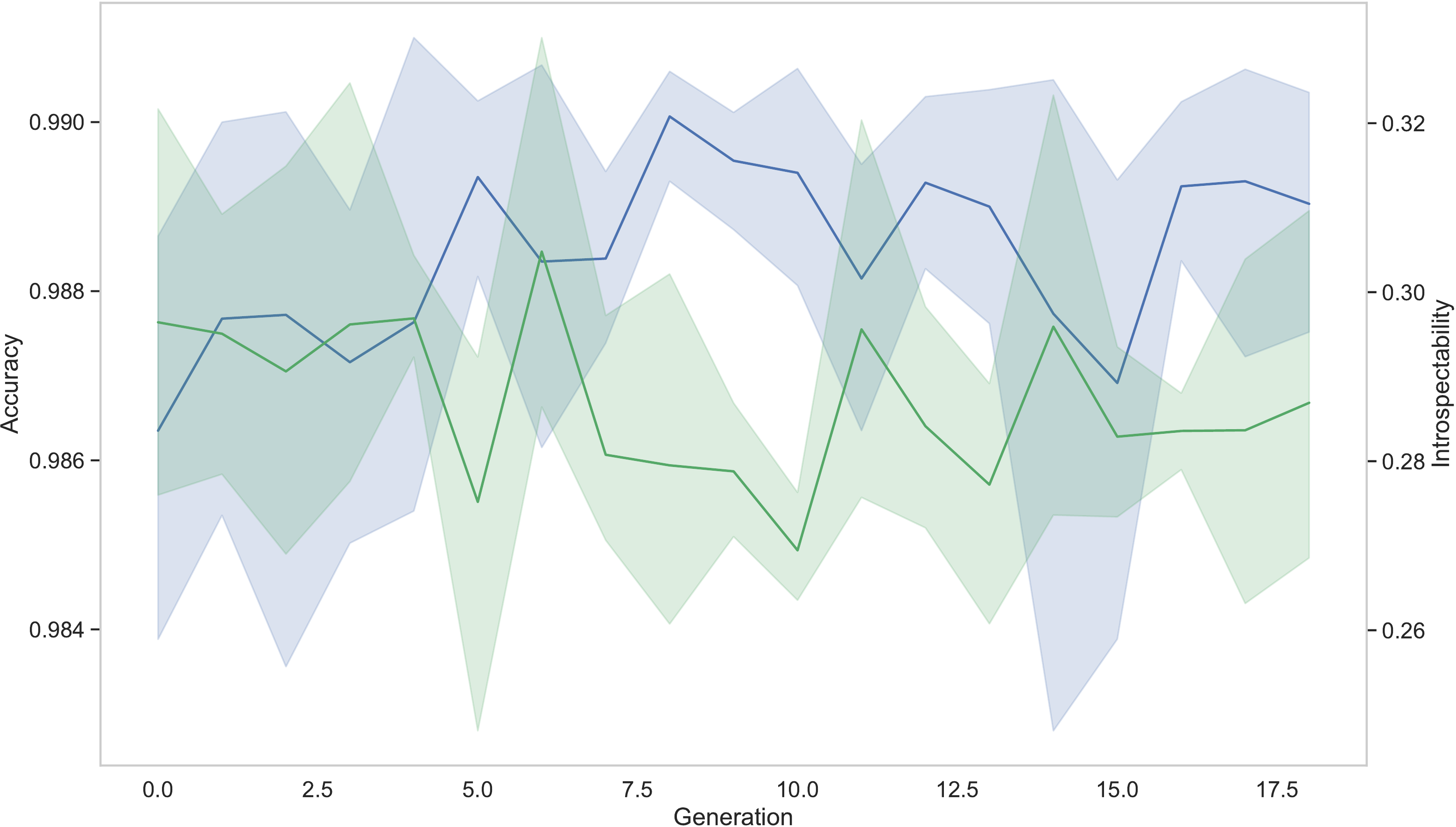}
        \caption{}
        \label{fig:a}
    \end{subfigure}\\%
    \begin{subfigure}[b]{.6\linewidth}%
        \centering
        \includegraphics[width=\linewidth]{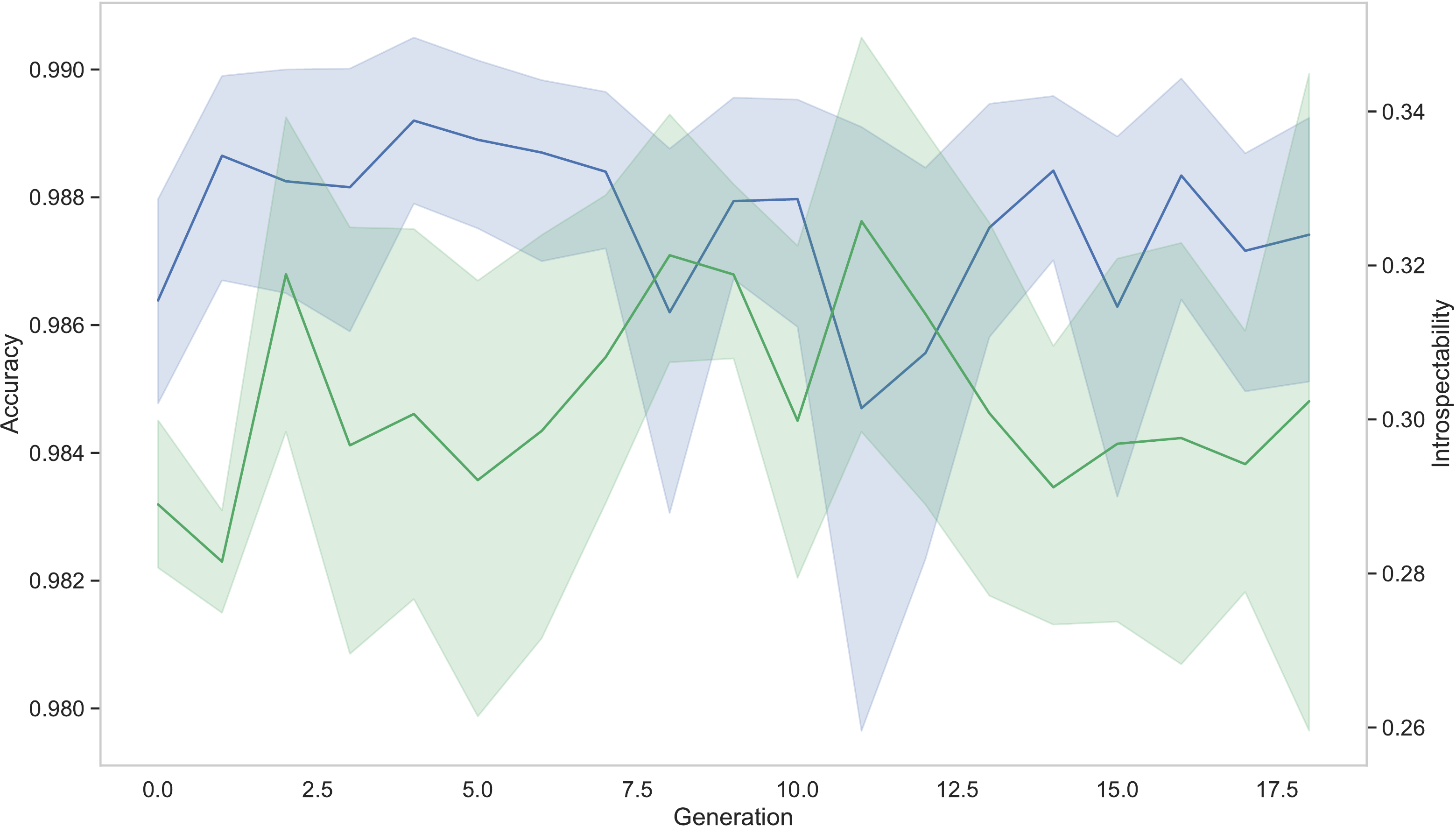}
        \caption{}
        \label{fig:b}
    \end{subfigure}
    \caption{The mean accuracy (blue) and introspectability (green) of the Pareto front solutions each generation on the MNIST task. (a) single-objective; (b) multi-objective. The shaded region indicates the 95\% confidence interval of solutions at each generation.}
    \label{fig:metrics_over_gens_first}
\end{figure}
\begin{figure}[H]
    \centering
    \begin{subfigure}[b]{.6\linewidth}%
        \centering
        \includegraphics[width=\linewidth]{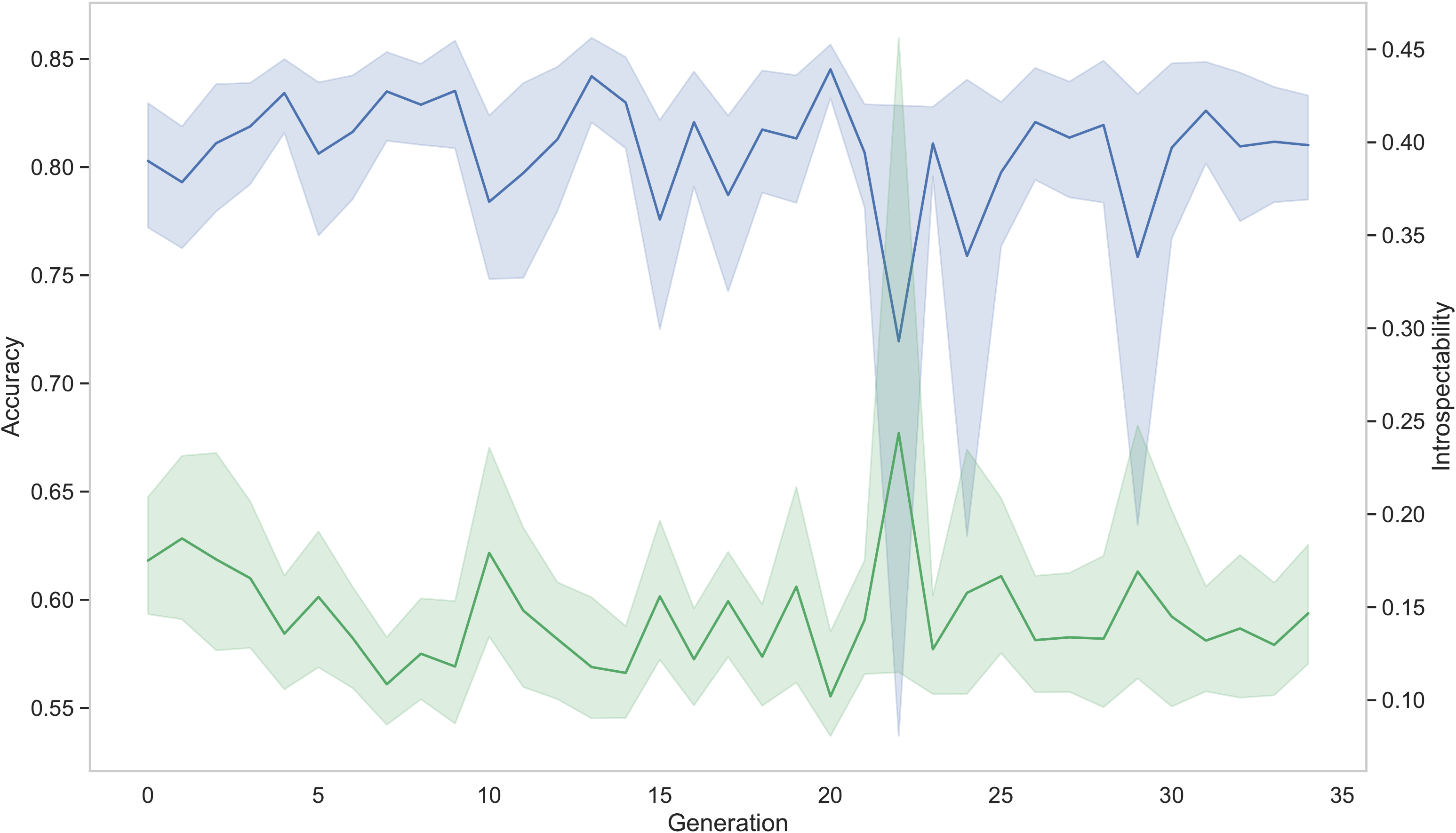}
        \caption{}
        \label{fig:a}
    \end{subfigure}\\%
    \begin{subfigure}[b]{.6\linewidth}%
        \centering
        \includegraphics[width=\linewidth]{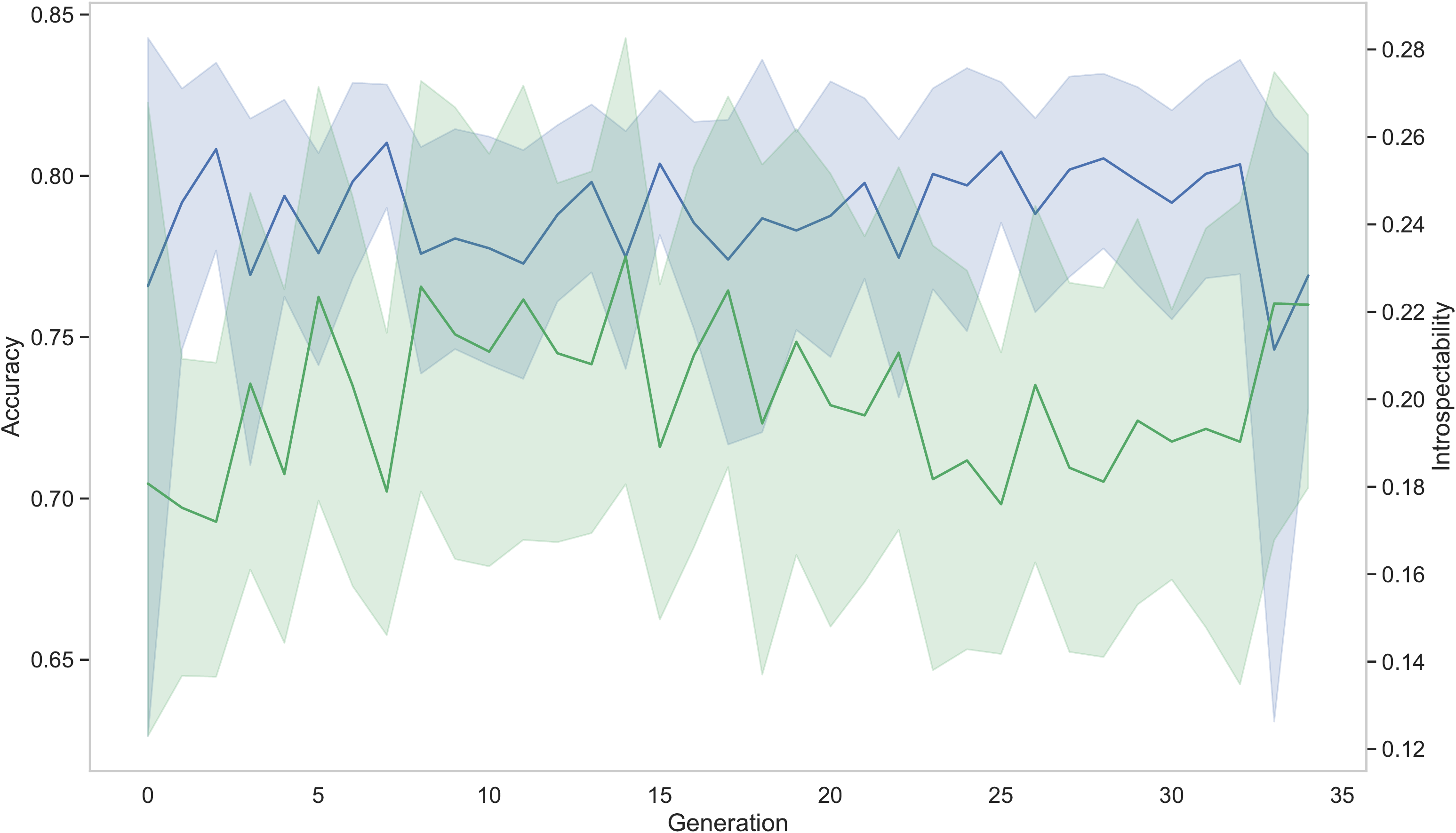}
        \caption{}
        \label{fig:b}
    \end{subfigure}
    \caption{The mean accuracy (blue) and introspectability (green) of the Pareto front solutions each generation on the CIFAR-10 task. (a) single-objective; (b) multi-objective. The shaded region indicates the 95\% confidence interval of solutions at each generation.}
    \label{fig:metrics_over_gens}
\end{figure}
\begin{figure}[H]
    \centering
    \begin{subfigure}[b]{.6\linewidth}%
        \centering
        \includegraphics[width=\linewidth]{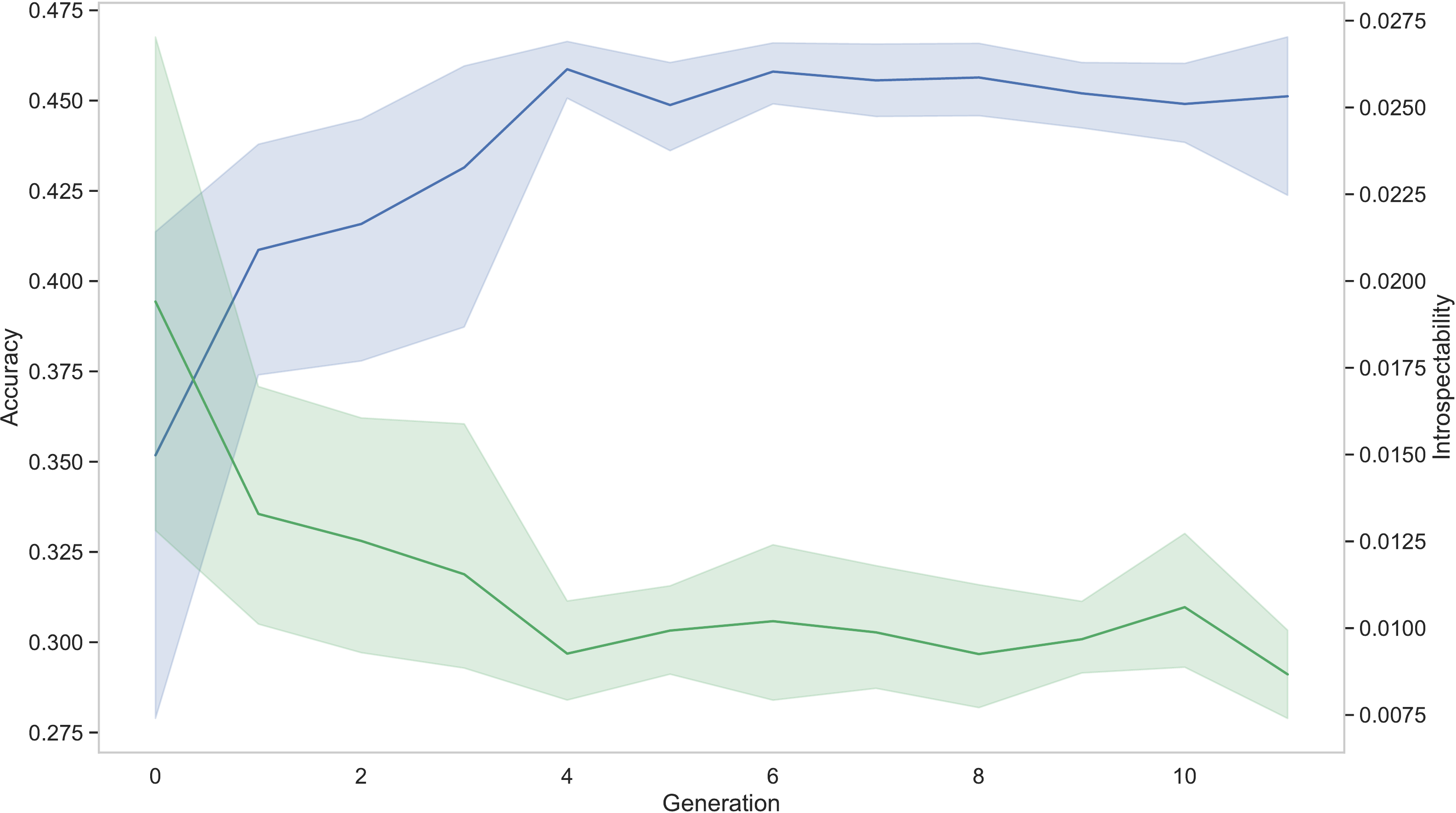}
        \caption{}
        \label{fig:a}
    \end{subfigure}\\%
    \begin{subfigure}[b]{.6\linewidth}%
        \centering
        \includegraphics[width=\linewidth]{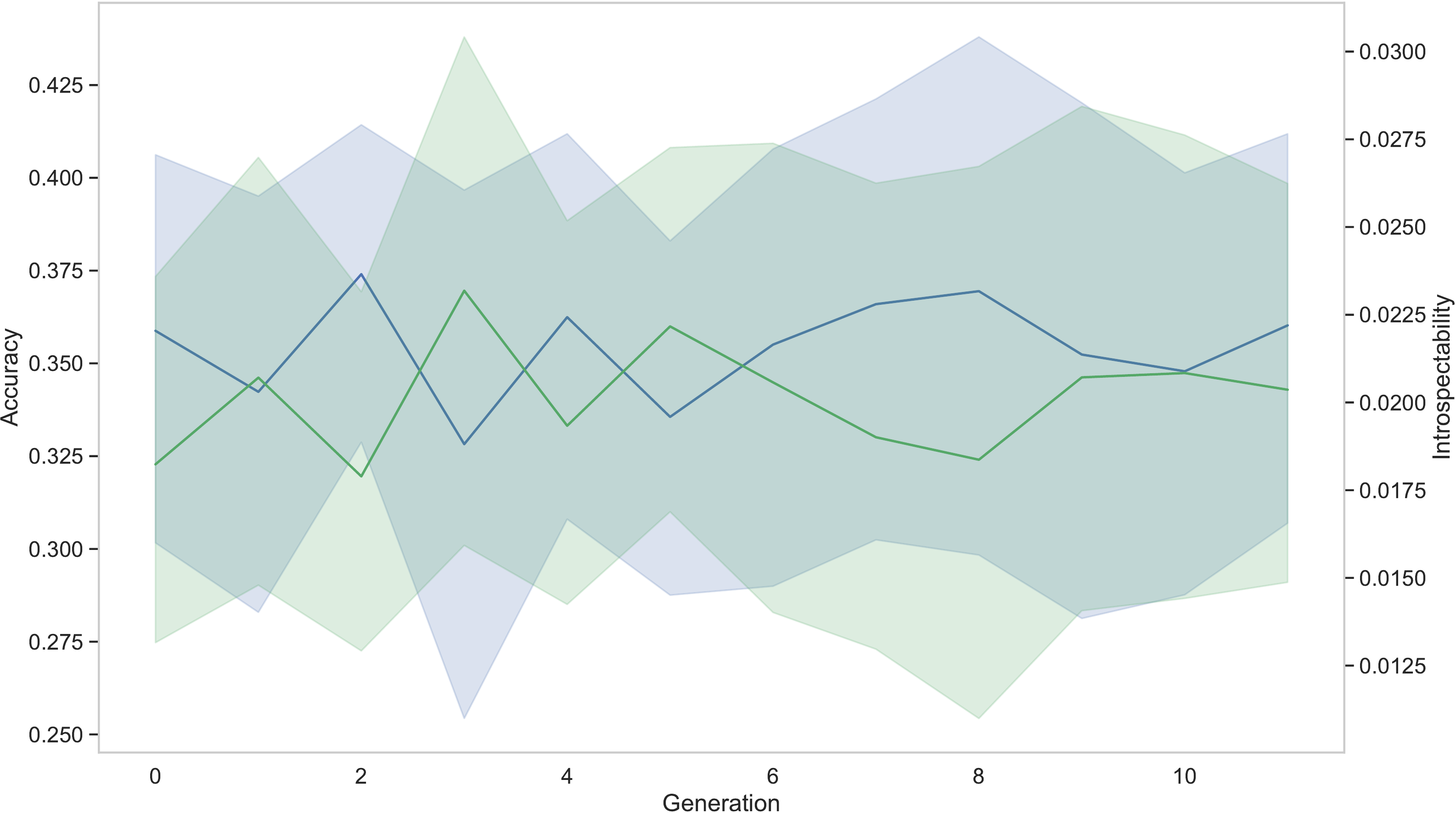}
        \caption{}
        \label{fig:b}
    \end{subfigure}
    \caption{The mean accuracy (blue) and introspectability (green) of the Pareto front solutions each generation on the ImageNet-16-120 task. (a) single-objective; (b) multi-objective. The shaded region indicates the 95\% confidence interval of solutions at each generation.}
    \label{fig:metrics_over_gens_last}
\end{figure}
\fi%

\section{Comparing \XNAS{} Accuracy with Related NAS Methods}

We compare \XNAS{} to other multi-objective approaches on the CIFAR-10 task. Building on the collected results and approach from~\cite{nsga-net}, we take the architecture with the best accuracy and increase the number of filters by a factor of four. We then perform full training on the CIFAR-10 dataset for 200 epochs. The comparison of results and methods is shown in Table~\ref{tab:nsganetRelated}. While \XNAS{} does not achieve the best accuracy (nor was this the objective of this research), the result is still competitive, especially considering the trade-off between accuracy and introspectability.

\begin{table}[H]
    \small
    \centering
    \ra{1.1}
    \begin{tabular}{@{}cccc@{}}
        \toprule
        Method & Error & Other Objective & Compute \\
        \midrule
        \makecell{PPP-Net\\\cite{ppp-net}} & 4.36\% & \makecell{FLOPs, \# \\ parameters, or \\ inference time} & Nvidia Titan X \\
        \makecell{MONAS\\\cite{monas}} & 4.34\% & Power & Nvidia 1080 Ti \\
        \makecell{NSGA-Net\\\cite{nsga-net}} & 3.85\% & FLOPs & \makecell{Nvidia 1080 Ti\\8 GPU Days} \\
        \XNAS{} & 4.45\% & Introspectability & \makecell{Nvidia Tesla P100 \\ 6 GPU Days} \\
        \bottomrule
    \end{tabular}
    \caption{Multi-objective methods for CIFAR-10 (best accuracy
for each method). Table adapted from~\cite{nsga-net}%
    }
    \label{tab:nsganetRelated}
\end{table}

\section{Additional Ablation Studies}

\begin{figure}[H]
    \centering
    \begin{subfigure}[b]{\linewidth}%
        \centering
        \includegraphics[width=\linewidth]{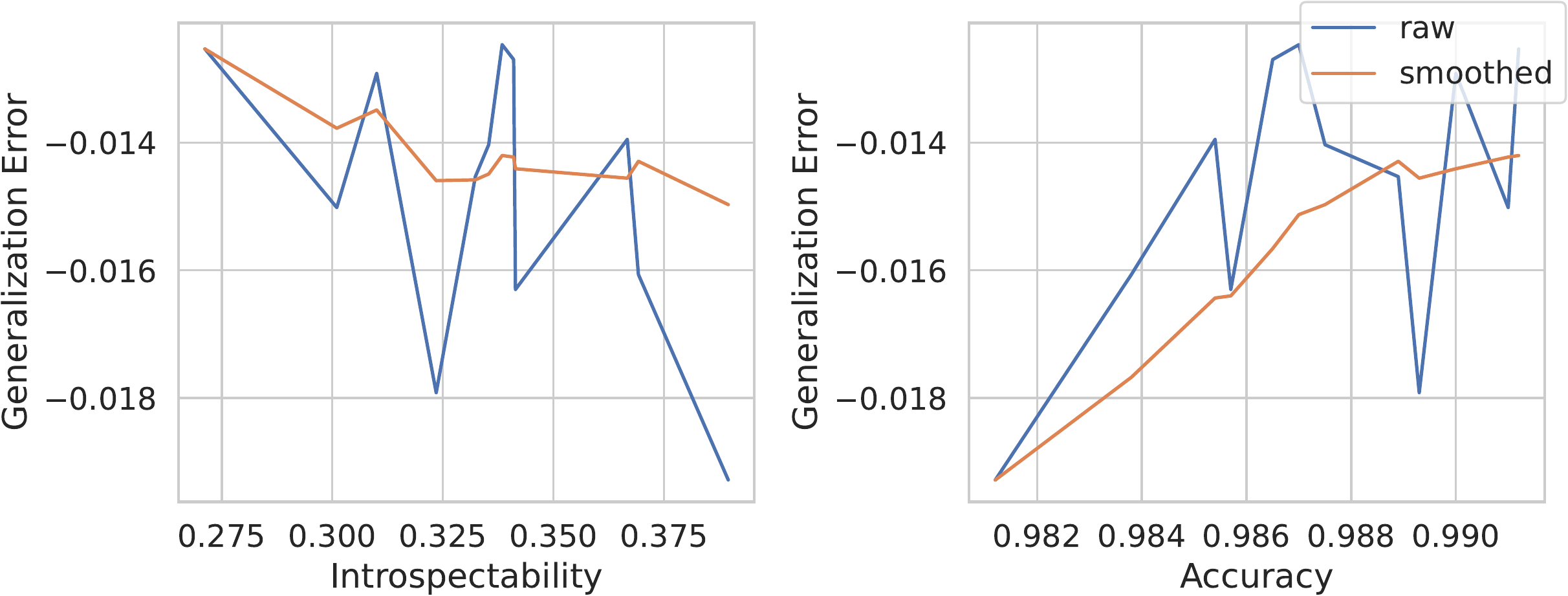}
        \caption{MNIST}
        \label{fig:a}
    \end{subfigure}\\%
    \begin{subfigure}[b]{\linewidth}%
        \centering
        \includegraphics[width=\linewidth]{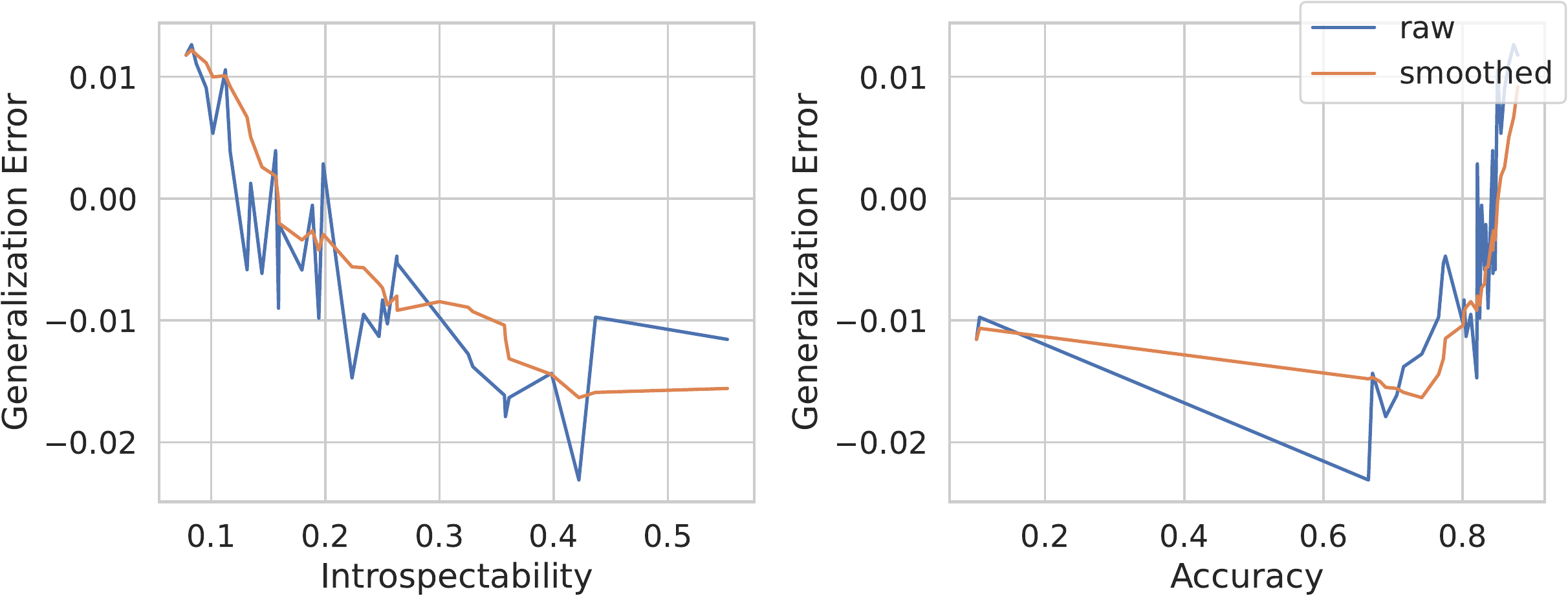}
        \caption{CIFAR-10}
        \label{fig:b}
    \end{subfigure}\\%
    \begin{subfigure}[b]{\linewidth}%
        \centering
        \includegraphics[width=\linewidth]{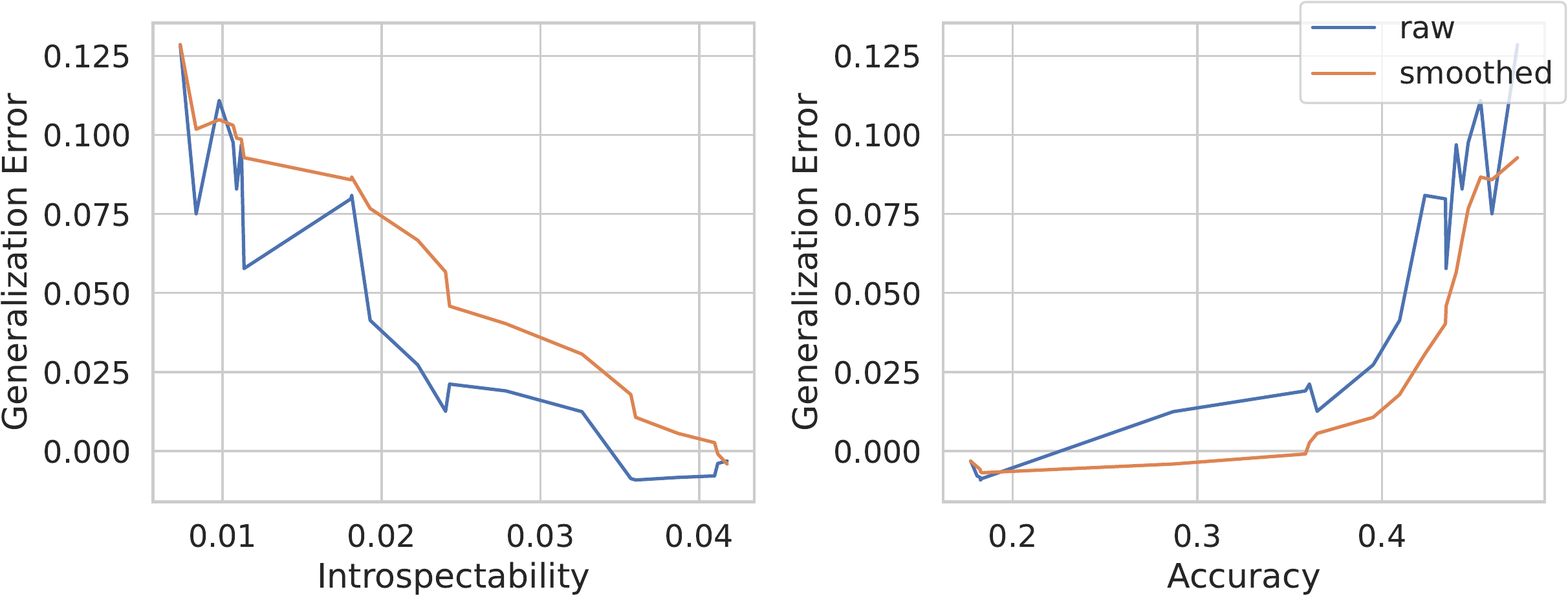}
        \caption{ImageNet-16-120}
        \label{fig:c}
    \end{subfigure}
    \caption{The effect of introspectability and accuracy on generalization error across the tasks.}
    \label{fig:ablate_gen}
\end{figure}

\begin{figure}[H]
    \centering
    \begin{subfigure}[b]{\linewidth}%
        \centering
        {\catcode`\#=12\includegraphics[width=\linewidth]{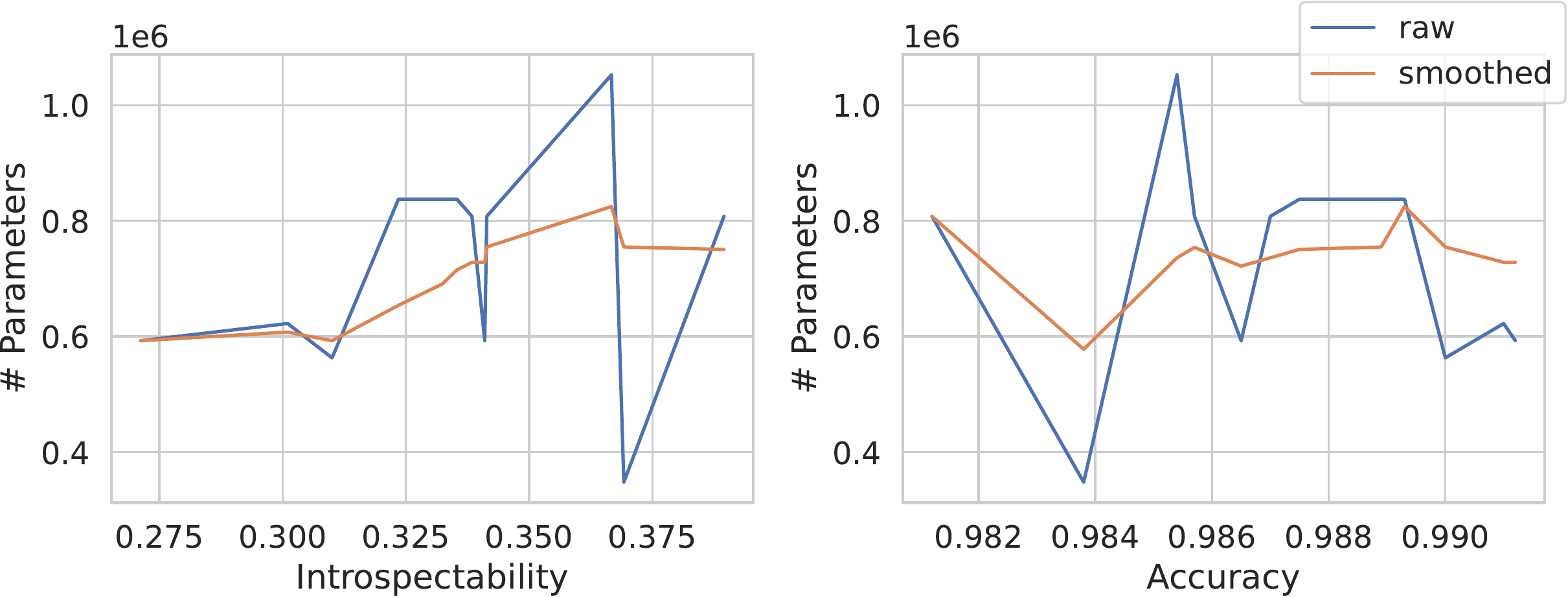}}
        \caption{MNIST}
        \label{fig:a}
    \end{subfigure}\\%
    \begin{subfigure}[b]{\linewidth}%
        \centering
        {\catcode`\#=12\includegraphics[width=\linewidth]{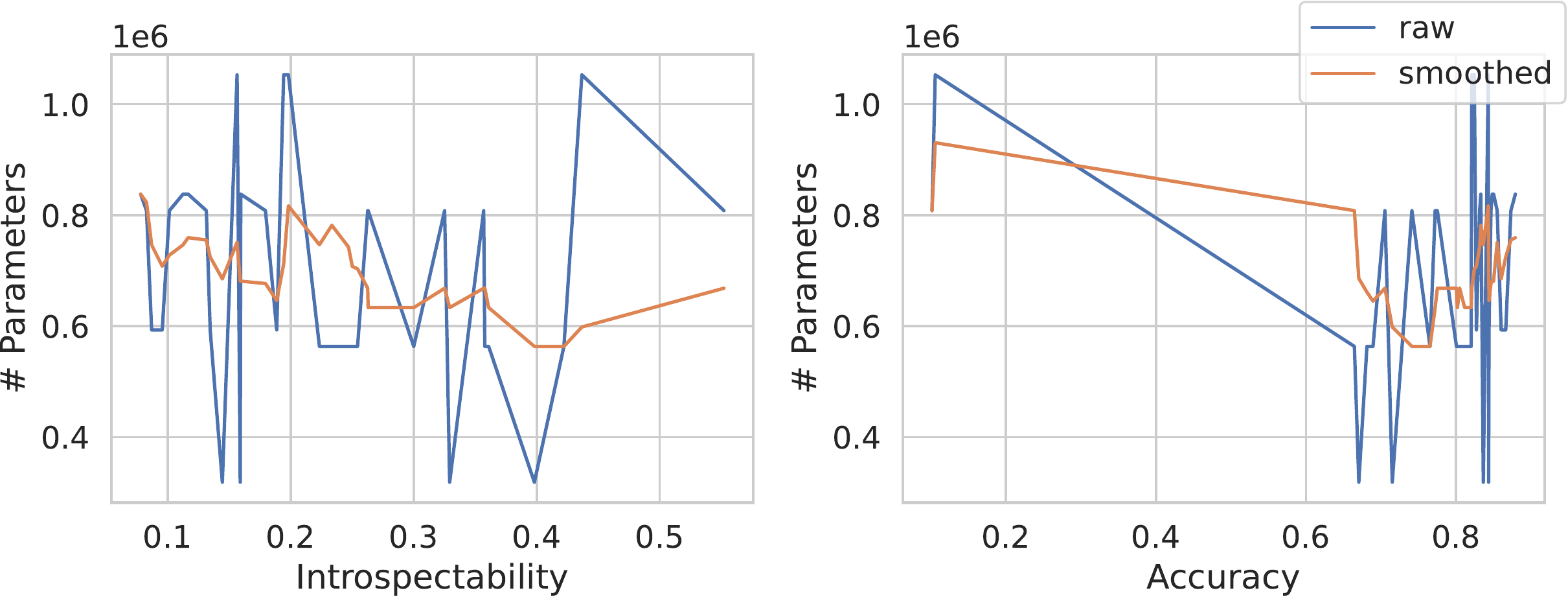}}
        \caption{CIFAR-10}
        \label{fig:b}
    \end{subfigure}\\%
    \begin{subfigure}[b]{\linewidth}%
        \centering
        {\catcode`\#=12\includegraphics[width=\linewidth]{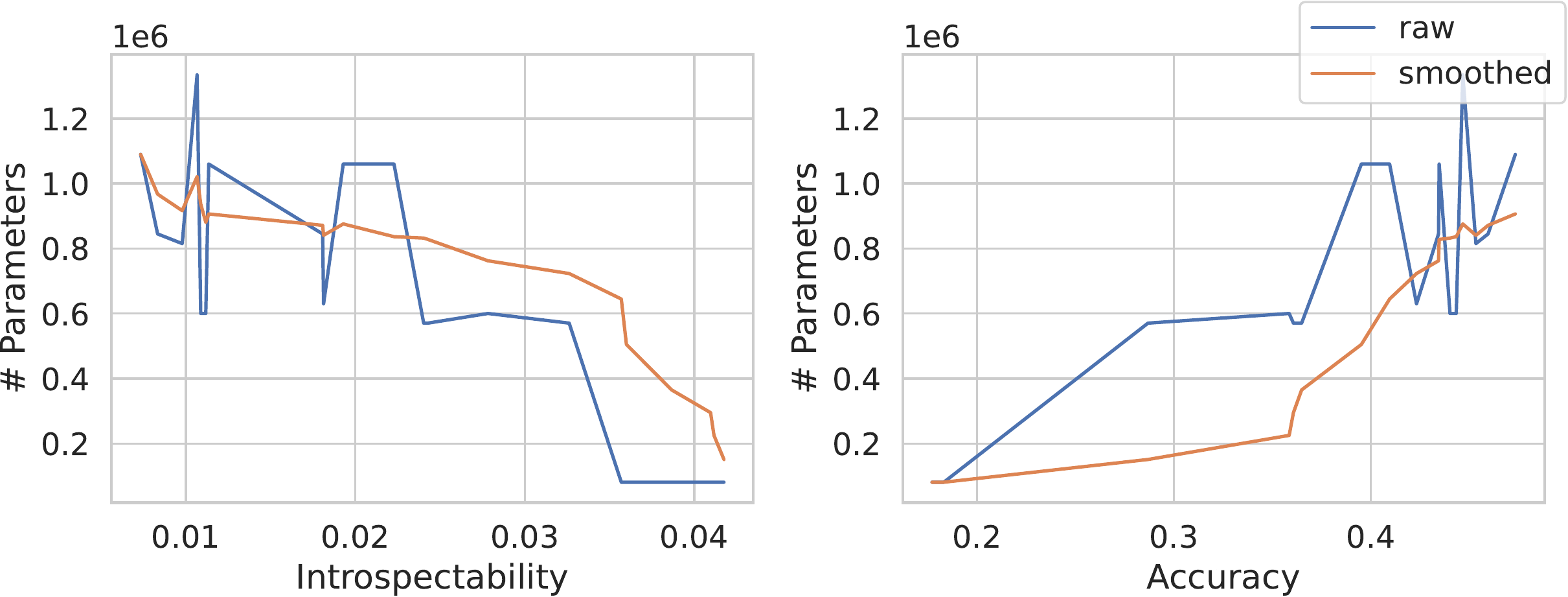}}
        \caption{ImageNet-16-120}
        \label{fig:c}
    \end{subfigure}
    \caption{The effect of introspectability and accuracy on the number of parameters across the tasks.}
    \label{fig:ablate_params}
\end{figure}

\begin{figure}[H]
    \centering
    \begin{subfigure}[b]{\linewidth}%
        \centering
        \includegraphics[width=\linewidth]{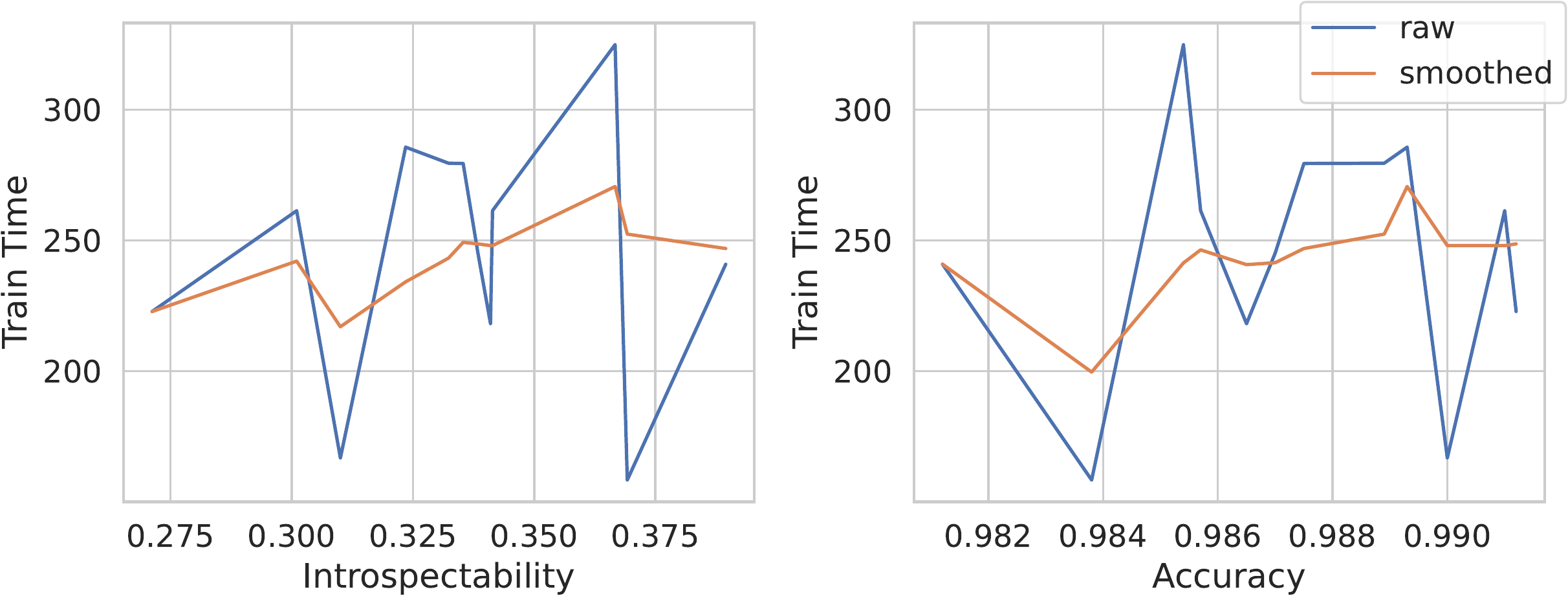}
        \caption{MNIST}
        \label{fig:a}
    \end{subfigure}\\%
    \begin{subfigure}[b]{\linewidth}%
        \centering
        \includegraphics[width=\linewidth]{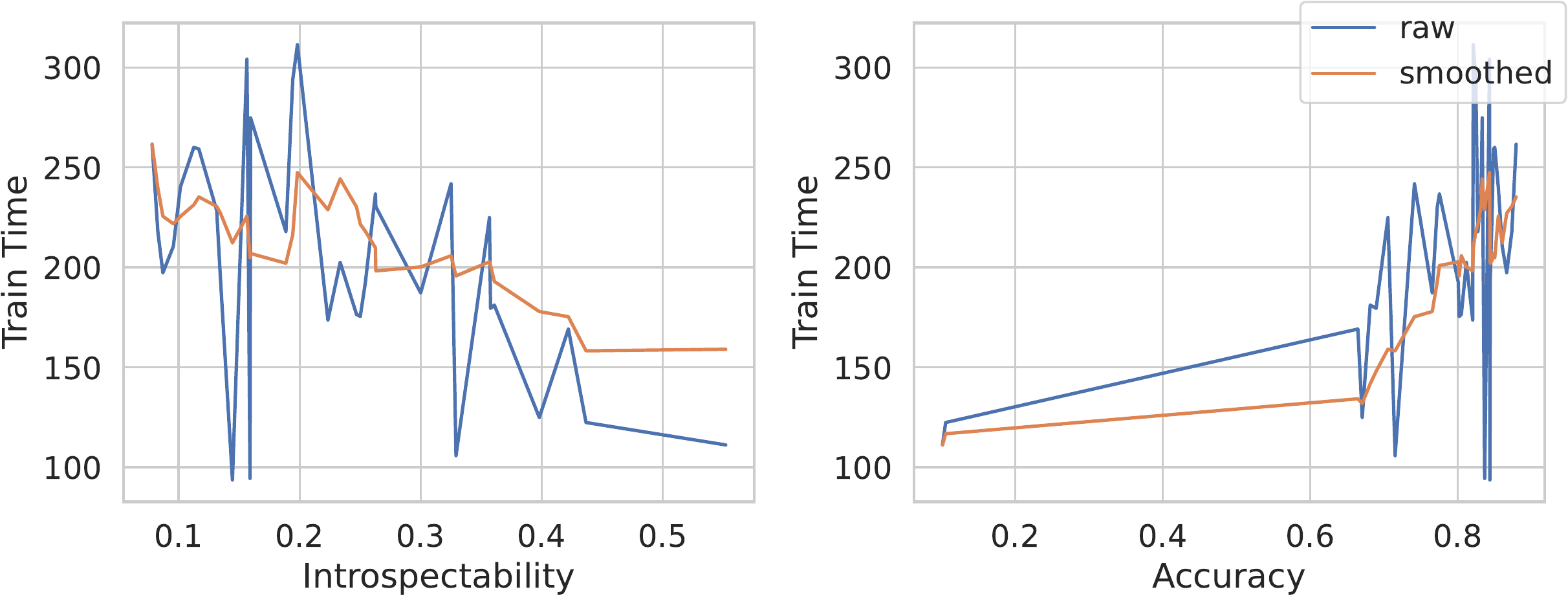}
        \caption{CIFAR-10}
        \label{fig:b}
    \end{subfigure}\\%
    \begin{subfigure}[b]{\linewidth}%
        \centering
        \includegraphics[width=\linewidth]{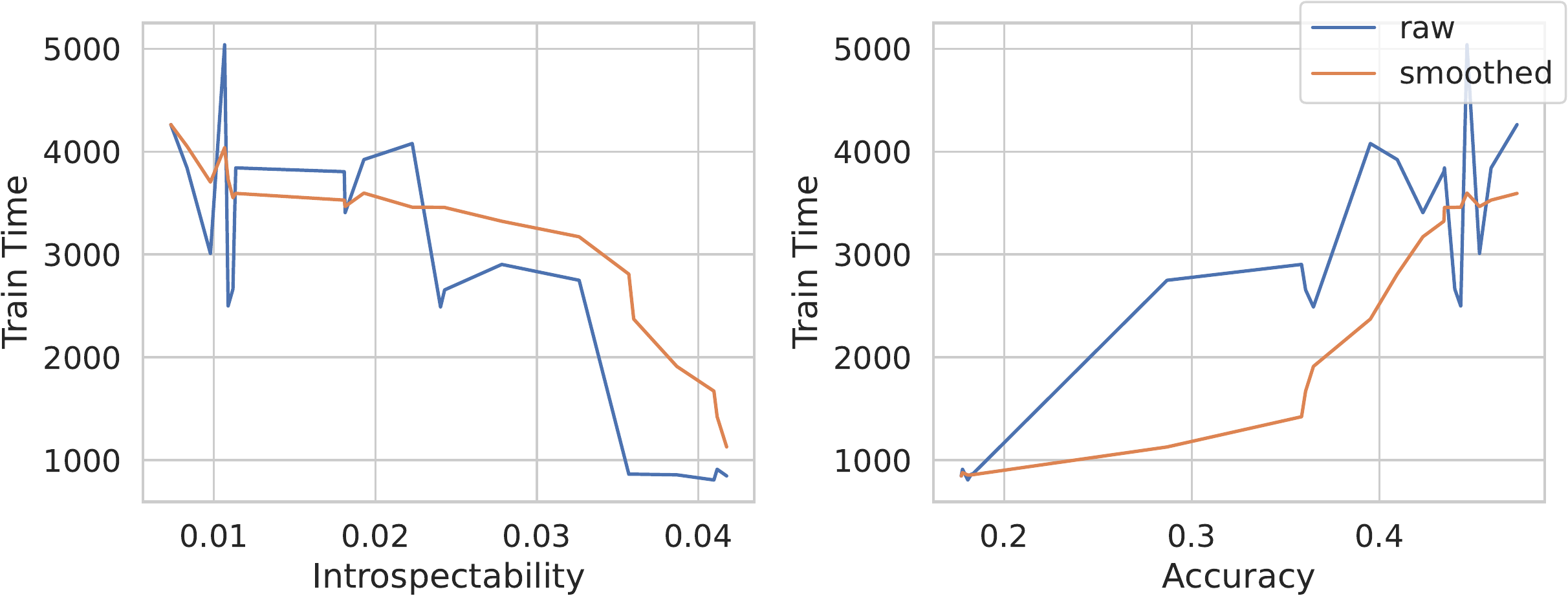}
        \caption{ImageNet-16-120}
        \label{fig:c}
    \end{subfigure}
    \caption{The effect of introspectability and accuracy on the training speed across the tasks.}
    \label{fig:ablate_time}
\end{figure}

We perform additional studies to understand the relationships between the objectives, accuracy and introspectability, and the generalization error, number of parameters, and training speed of architectures. Figure~\ref{fig:ablate_gen} demonstrates that introspectability and accuracy have an inverse relationship on the generalization error -- this error increases with high-accuracy models and decreases with high-introspectability models. Likewise, the trend can be observed with the number of parameters and training speed as shown in Figures~\ref{fig:ablate_params} and~\ref{fig:ablate_time}, respectively. These figures follow a similar trend as the number of parameters correlates with the number of FLOPs and thus the training time. As discussed in Section 4.4 of the main text, high-introspectability networks tend to have a more pooling layers whereas high-accuracy networks have more convolutional layers. This helps to explain the trends observed in the number of parameters. A takeaway from this analysis is that the trade-off between accuracy and introspectability also implies a trade-off in parameters (and FLOPs), training time, and generalization error.

\section{Model Debugging Experiments}

Here, we study the ability of our activations calibration approach to correct bugs in models. We first demonstrate that there is a strong connection between the pairwise activation distances used in the formulation of introspectability and the ground truth confusion matrix. To make this comparison, the pairwise distances are negated as disentanglement (separation) between class representations is posited to correlate with confounding. Since the distance between the activations of a class and itself is 0, ideally, the distance between such and the activations of other classes is maximized. There is no information about ground truth available in computing pairwise distances, i.e.\ the computation is symmetrical and unconditioned. In turn, we compare this information to a confusion matrix folded along the diagonal. This means that element $x_{i, j}, i\ne j$ in the folded confusion matrix is equivalent to the sum $x_{i, j} + x_{j, i}$ in the original confusion matrix. On a higher level, each element $x_{i, j}$ is either the number of true positives for a class (when considering the diagonal), or the support of class $i$ being predicted when class $j$ were true and the support of the converse. To support this, we measure the correlation between the negated pairwise activation distances and the folded ground truth confusion matrix across Pareto optimal models trained on CIFAR-10. High-introspectability models achieve a correlation of $\rho = 0.85$ while low-introspectability models achieve a correlation of $\rho = 0.59$. With this motivation, we demonstrate how model bugs can be identified and corrected in the following case study.

\begin{figure}[H]
    \centering
    \includegraphics[width=\linewidth]{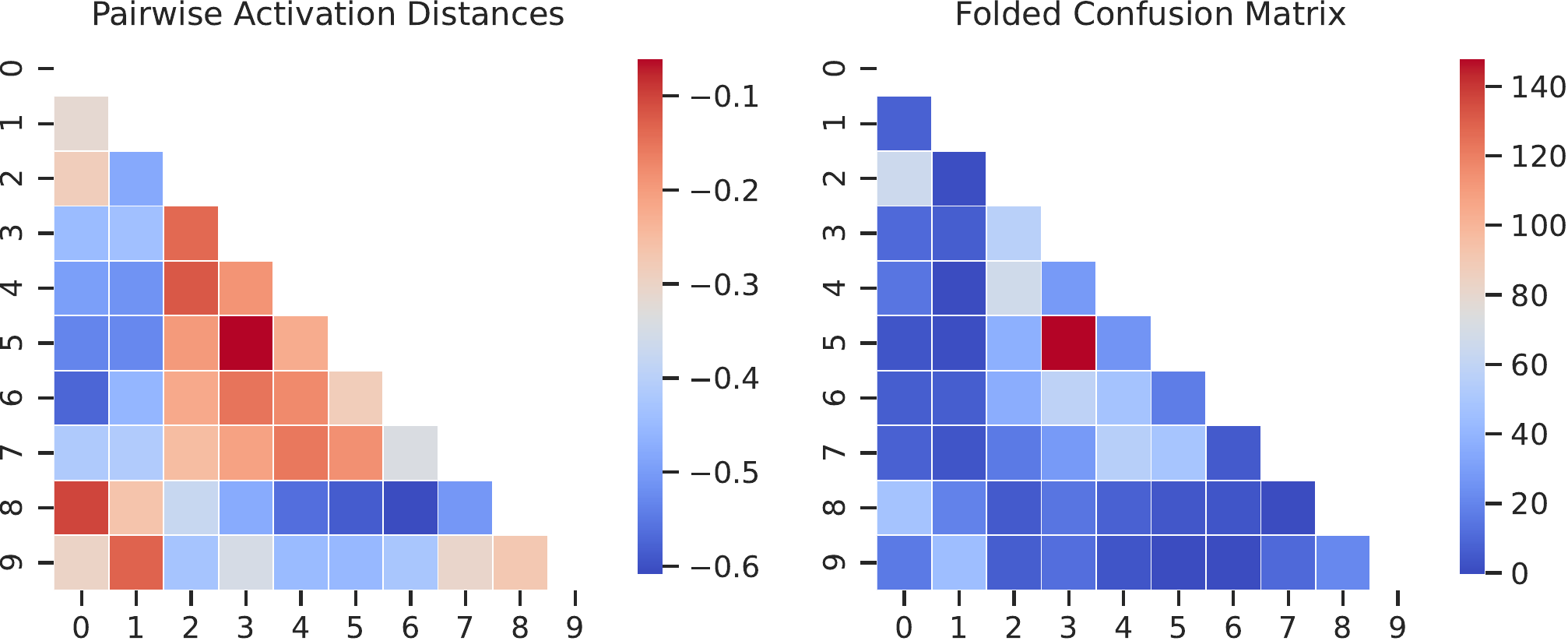}
    \caption{Heat maps of (left) the negated pairwise activation distances as part of the introspectability computation, and (right) the ground truth confusion matrix folded along the diagonal. The heat maps are shown for a random model trained on CIFAR-10. As can be seen, the model confounds the classes 3 and 5, which is reflected in the distance between activations.}
    \label{fig:cm_corr}
\end{figure}

\paragraph{Case Study: Bug Identification and Correction}
Figure~\ref{fig:cm_corr} demonstrates for a random higher-introspectability model trained on CIFAR-10 that there is strong correlation between the negated pairwise activation distances and the ground truth confusion matrix folded along the diagonal ($\rho = 0.81$). Noticeably, the model confounds the classes 3 and 5, which is reflected in the pairwise activations as the smallest distance (largest negated distance).

\begin{figure}
    \centering
    \includegraphics[width=\linewidth]{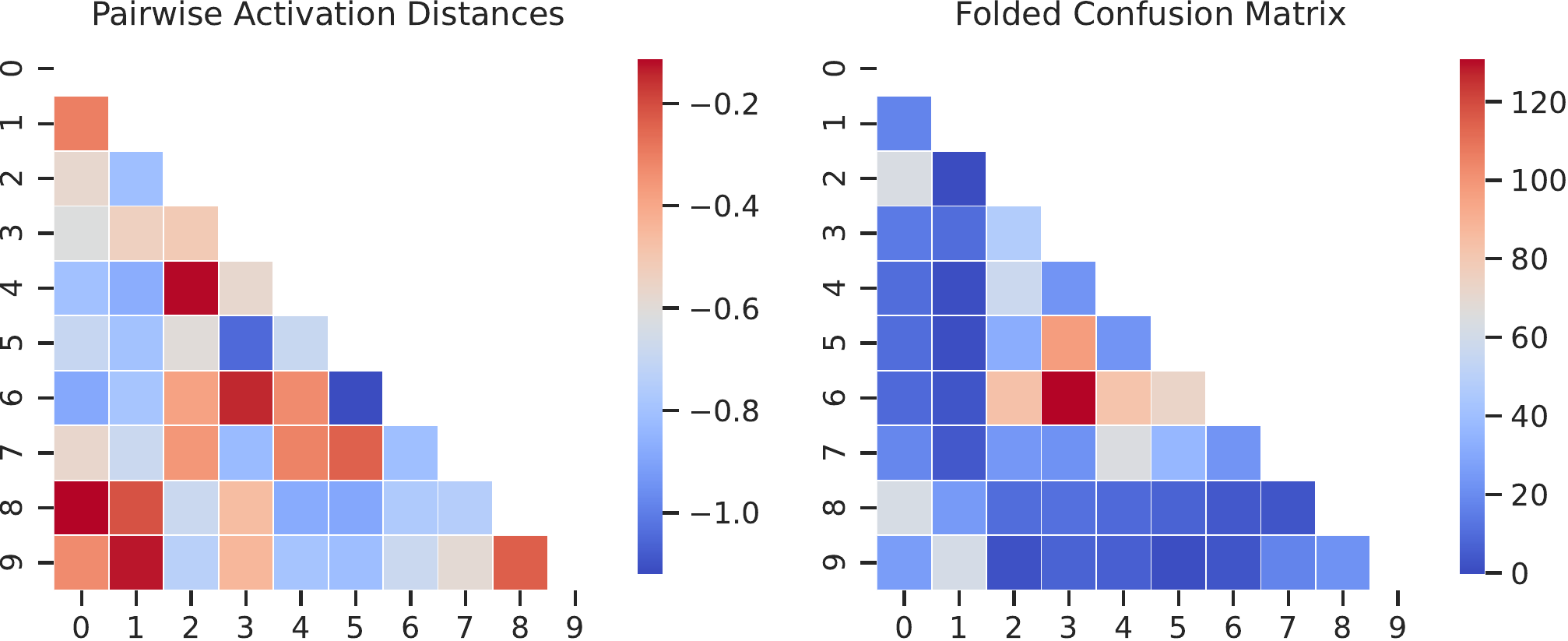}
    \caption{Heat maps after the correction procedure is run of (left) the negated pairwise activation distances as part of the introspectability computation, and (right) the ground truth confusion matrix folded along the diagonal. Note the difference in color scale from Figure~\ref{fig:cm_corr}.}
    \label{fig:cm_corr_fixed}
\end{figure}

With the bug in the model identified, we formulate a strategy to mitigate the issue. The key of our approach is to push the representations of classes 3 and 5 apart in order to reduce the confounding of one another. We accomplish this by using the introspectability regularizer approach with pairwise coefficients. The generalization of this to arbitrary pairs is formalized in Eq.\ \eqref{eq:introspectability_targetted_regularizer}.
\begin{align}
\begin{split}\label{eq:introspectability_targetted_regularizer}
\text{Introspectability}_{\text{reg}}(\mathcal{M}, \tensor{X}) =\\
\frac{-1}{\binom{N_C}{2}} \sum_{c = 1}^{N_C} \sum_{k = c + 1}^{N_C} D(\bar{\mathbf{\Phi}}^{(c)}, \bar{\mathbf{\Phi}}^{(k)}) \times \omega_{i,j}
\end{split}
\end{align}
\noindent
where $\omega_{i,j}$ is a weight for each class pair $(i, j)$. With every $\omega_{i,j} = 1$ this is equivalent to the untargeted introspectability regularizer. If the aim is to target all confounded predictions, one can set all $\omega_{i,j}$ proportionally to the pairwise activation distances (or folded confusion matrix). However, we target a single pair in this case study.
In our experiment, the model is trained with the regularization term for an additional 5 epochs, a learning rate of 0.001, $\omega_{3,5} = 25$, and all other $\omega_{i,j} = 1$. The results are visualized in Figure~\ref{fig:cm_corr_fixed}.
The approach is stronger in identifying bugs than mitigating them, although there is improvement without significant degradation of accuracy~($\pm 0.6\%$ across 10 trials). We leave the tuning of hyperparameters and alternative weighting schemes to future exploration.

\section{Extended Background}
\paragraph{DNN Inspection within Explainable AI (XAI)}

The opaque nature of deep neural networks~(DNNs) has ultimately led to the sub-field of explainable AI (XAI)~\cite{darpa_xai}, which was denominated in 2016 by DARPA, although relevant work predates this by years.
Relevant to the subject matter of this work are XAI methods of DNN inspection.
This suite of methods enables the debugging of model behavior, the detection of dataset errors, and the development of adversarial attacks.
The authors of~\cite{DBLP:conf/icml/KohL17} scale influence functions, a robust statistics method, to DNNs to understand the effect of training points on a prediction.
DNN visualization tools have been proposed to provide qualitative modes of analysis. Notably, \cite{erhan2009visualizing,DBLP:journals/corr/YosinskiCNFL15} provide tools for visualizations by gradient ascent, deconvolution for highlighting input images, and discovering preferred input patterns for each class.
Probing-based methods aim to qualify the role of DNN internal elements (neurons, latent representations, etc.). In~\cite{DBLP:conf/icml/KimWGCWVS18,DBLP:journals/pnas/BauZSLZ020}, methods are proposed to relate DNN internals to semantic concepts, such as textures, shapes, colors, or even people. Another approach introduced in~\cite{DBLP:conf/nips/GhorbaniZ20} is to use Shapley values from game theory to quantify the influence each neuron has on overall DNN error.

More related to our work are those related to disentanglement, i.e.\ the separation of concept- or class-relevant information in a network. For instance, \cite{DBLP:conf/cvpr/ZhangWZ18a} proposes the learning of interpretable CNN filters by coercing feature maps to resemble hand-crafted templates. Moreover, the variational autoencoder~(VAE)~\cite{DBLP:journals/corr/KingmaW13} has been extended to produce a disentangled latent space by regularizing the bottleneck layer~\cite{DBLP:conf/iclr/HigginsMPBGBML17}. 
In contrast, we optimize for DNNs with disentangled internal representations of classes without explicit constraints on the loss, modifications to the architecture, or hand-crafted activation patterns.
This also allows for the use of non-differentiable objectives.